%% file: neurips_2026.tex
\pdfoutput=1
\documentclass{article}

\input{math_commands.tex}

\usepackage{natbib}
\usepackage[main, final]{neurips_2026}

\usepackage[utf8]{inputenc} %
\usepackage[T1]{fontenc}    %
\usepackage{hyperref}       %
\usepackage{url}            %
\usepackage{booktabs}       %
\usepackage{amsfonts}       %
\usepackage{nicefrac}       %
\usepackage{microtype}      %
\usepackage{xcolor}         %
\title{Meta-RL with Bayesian Linear Task Models}

\author{%
  Jingyang You and Hanna Kurniawati \\
  School of Computing\\
  Australian National University\\
  \texttt{\{jingyang.you, hanna.kurniawati\}@anu.edu.au} \\
}

\begin{document}

\maketitle

\begin{abstract}
Deep Bayesian reinforcement learning adapts to unseen tasks by inferring latent transition and reward models, but existing methods typically rely on variational posteriors and evidence lower bounds, introducing approximation error and unstable task representations. We introduce \nom, a deep Bayesian RL framework that combines generalised linear task models with learnable non-linear basis functions. \nom features conjugate Bayesian inference, yielding exact, sequential posterior updates over task parameters and model noise, together with a closed-form marginal likelihood that eliminates variational inference. The update is naturally permutation-invariant, allowing \nom to integrate with both off- and on-policy algorithms. \nom also learns task representation admitting an exact kernel identity, relating distances between task representations to kernel discrepancies over the task contexts. Compared against eight representative or recent meta reinforcement learning methods, \nom achieves the highest aggregate zero-shot test performance on both the MuJoCo locomotion and MetaWorld manipulation benchmarks. 
\end{abstract}

\input{sections/intro}
\input{sections/related}
\input{sections/background}
\input{sections/method}
\input{sections/experiment}
\input{sections/conclusion}

\newpage
\section*{Acknowledgement}
We thank anonymous reviewers for their valuable feedback. This work was supported by the ARC Research Hub in Intelligent Robotic Systems for Real-Time Asset Management (IH210100030). Hanna Kurniawati is partially supported by the Australasian Space Innovation Institute (fka SmartSat CRC).

\bibliography{neurips_2026}
\bibliographystyle{plain}

\appendix
\input{sections/appendix}

\end{document}

%% file: math_commands.tex
\usepackage{amsmath,amsfonts,bm}
\usepackage{xspace}
\usepackage{multicol}
\usepackage{amssymb, setspace}
\usepackage{xr}
\usepackage{graphicx}
\usepackage{subcaption}
\usepackage{booktabs}
\usepackage{array}
\usepackage{yhmath}

\usepackage[ruled,vlined]{algorithm2e}

\newcommand\labelAndRemember[2]
  {\expandafter\gdef\csname labeled:#1\endcsname{#2}%
   \label{#1}#2}
\newcommand\recallLabel[1]
   {\csname labeled:#1\endcsname\tag{\ref{#1}}}

\def\Figref#1{Figure~\ref{#1}}

\def\Secref#1{Section~\ref{#1}}

\def\eqref#1{equation~\ref{#1}}
\def\Eqref#1{Equation~\ref{#1}}

\def\Algref#1{Algorithm~\ref{#1}}

\def\Apref#1{Appendix~\ref{#1}}

\def\1{\bm{1}}

\def\thetat{\theta_T}
\def\thetar{\theta_R}
\def\Thetat{\Theta_T}
\def\Thetar{\Theta_R}

\def\phit{\phi_T}
\def\phir{\phi_R}

\def\mut{T_\mu}
\def\sigmat{T_\sigma}
\def\mur{R_\mu}
\def\sigmar{R_\sigma}

\def\Ct{\mathbf{C}_T}
\def\Cr{\mathbf{C}_R}

\def\Mt{\mathbf{M}_T}
\def\Mr{\mathbf{M}_R}

\def\xit{\boldsymbol{\Xi}_T}
\def\xir{\boldsymbol{\Xi}_R}

\def\omegat{\boldsymbol{\Omega}_T}
\def\omegar{\boldsymbol{\Omega}_R}

\def\nut{\nu_T}
\def\nur{\nu_R}

\def\SigmaT{\boldsymbol{\Sigma}_T}
\def\SigmaR{\boldsymbol{\Sigma}_R}

\def\alpaca{ALPaCA\xspace}
\def\pearl{PEARL\xspace}
\def\vbad{VariBAD\xspace}
\def\rltwo{RL$^2$\xspace}
\def\maml{MAML\xspace}
\def\nom{GLiBRL\xspace}

\DeclareMathAlphabet{\mathsfit}{\encodingdefault}{\sfdefault}{m}{sl}
\SetMathAlphabet{\mathsfit}{bold}{\encodingdefault}{\sfdefault}{bx}{n}

\DeclareMathOperator{\Tr}{Tr}

%% file: sections/intro.tex
\section{Introduction}
Reinforcement Learning (RL) algorithms possess significant potential to enable autonomous, intelligent robotic control. However, standard RL algorithms typically do not account for variations in transition and reward dynamics, causing them to generalise poorly to unseen tasks where the underlying environmental models deviate from those encountered during training. Bayesian Reinforcement Learning (BRL), a special case of Meta-Reinforcement Learning (Meta-RL), offers an effective framework to overcome this limitation.

Instead of ignoring environmental variations, BRL explicitly accounts for them by assuming distributions over the transition and reward models, continually performing Bayesian inference on these parameters  \citep{ghavamzadeh2015bayesian}. The resulting parameter distributions implicitly encode varying task dynamics. To solve these BRL problems, many classical methods employ planners \citep{guez2013scalable, poupart2006analytic} to search for Bayes-optimal policies. However, these methods scale poorly and require explicit prior knowledge regarding the structural forms of the transition and reward models, severely restricting their generalisation across diverse tasks.

To address such rigidity, recent deep BRL methods \citep{rakelly2019efficient, zintgraf2021varibad} enable model learning by optimising the marginal likelihood of the observed data. Unfortunately, the direct application of neural networks to the joint space of data and task parameters renders exact Bayesian inference intractable, preventing direct optimisation of the exact marginal likelihood. Consequently, these methods rely on variational inference to optimise the Evidence Lower Bound (ELBO). This introduces challenges such as high-variance Monte Carlo estimates, amortisation gaps \citep{cremer2018inference} and posterior collapse \citep{bowman2016generating, dai2020usual}. In the context of BRL, these issues often prevent the extraction of meaningful and distinctive task parameters which are crucial for effective BRL.

To alleviate these issues, we propose \textbf{G}eneralised \textbf{Li}near Models in Deep \textbf{B}ayesian \textbf{RL} with Learnable Basis Functions (\textbf{\nom})\footnote{\href{https://github.com/RDLLab/GLiBRL}{https://github.com/RDLLab/GLiBRL}}. Inspired by the deep Bayesian linear regression structure proposed in \alpaca \citep{harrison2018meta},  \nom enforces a generalised linear relation between task parameters and  data features computed via learnable basis functions. This architecture circumvents variational inference entirely, permitting exact posterior inference and the computation of a closed-form marginal likelihood. By performing exact inference over the model noise, \nom naturally generalises \alpaca while reducing the error of predictions in unseen tasks. The permutation-invariance of \nom induced by the exact Bayesian update also allows seamless integration with both on- and off-policy RL algorithms. Moreover, we show an exact kernel identity makes \nom's task-representation geometry analytically traceable, relating distances between task representations to signed-measure kernel discrepancies over the task contexts (\Apref{theory}).

We evaluate \nom on the MuJoCo \citep{todorov2012mujoco} locomotion benchmark and the challenging MetaWorld \citep{yu2020meta, mclean2025meta} ML10/45 benchmarks. Compared against a diverse suite of Meta-RL baselines, including \pearl \citep{rakelly2019efficient}, \vbad \citep{zintgraf2021varibad}, SDVT \citep{lee2023parameterizing}, \rltwo \citep{wang2016learning, duan2016rl}, \maml \citep{finn2017model}, AMAGO \citep{grigsby2024amago}, TrMRL \citep{melo2022transformers} and ECET \citep{shala2025efficient}, \nom achieves the highest observed aggregate performance among the evaluated methods across both locomotion and manipulation domains. 

%% file: sections/related.tex
\section{Related Work}
\textbf{Bayesian Last Layers}. \nom is most relevant to \alpaca \citep{harrison2018meta}, which introduces the concept of performing Bayesian inference only on the output layer of neural networks. \alpaca provides an efficient, online Bayesian linear regression framework that also employs learnable basis functions. However, \alpaca, alongside its subsequent extensions \citep{harrison2024variational}, was not formulated as a BRL method. While related work such as CAMeLiD \citep{harrison2018control} employs similar controllers for policy computation, they rely on the restrictive assumption of fully known reward functions. \nom generalises \alpaca and CAMeLiD in two ways: (1) whereas \alpaca and CAMeLiD assume a \textit{known} noise in the likelihood function, \nom performs exact Bayesian inference over the model noise, (2) \nom operates seamlessly in online BRL settings with unknown rewards, whereas \alpaca and CAMeLiD focus on offline problems with simple, known rewards. As we will show in \Apref{ablation}, the assumption of known noise induces predictive errors in both transition and reward models.

\textbf{Reinforcement Learning}. RL methods are broadly categorised into model-free and model-based approaches. Here, we use model-free algorithms, specifically Proximal Policy Optimization (PPO) \citep{schulman2017proximal} and Soft Actor-Critic (SAC) \citep{haarnoja2018soft}, to learn the BRL policies, aligning with a substantial portion of the Meta-RL literature. Furthermore, as \nom learns transition and reward models, it is naturally compatible with model-based methods.

{\color{black} \textbf{Hidden-Parameter MDPs}. Hidden-Parameter MDPs (HiP-MDPs) \citep{doshi2016hidden} provide a framework for parametric Bayesian Reinforcement Learning. Initially modelled using Gaussian Processes (GPs) \citep{doshi2016hidden}, HiP-MDPs were subsequently scaled by replacing GPs with Bayesian Neural Networks (BNNs) \citep{killian2017robust}. However, updating BNN weights at test-time has proved inefficient \citep{yang2019single}. While follow-up work mitigated such inefficiency by fixing the test-time BNN weights and exclusively optimising task parameters \cite{yao2018direct}, doing so strips away the true Bayesian features of the framework. Optimising task parameters while fixing BNN weights is mathematically equivalent to performing Maximum Likelihood Estimation (MLE), diverting the agent from Bayes-optimal exploration strategies (a limitation similarly noted by \citep{zintgraf2021varibad}). In contrast, recent parametric deep BRL methods \citep{rakelly2019efficient, perez2020generalized, zintgraf2021varibad, lee2023parameterizing}, including \nom, are considered orthogonal to this line of work, as they do not assume known forms of reward functions and perform (approximate) Bayesian inference directly on task parameters rather than the neural network weights.}

\textbf{Classical Bayesian Reinforcement Learning}. Classical BRL methods assume known forms of transitions and rewards. Early approaches formulated BRL as a Partially Observable MDP (POMDP) \citep{kaelbling1998planning} utilising sampling-based offline solvers \cite{poupart2006analytic}, while later methods introduced online, tree-based solvers leveraging posterior sampling \citep{strens2000bayesian, osband2013more} for efficiency \cite{guez2013scalable}. These methods plan to find approximately optimal policies, which is orthogonal to the direct policy learning in \nom. \nom shares the idea of using generalised linear models with \cite{tziortziotis2013linear}, while it differs fundamentally in that \cite{tziortziotis2013linear} \textit{specifies} the basis functions, instead of learning them as in \nom. Hand-crafting even a simple non-linear basis function, e.g., quadratic function, yields an $O(d^2)$ dimensional feature space, where $d$ is the dimension of the raw input. As detailed in \Secref{method}, performing online Bayesian inference scales at least quadratically with respect to the feature space dimension, meaning a prohibitive $O(d^4)$ complexity is required for a single inference step. In contrast, by \textit{learning} basis functions, \nom enables the use of compact low-dimensional feature spaces that capture non-linearities while preserving scalability.

\textbf{Meta-Reinforcement Learning}. Meta-Reinforcement Learning (Meta-RL) aims to learn policies from \textit{seen} tasks that are capable of adapting to \textit{unseen} tasks following similar task distributions \citep{beck2023survey}. According to \cite{beck2023survey}, Meta-RL methods can be categorised as \textbf{(1) parameterised policy gradient (PPG)} methods (e.g., \cite{finn2017model, yoon2018bayesian, finn2018probabilistic}) that learn by performing meta-policy gradients on the meta-parameterised policy; \textbf{(2) black-box methods} (e.g., \cite{duan2016rl, wang2016learning, melo2022transformers, grigsby2024amago, shala2025efficient}) that learn from summaries of histories and \textbf{(3) task-inference methods} (e.g., \cite{rakelly2019efficient, zintgraf2021varibad, lee2023parameterizing}) that learn from the belief of the task parameters (i.e., $\thetat, \thetar$ in \nom). Most of the deep BRL methods can be viewed as task inference methods, hence a subset of Meta-RL. BRL differs from other Meta-RL methods in that it performs Bayesian inference on task parameters, affording advantages such as uncertainty quantification. As a result, BRL methods have natural integrations with model-based algorithms, holding significance to research fields such as risk-aware control and planning under uncertainty.

%% file: sections/background.tex
\section{Background}
\label{background}
\subsection{Markov Decision Processes and Reinforcement Learning}
Markov Decision Processes (MDPs) are 5-tuples defined as $\mathcal{M} = \left ( \mathcal{S}, \mathcal{A}, R, T, \gamma\right) $, where $\mathcal{S}$ is the state space, $\mathcal{A}$ is the action space, $R(s_t , a_t , s_{t+1} , r_{t+1})= p(r_{t+1}|s_t, a_t, s_{t+1})$ is the \textit{reward} function, $T(s_{t}, a_{t}, s_{t+1}) = p(s_{t+1}|s_t, a_t)$ is the \textit{transition} function, and $\gamma \in [0, 1]$ a discount factor. In the above definition, $s_t \in \mathcal{S}, s_{t+1} \in  \mathcal{S}, a_t \in \mathcal{A}$ and $r_{t+1} \in \mathbb{R}$. The goal of solving an MDP is to find a policy $\pi(a_t|s_t): \mathcal{S} \rightarrow \mathcal{A}$ that maximises the expected return over a finite horizon $H > 0$, given by $\mathcal{J}(\pi) = \mathbb{E}_{\pi}[\sum_{t=0}^{H} \gamma^tr_{t+1}]$. Standard Reinforcement Learning (RL) problems optimise this exact objective; however, they assume the transition dynamics $T$, the reward function $R$, or both are unknown and must be learnt from data. 

\subsection{Bayes-Adaptive MDPs and Bayesian Reinforcement Learning}

Bayes-Adaptive MDPs (BAMDPs) \citep{duff2002optimal, ghavamzadeh2015bayesian} parameterise transition and reward functions by unknown variables $\thetat \in \Thetat$ and $\thetar \in \Thetar$, respectively. Following POMDP terminology \citep{kaelbling1998planning}, a BAMDP maintains their joint distribution, or \textit{belief}, $b_t=p_t(\thetat,\thetar)$. After observing $(s_t,a_t,s_{t+1},r_{t+1})$, Bayesian inference updates it as $b_{t+1}=p_{t+1}(\thetat,\thetar)$.

To efficiently use existing MDP frameworks, beliefs can be absorbed into the original state space to form hyper-states $\mathcal{S}^{+}=\mathcal{S}\times\mathcal{B}(\Thetat\times\Thetar)$. Hence, BAMDPs can be defined as 5-tuple $(\mathcal{S}^{+},\mathcal{A},R^+,T^+,\gamma)$ MDPs. Let $\bar b_{t+1}=\tau_s(b_t,s_t,a_t,s_{t+1})$ be the intermediate joint belief after observing the next state. The joint one-step kernel factors as
\begin{align}
\begin{split}
    &p(s_{t+1},r_{t+1},b_{t+1}\mid s_t,b_t,a_t)\\
    &=\mathbb{E}_{(\thetat,\thetar)\sim b_t}\!\left[p(s_{t+1}\mid s_t,a_t,\thetat)\right]
      \mathbb{E}_{(\thetat,\thetar)\sim\bar b_{t+1}}\!\left[p(r_{t+1}\mid s_t,a_t,s_{t+1},\thetar)\right]\\
    &\quad\cdot\delta\!\left(b_{t+1}=\tau_r(\bar b_{t+1},s_t,a_t,s_{t+1},r_{t+1})\right). \label{hypert}
\end{split}
\end{align}
The first and second expectations are the predictive factors $T^+$ and $R^+$, respectively. Thus, observing $s_{t+1}$ produces $\bar b_{t+1}$, which predicts $r_{t+1}$; observing the reward then produces the final belief $b_{t+1}$ used at the next decision. Accordingly, the expected return to maximise becomes $\mathcal{J}^{+}(\pi^+) = \mathbb{E}_{T^+, \pi^+, R^+}[\sum_{t=0}^{H^+} \gamma^t  r_{t+1}]$, where $H^{+} > 0$ is the BAMDP horizon, and $\pi^+: \mathcal{S}^{+} \rightarrow \mathcal{A}$ is the policy of BAMDPs \footnote{We will use BAMDP policy and BRL policy interchangeably when referring to $\pi^+$.}. Problems that require solving BAMDPs are named Bayesian Reinforcement Learning (BRL). Beyond improving generalisation, BRL offers a principled mechanism for addressing the exploration-exploitation trade-off \citep{ghavamzadeh2015bayesian}.

Classical BRL methods \citep{poupart2006analytic, guez2013scalable, tziortziotis2013linear} typically assume known functional forms for $T^+$ and $R^+$, limiting flexibility. HiP-MDPs \citep{doshi2016hidden, killian2017robust, yao2018direct} instead infer neural-network weights, but scale poorly \citep{yang2019single} and retain known reward forms. Recent deep BRL methods \citep{perez2020generalized, harrison2018control, rakelly2019efficient, zintgraf2021varibad} use neural networks while performing approximate inference over task parameters; \nom follows this scalable paradigm while accommodating unknown rewards.

To contextualise our method, we formalise model learning in the  recent deep BRL setting. The agent interacts with a set of MDPs with unknown transition dynamics and/or reward functions. A single sample at timestep $t$ is denoted as $c_t:=\left(s_t, a_t, s_{t+1}, r_{t+1}\right)$. Within a given MDP $\mathcal{M}$, we define the collection of $N$ samples  as $\mathcal{C}^\mathcal{M} := \left\{ c_t\right\}_{t=1}^N$.
During training, assuming access to a batch of $M$ MDPs $\left\{\mathcal{M}_i\right\}_{i=1}^M$, the resulting joint dataset is given by $\mathcal{C} = \bigcup_{i=1}^M \mathcal{C}^{\mathcal{M}_i}$. The objective in most recent deep BRL methods \citep{rakelly2019efficient, perez2020generalized,    zintgraf2021varibad} is to maximise the marginal log-likelihood of this joint data 
{\small
\begin{align}
    \begin{split}
         \log p_{\zeta, \phit, \phir}(\mathcal{C}) = \sum_{i=1}^M \log p_{\zeta, \phit, \phir} (\mathcal{C}^{\mathcal{M}_i}) 
         = \sum_{i=1}^M \log \iint p_{\zeta}(\thetat, \thetar) p_{\phit, \phir} (\mathcal{C}^{\mathcal{M}_i} | \thetat, \thetar) d\thetat d\thetar \label{logllh}
    \end{split}
\end{align} 
}

\noindent where $\zeta$ is an optional neural network parameterising the prior $p(\thetat, \thetar)$. The likelihood of an individual task factors as $p_{\phit, \phir}(\mathcal{C}^{\mathcal{M}_i} | \thetat, \thetar) \propto \prod_{t=1}^N p_{\phit}(s_{t+1}|s_t, a_t, \thetat) p_{\phir} (r_{t+1}|s_t, a_t, s_{t+1}, \thetar)$, where $\phit$ and $\phir$ denote neural networks to learn the forms of transition and reward functions, respectively. 

To facilitate optimisation, $p_{\phit}$ and $p_{\phir}$ are typically modelled as Gaussians with means and diagonal covariances determined by the output of neural networks $\phit, \phir$. The prior $p_{\zeta}(\thetat, \thetar)$ is also assumed to be a Gaussian. However, note that even with these distributional simplifications, the exact marginal likelihood in \Eqref{logllh} remains intractable. Because the neural network introduces high non-linearity between task parameters $\thetat, \thetar$ and the data, the analytical integration is impossible.  Consequently, these methods are forced to resort to variational inference, optimising the Evidence Lower Bound (ELBO) as a tractable surrogate objective (proof see \Apref{elboproof}):  
\begin{align}
    \begin{split}
        \log p_{\zeta, \phit, \phir}(\mathcal{C}) 
        &\ge \sum_{i=1}^M { \mathbb{E}_{q}\left[\log p_{\phit, \phir}(\mathcal{C}^{\mathcal{M}_i}|\thetat, \thetar) \right] - D_{KL}\left({q(\thetat, \thetar|\mathcal{C}^{\mathcal{M}_i})} \Vert p_{\zeta}(\thetat, \thetar)\right) }  \label{elbo}
    \end{split}
\end{align}
where $D_{KL}(\cdot||\cdot)$ is the KL-divergence, and $q(\cdot)$ is an approximate Gaussian posterior of $\thetat, \thetar$.

Variational inference enables model learning, but it introduces optimisation challenges such as amortisation gaps \citep{cremer2018inference} and posterior collapse \citep{bowman2016generating, dai2020usual}. These issues frequently prevent learning meaningful and distinctive task representations. Different from other tasks where the quality of learnt representations might be less important, BRL policies rely heavily on continually updated task representations to compute actions. Indistinctive representations will substantially harm the performance of BRL policies. 

The other challenge is to compute the approximate posterior $q(\thetat, \thetar|\mathcal{C}^{\mathcal{M}_i})$ directly on variable-length $\mathcal{C}^{\mathcal{M}_i}$. The most common approach is to use sequential models like RNNs or Transformers \citep{vaswani2017attention, grigsby2024amago, melo2022transformers, shala2025efficient} to summarise $\mathcal{C}^{\mathcal{M}_i}$. However, because these models are permutation-variant, directly combining them with sample-efficient off-policy algorithms like Soft Actor-Critic (SAC) \citep{haarnoja2018soft} causes a severe performance drop, as demonstrated by an ablation study in \pearl \cite{rakelly2019efficient}. The other approach, used by \pearl, is a factored approximation $q(\thetat, \thetar|\mathcal{C}^{\mathcal{M}_i}) \approx \prod_{t=1}^N \mathcal{N}(\thetat, \thetar|g(\mathcal{C}^{\mathcal{M}_i}[t]))$, where $g(\cdot)$ is a neural network that takes the $t$-th sample in $\mathcal{C}^{\mathcal{M}_i}$ as the input and returns the mean and covariance of a Gaussian as the output. However, Bayes' rule reveals that $g(\cdot)$ must essentially predict 
$p(\thetat, \thetar)^{(1/N)-1}q(\thetat, \thetar|\mathcal{C}^{\mathcal{M}_i}[t])$. Because this target relies heavily on the length $N$, it creates shifting optimisation targets as the dataset grows, resulting in inaccurate posterior approximations. In contrast, \nom completely circumvents the need to compute an approximate posterior in the first place; its exact Bayesian update is naturally permutation-invariant, enabling seamless integration with SAC, as we will demonstrate in \Secref{exp}.

%% file: sections/method.tex
\section{Meta-RL with Bayesian Linear Task Models}
\label{method}

\begin{figure*}[h]
  \centering
  \includegraphics[width=\textwidth]{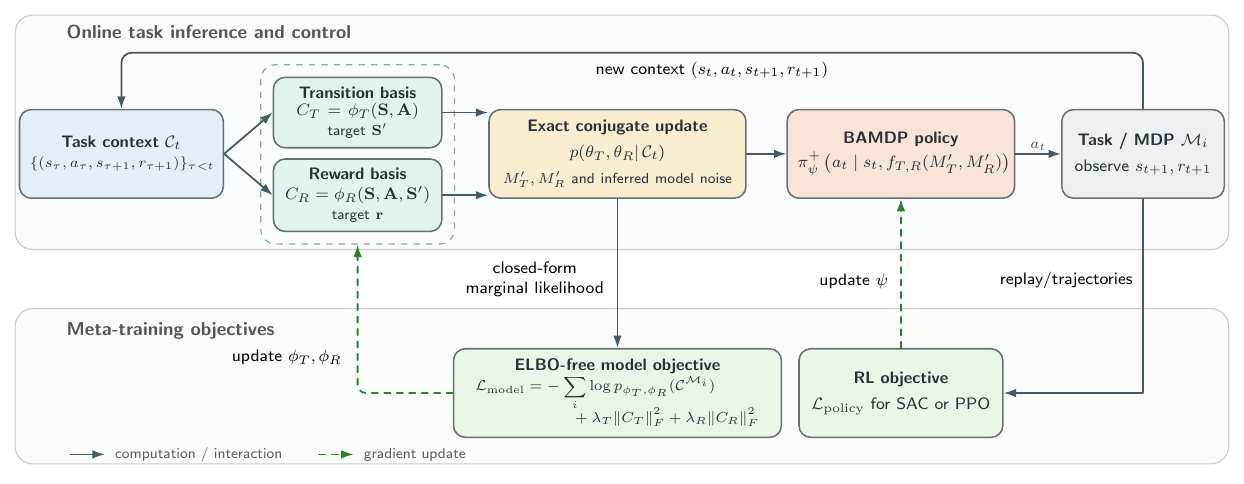}
  \caption{
    Overview of \nom. A task context is mapped through learned transition and reward basis functions to exact Bayesian posteriors. The posterior means form a task representation supplied to a BAMDP policy. During meta-training, the basis functions are optimised using the closed-form marginal-likelihood objective, while the policy is optimised using the corresponding reinforcement-learning objective. At test time, adaptation consists only of sequential posterior updates. \label{diagram}
  }
\end{figure*}

\nom is an online algorithm whose training alternates between data collection and learning. During data collection, a batch of MDPs $\{\mathcal{M}_{i}\}_{i=1}^K$ is sampled from the task distribution. Samples $\left\{\mathcal{C}^{\mathcal{M}_i}\right\}_{i=1}^K$ are maintained for each MDP separately. Starting from state $s$, to collect samples $(s, a, s', r)$ to be stored in $\mathcal{C}^{\mathcal{M}_i}$, \nom sequentially performs the following at each timestep: (1)  updates the current belief $b$  to the exact posterior distribution $p_{\phit, \phir}({\thetat}, {\thetar} | \mathcal{C}^{\mathcal{M}_i})$ (2) samples an action $a$ from the BAMDP policy $\pi_{\psi}^{+} (a|s, b)$ where $\psi$ is a neural network, and (3) executes $a$ to observe the next state $s'$ and reward $r$. Once sufficient samples are collected across tasks, \nom samples a batch of samples $\mathcal{D}$ from the joint buffer $\mathcal{C} = \bigcup_i \mathcal{C}^{\mathcal{M}_i}$ and updates the BAMDP policy network $\psi$ and transition/reward models $\phit, \phir$ via gradient descent on $\mathcal{L}_{\text{policy}}$ and $\mathcal{L}_{\text{model}}$, respectively. The complete procedure is detailed in \Algref{glibrlalg}, and we include a diagram of \nom in \Figref{diagram}.

The success of \nom relies on solving three primary challenges: (1) formulating a tractable, sequentially updated posterior $p_{\phit, \phir}({\thetat}, {\thetar} | \mathcal{C}^{\mathcal{M}_i})$, (2) deriving an ELBO-free objective $\mathcal{L}_{\text{model}}$ for updating $\phit, \phir$, and (3) constructing theoretically grounded task representations for learning the policy $\pi_{\psi}^+$. We will detail these solutions in \Secref{mlearning1},  \Secref{mlearning2}, and  \Secref{plearning}, respectively.

\subsection{Tractable and Sequentially Updated Posterior Distributions}
\label{mlearning1}
Inspired by \alpaca \citep{harrison2018meta}, to derive the exact posterior $p_{\phit, \phir}({\thetat}, {\thetar} | \mathcal{C}^{\mathcal{M}_i})$, we must explicitly define the prior $p(\thetat, \thetar)$ and the likelihood $p_{\phit, \phir}(\mathcal{C}^{\mathcal{M}_i} | \thetat, \thetar)$. Let the task data be denoted compactly as $\mathcal{C}^{\mathcal{M}_i} = (\mathbf{S}_i \in \mathbb{R}^{N \times D_S}, \mathbf{A}_i \in \mathbb{R}^{N \times D_A} , \mathbf{S}_i'  \in \mathbb{R}^{N \times D_S}, \mathbf{r}_i  \in \mathbb{R}^{N \times 1})$, where $N$ is the number of samples and $D_S, D_A$ are the dimensions of the state and action space. Let the task variables be defined as $\thetat = (\mut \in \mathbb{R}^{D_T \times D_S}, \sigmat \in \mathbb{R}^{D_S \times D_S})$ and $\thetar = (\mur \in \mathbb{R}^{D_R \times 1}, \sigmar \in \mathbb{R}^{1 \times 1})$, where $D_T, D_R$ are task dimensions. We formulate the prior as Normal-Wishart distributions:
{
\small
\begin{align}
    \begin{split}
        &p(\thetat, \thetar) = p(\thetat) \cdot p(\thetar) \qquad \text{(Assuming independent $\thetat, \thetar$)}
    \end{split}\\
\labelAndRemember{tprior}{
    \begin{split}
         &p(\thetat)  = \mathcal{MN}(\mut|\Mt \in \mathbb{R}^{D_T \times D_S}, \xit^{-1} \in \mathbb{R}^{D_T \times D_T}, \sigmat) \cdot \mathcal{W}(\sigmat^{-1}|\omegat^{-1} \in \mathbb{R}^{D_S \times D_S}, \nut) 
    \end{split}}
   \\
   \begin{split}
        &p(\thetar)  =\mathcal{MN}(\mur|\Mr\in \mathbb{R}^{D_R \times 1}, \xir^{-1}\in \mathbb{R}^{D_R \times D_R}, \sigmar) \cdot \mathcal{W}(\sigmar^{-1}|\omegar^{-1} \in \mathbb{R}^{1 \times 1}, \nur)
   \end{split}
\end{align}
}

Here, $\mathcal{W}(\mathbf{W}|\mathbf{X}, \nu)$ defines a Wishart distribution on positive definite random matrix $\mathbf{W}$ with scale $\mathbf{X}$ and degrees of freedom $\nu$, and $\mathcal{MN}(\mathbf{W|\mathbf{X}, \mathbf{Y}, \mathbf{Z}})$ defines a matrix normal distribution with random matrix $\mathbf{W}$, mean $\mathbf{X}$, row covariance $\mathbf{Y}$ and column covariance $\mathbf{Z}$. Normal-Wishart priors enable fully tractable Bayesian inference over the model noises $\sigmat, \sigmar$, generalising previous work \cite{harrison2018control, harrison2018meta}. We then define the Matrix-Normal likelihood function:
\begin{align}
    \begin{split}
    &p_{\phit, \phir}(\mathcal{C}^{\mathcal{M}_i} | \thetat, \thetar) = p_{\phit}(\mathbf{S}_i'|\mathbf{S}_i, \mathbf{A}_i, \thetat) \cdot p_{\phir}(\mathbf{r}_i|\mathbf{S}_i, \mathbf{A}_i, \mathbf{S}_i', \thetar) \cdot p(\mathbf{S}_i, \mathbf{A}_i)
    \end{split}\\
    \labelAndRemember{tllh}{
    &p_{\phit}(\mathbf{S}_i'|\mathbf{S}_i, \mathbf{A}_i, \thetat) = \mathcal{MN}\left(\mathbf{S}_i'|\Ct \mut, \mathbf{I}_N,  \sigmat \right) }\\
    &p_{\phir}(\mathbf{r}_i|\mathbf{S}_i, \mathbf{A}_i, \mathbf{S}_i', \thetar) = \mathcal{MN}\left(\mathbf{r}_i|\Cr \mur, \mathbf{I}_N, \sigmar\right)\\
    &\Ct \in \mathbb{R}^{N \times D_T} = \phit(\mathbf{S}_i, \mathbf{A}_i) \qquad
    \Cr \in \mathbb{R}^{N \times D_R} = \phir(\mathbf{S}_i, \mathbf{A}_i, \mathbf{S}_i')
\end{align}
The term $p(\mathbf{S}_i, \mathbf{A}_i)$ acts as a constant irrelevant to the optimisation, while $\Ct$ and $\Cr$ \footnote{Dependence on $i$ of $\Ct, \Cr$ is omitted for conciseness. This also applies to $\Mt', \xit', \omegat', \Mr', \xir', \omegar'$.} represent data features generated by neural networks $\phit, \phir$ acting as expressive and learnable basis functions. By restricting the linearisation exclusively to the mapping between the feature space and the Bayesian parameters, we preserve exact mathematical tractability.

Because the Normal-Wishart prior is conjugate to the Matrix-Normal likelihood, the resulting posteriors also follow Normal-Wishart distributions (proof in \Apref{nwconjugacy})
\begin{align}
    p_{\phit, \phir}({\thetat}, {\thetar} | \mathcal{C}^{\mathcal{M}_i}) &= p_{\phit}(\thetat|\mathbf{S}_i, \mathbf{A}_i, \mathbf{S}_i') \cdot p_{\phir}(\thetar|\mathbf{S}_i, \mathbf{A}_i, \mathbf{S}_i', \mathbf{r}_i)\\
    \labelAndRemember{tpost}{p_{\phit}(\thetat|\mathbf{S}_i, \mathbf{A}_i, \mathbf{S}_i') &= \mathcal{MN}(\mut|\Mt', \xit'^{-1}, \sigmat) \cdot \mathcal{W}(\sigmat^{-1}|\omegat'^{-1}, \nut')} \\
    p_{\phir}(\thetar|\mathbf{S}_i, \mathbf{A}_i, \mathbf{S}_i', \mathbf{r}_i) &= \mathcal{MN}(\mur|\Mr', \xir'^{-1}, \sigmar) \cdot \mathcal{W}(\sigmar^{-1}|\omegar'^{-1}, \nur')
\end{align}
where
\begin{align}
    \labelAndRemember{post}{
    \begin{split}
         \Mt' &= {\xit'}^{-1} \left[\Ct^\text{T}\mathbf{S}' + \xit \Mt \right]\\
        {\xit'} &=\Ct^\text{T}\Ct + \xit \\
        {\omegat'} &= {\omegat} + {\mathbf{S}'}^\text{T} \mathbf{S}'\\
                &\quad+ \Mt^\text{T}\xit \Mt -  {\Mt'}^\text{T} \xit' {\Mt'}\\
       \nut' &= \nut + N 
   \end{split}
   }
    \begin{split}
        \Mr' &= {\xir'}^{-1} \left[\Cr^\text{T}\mathbf{r} +  \xir \Mr \right]\\
        {\xir'} &= \Cr^\text{T}\Cr + \xir\\
        {\omegar'} &= \omegar + \mathbf{r}^\text{T}  \mathbf{r}  \\
        &\quad+ \Mr^\text{T}  \xir \Mr -  {\Mr'}^\text{T} \xir ' {\Mr'}\\
        \nur' &= \nur + N 
    \end{split}
\end{align}
During online execution, the agent must sequentially update priors to posteriors with each newly observed sample $c_t = \{s_t, a_t, s_t', r_{t+1}\}$ for arbitrary timestep $t$. Naively computing \Eqref{post} requires inversion of $\xit'$ and $\xir'$, which scale at $\mathcal{O}(D_T^3)$ and $\mathcal{O}
(D_R^3)$, respectively. Applying the matrix inversion lemma yields
\begin{align}
    \begin{split}
        \xit'^{-1} &= \left(\Ct^{\text{T}}\Ct + \xit\right)^{-1} = \xit^{-1} - \xit^{-1} \Ct^{\text{T}} \left( \mathbf{I}_{N} + \Ct \xit^{-1} \Ct^{\text{T}}\right)^{-1} \Ct \xit^{-1}
    \end{split} \\
    \begin{split}
        \xir'^{-1} &= \left(\Cr^{\text{T}}\Cr + \xir\right)^{-1} = \xir^{-1} - \xir^{-1} \Cr^{\text{T}} \left( \mathbf{I}_{N} + \Cr \xir^{-1} \Cr^{\text{T}}\right)^{-1} \Cr \xir^{-1}
    \end{split}
\end{align}
With a single sample, $\Ct \in \mathbb{R}^{1 \times D_T}$ and $\Cr \in \mathbb{R}^{1 \times D_R}$. Hence, $\left( \mathbf{I}_{N} + \Ct \xit^{-1} \Ct^{\text{T}}\right)^{-1}$ and $\left( \mathbf{I}_{N} + \Cr \xir^{-1} \Cr^{\text{T}}\right)^{-1}$ are reduced to reciprocals of scalars. By caching the inverted matrices, this reduces the entire online sequential update complexity to $\mathcal{O}(\max\left(D_S^2, D_T^2\right))$ for transitions and $\mathcal{O}(D_R^2)$ for rewards, enabling efficient data collection.
 
\subsection{ELBO-free Model Updates}
\label{mlearning2}
The fully tractable posterior update yields a closed-form marginal likelihood, which contributes to the model loss $\mathcal{L}_\text{model}$ for optimising model parameters $\phit, \phir$. Dropping the optional neural network $\zeta$ , \Eqref{logllh} can be written analytically as
\begin{equation}
\begin{split}
    \log p_{\phit, \phir}(\mathcal{C}) &= \sum_{i=1}^M \log \int p(\thetat) p_{\phit}(\mathbf{S}_i'|\mathbf{S}_i, \mathbf{A}_i, \thetat)  d\thetat \\
    &+\sum_{i=1}^M \log \int p(\thetar) p_{\phir}(\mathbf{r}_i|\mathbf{S}_i, \mathbf{A}_i, \mathbf{S}_i', \thetar) d\thetar + \text{const.} \label{logllhnew}
\end{split}
\end{equation}
Substituting accordingly with equations in \Secref{mlearning1}, we reduce the integral to a closed-form marginal log-likelihood (proof in \Apref{llhproof}):
\begin{align}
    \labelAndRemember{llhour}{
    \log p_{\phit, \phir}(\mathcal{C}) &= -\frac{1}{2}\sum_{i=1}^M \left(D_S\log |\xit'| +\nut' \log |\frac{1}{2} \omegat'|   +  \log |\xir'| + \nur' \log |\frac{1}{2} \omegar'|\right) +  \text{const.}
    }
\end{align}
To optimise $\phit$ and $\phir$, \Eqref{llhour} is maximised with respect to features $\Ct$ and $\Cr$. To prevent overfitting, we apply squared Frobenius norms $\Vert \Ct \Vert_F^2$ and $\Vert \Cr \Vert_F^2$ to \Eqref{llhour} as regularisations, the effect of which is discussed in \Apref{ablation}. The regularised loss is defined as
\begin{align}
    \mathcal{L}_{\text{model}} := -\log p_{\phit, \phir}(\mathcal{C}) + \lambda_{T} \Vert \Ct \Vert_F^2 +  \lambda_{R} \Vert \Cr \Vert_F^2   \label{llhourfinal}
\end{align}
where $\lambda_{T} > 0$ and $\lambda_{R} > 0$ are hyperparameters. Note that $\mathcal{L}_{\text{model}}$ can be directly minimised with gradient descent, without any variational inference and the evaluation of the ELBO term. 

\subsection{Task Representation in Policy Networks}
\label{plearning}
The final component in \nom involves integrating beliefs, which define tasks, into the policy network $\pi_{\psi}^{+}(a|s, b)$ to drive decision-making and sample collection. We concisely represent this belief using the deterministic means of the task posteriors:
\begin{align}
    \pi_{\psi}^{+}(a|s, b) \approx \pi_{\psi}^{+}\left(a|s, f\left(\Mt', \Mr'\right)\right)
\end{align}
where $f: \mathbb{R}^{D_T \times D_S} \times \mathbb{R}^{D_R \times 1} \rightarrow \mathbb{R}^{D_P \times 1}$ and $D_P$ is the dimension of the belief representation. 

For $\mathbf{x}\in\mathbb{R}^d$ and a fixed $\epsilon>0$, define the $\epsilon$-normalisation
\begin{align}
    \mathcal{N}_\epsilon(\mathbf{x})=\mathbf{x} / \left(\lVert \mathbf{x}\rVert_2+\epsilon\right) \label{zerosafenorm}
\end{align}
for normalisation to be well-defined at the empty context.  $\mathcal{N}_\epsilon$ agrees with ordinary normalisation up to the factor $\rho(x)=\lVert x\rVert_2/(\lVert x\rVert_2+\epsilon)\in[0,1)$, which is close to one whenever $\lVert x\rVert_2\gg\epsilon$. We focus on the following concatenated representation \footnote{We list alternatives and empirical comparisons in \Apref{beliefrep}.}
\begin{align}
    \label{usedrep}
    f_{T, R}\left(\Mt', \Mr'\right)
    =\left[\mathcal{N}_\epsilon\!\left(\mathrm{tril}\left(\Mt'{\Mt'}^{\mathrm{T}}\right)\right)^{\mathrm{T}}\;\;
    \mathcal{N}_\epsilon\!\left(\Mr'\right)^{\mathrm{T}}\right]^{\mathrm{T}}
\end{align}
Here, $\mathrm{tril}(\cdot)$ retrieves the lower triangle of a matrix and flattens it into a vector. Let samples from two MDPs $\mathcal{M}_1$, $\mathcal{M}_2$ be $\mathcal{C}^{\mathcal{M}_1}$ and $\mathcal{C}^{\mathcal{M}_2}$, where $
    \mathcal{C}^{\mathcal{M}_1} =\left(\mathbf{S}_1, \mathbf{A}_1, \mathbf{S}'_1, \mathbf{r}_1\right), \mathcal{C}^{\mathcal{M}_2} =\left(\mathbf{S}_2, \mathbf{A}_2, \mathbf{S}'_2, \mathbf{r}_2\right)$. 
Given posterior means ${\Mt^1}'$, ${\Mr^1}'$, ${\Mt^2}'$, ${\Mr^2}'$ and zero prior means, \Eqref{post} gives
\begin{align}
    \labelAndRemember{finalrep}{
            &\left\Vert f_{T, R}\left({\Mt^1}', {\Mr^1}'\right) - f_{T, R}\left({\Mt^2}', {\Mr^2}'\right) \right\Vert_2^2 \nonumber \\ &\hspace{5em} =\sum_{k\in\{k_{TT},\,k_R\}}\Big(\rho_k\big(\mathcal{C}^{\mathcal{M}_1}\big)^2+\rho_k\big(\mathcal{C}^{\mathcal{M}_2}\big)^2-2\,\mathrm{sCMS}_{k}\big(\mathcal{C}^{\mathcal{M}_1},\mathcal{C}^{\mathcal{M}_2}\big)\Big)
    }
\end{align}
where $k_{TT}$ and $k_R$ are kernels induced by the learned basis functions $\phit, \, \phir$ on pairs of transition samples and on reward samples respectively, $\mathrm{sCMS}_k$ is the signed-measure Cosine Mean Similarity \citep{gruber2024disentangling} between task datasets under kernel $k$, and $\rho_k(\mathcal{C})^2=\mathrm{sCMS}_k(\mathcal{C},\mathcal{C})\in[0,1)$.

The unnormalised version $\widetilde{f_{T,R}} = \left[\mathrm{tril}\left(\Mt'{\Mt'}^{\mathrm{T}}\right)^{\mathrm{T}}\;\;\Mr'^{\mathrm{T}}\right]^{\mathrm{T}}$ features a cleaner identity:
\begin{align}
\labelAndRemember{finalrepunnorm}{
    \left\Vert \widetilde{f_{T, R}}\left({\Mt^1}', {\Mr^1}'\right) - \widetilde{f_{T, R}}\left({\Mt^2}', {\Mr^2}'\right) \right\Vert_2^2 = \mathrm{sMMD}_{k_{TT}}^2\big(\mathcal{C}^{\mathcal{M}_1},\mathcal{C}^{\mathcal{M}_2}\big)+\mathrm{sMMD}_{k_R}^2\big(\mathcal{C}^{\mathcal{M}_1},\mathcal{C}^{\mathcal{M}_2}\big)
    }
\end{align}
where $\mathrm{sMMD}_k$ is the signed-measure Maximum Mean Discrepancy \citep{gretton2012kernel} between the two task datasets under kernel $k$.  Detailed definitions and proofs are deferred to \Apref{theory}. 

\subsection{Diagonal Noise Inference}
\label{diagonalnoise}
The Matrix-Normal-Wishart formulation derived above learns a full covariance and therefore captures cross-dimensions correlations. For cases where only diagonal noise inference suffices, we provide a Matrix-Normal-Gamma formulation. We assign an independent Gamma prior to each output dimension, preserving the exact tractability. Its closed-form update, online execution complexity, ELBO-free marginal log-likelihood and task-representation result are discussed in \Apref{mngconjugacy}.

%% file: sections/experiment.tex
\section{Experiments}
\label{exp}
We evaluate \nom on locomotion tasks from the MuJoCo benchmark \citep{todorov2012mujoco} and manipulation tasks from the MetaWorld \citep{yu2020meta, mclean2025meta} benchmark. For MuJoCo, following standard Meta-RL evaluation protocols, we assess performance on \texttt{HalfCheetahDir}, \texttt{AntDir}, and \texttt{HalfCheetahVel}. In these environments, the agent must adapt its policy from the observed reward  to run in the randomly determined forward/backward direction or at a targeted velocity. For MetaWorld, which is considered one of the most challenging meta-RL benchmarks, we focus on its most complex subsets: Meta-Learning 10 (ML10) and Meta-Learning 45 (ML45). ML10 and ML45 consist of 10 and 45 training tasks, respectively, alongside 5 held-out testing tasks where the agent must adapt to both unseen transition and reward functions. To ensure fair comparison, we use the V2 reward in MetaWorld-V3 \citep{mclean2025meta}, matching the reward version of all comparators.

We benchmark \nom against a diverse set of Meta-RL baselines spanning all three main categories: (1) PPG-based methods: \textbf{ MAML \citep{finn2017model}}, (2) black-box methods: \textbf{\rltwo \citep{duan2016rl}, AMAGO \citep{grigsby2024amago}, TrMRL \citep{melo2022transformers}} and \textbf{ECET \citep{shala2025efficient}}, and (3) task-inference (deep BRL) methods: \textbf{\pearl \citep{rakelly2019efficient}, \vbad \citep{zintgraf2021varibad}} and \textbf{SDVT \citep{lee2023parameterizing}}. Specifically, we compare with \rltwo, TrMRL, \pearl, \vbad in the locomotion tasks in MuJoCo, and \maml, \rltwo, \vbad, AMAGO, SDVT, TrMRL and ECET in the manipulation tasks in MetaWorld. \footnote{Methods excluded from these comparisons either lack reported results or demonstrate uncompetitive performance. For example, \maml has return $<0$ with $1e7$ steps in \texttt{HalfCheetahDir} \citep{melo2022transformers}. \pearl has $<3\%$ success rate in ML10 with $1e8$ steps \citep{yu2020meta}. We exclude ECET in MuJoCo for it not being open-source and reporting MuJoCo results in \href{https://openreview.net/forum?id=UENQuayzr1}{meta-steps}.} Full implementation details are provided in \Apref{impdetails}.

Across all experiments, we focus on zero-shot test-time performance, which is critical in Meta-RL settings that require adaptation with minimal data \citep{beck2023survey}. Here, zero-shot means that the agent receives no task-specific observations and performs no adaptation before evaluation begins. During evaluation, it updates its posterior online using only transitions collected through interaction with the test environment; no gradient updates are performed. For MuJoCo experiments (\textbf{\Secref{mujocores}}), we follow the setting in \pearl and test \nom with SAC \citep{haarnoja2018soft} (\Algref{alg:glibrl_sac}) over five random seeds. Both \pearl and \nom are trained with $1e6$ total steps, while others are trained with $1e7$ total steps. For MetaWorld experiments (\textbf{\Secref{metares}}), we follow \cite{mclean2025meta} and run \nom with PPO \cite{schulman2017proximal} (\Algref{alg:glibrl_ppo}) over ten random seeds. ML10 and ML45 experiments are run for $2e7$ steps and $5e7$ steps, respectively. Beyond quantitative evaluations, we qualitatively inspect the learnt task representations (\textbf{\Secref{qualitativeres}}). Finally, using ML10, we conduct ablation studies to empirically validate the necessity of inference on model noises (\textbf{\Secref{ablationstudy}}). Each experiment is run on a single L40S GPU.

\begin{figure}[t]
    \centering
    \includegraphics[width=\textwidth]{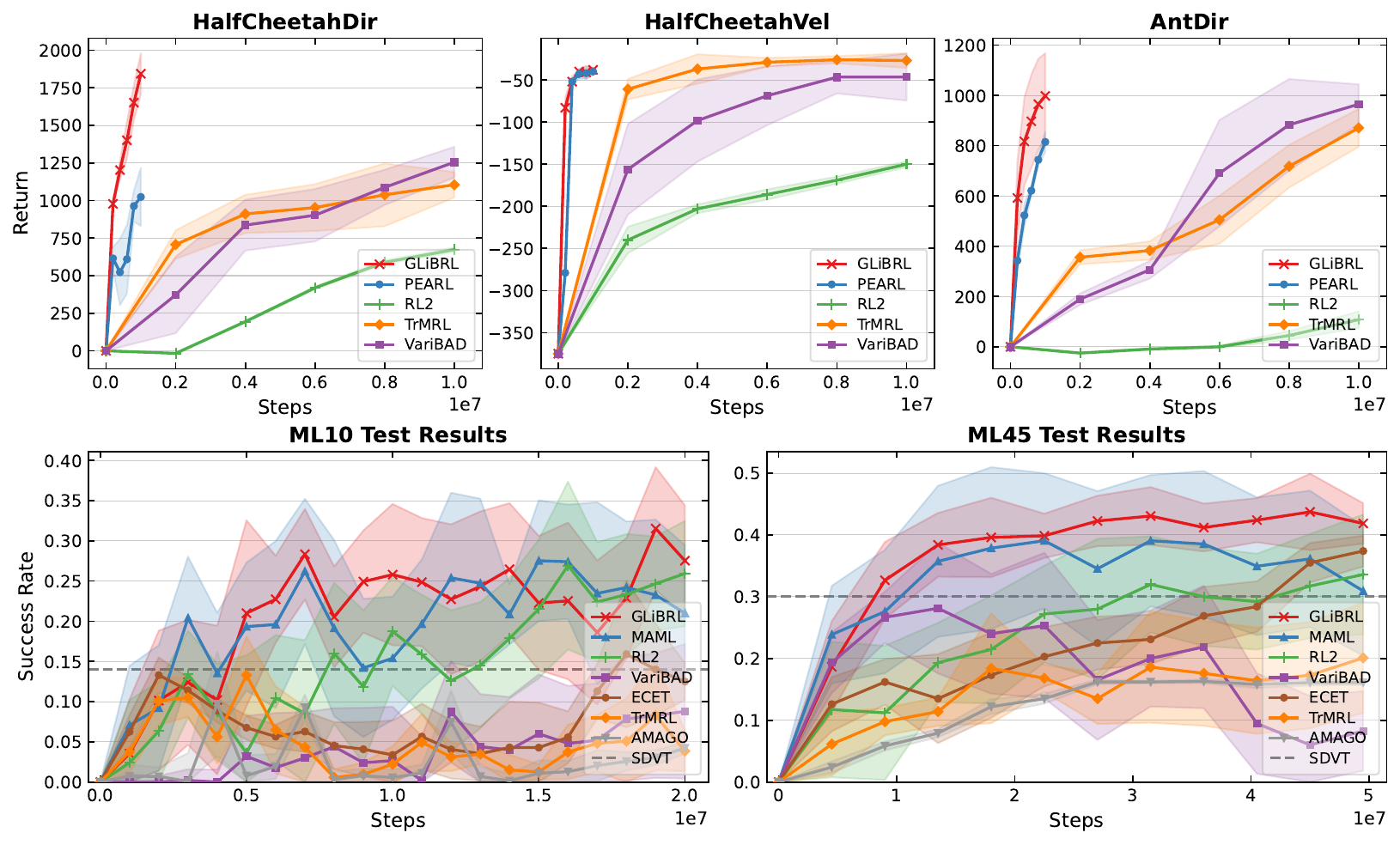}
    \caption{Test-time return and success rate in all benchmarks. Curves report the interquartile mean (IQM) \citep{agarwal2021deep}, and shaded regions denote $95\%$ confidence intervals . In \nom, $D_T=16$ and $D_R=256$ for MetaWorld tasks, and $D_R=8$ for MuJoCo tasks which do not require inference in transition. In MuJoCo tasks, we run \pearl and \nom with SAC for $1e6$ steps and other baselines with PPO for $1e7$ steps. \label{allres}}
\end{figure}

\subsection{MuJoCo Results}
\label{mujocores}
As shown in the top row in \Figref{allres}, \nom combined with SAC demonstrates superior sample efficiency on MuJoCo locomotion tasks. Using only $1e6$ training steps, \nom outperforms or matches all PPO-based baselines (\rltwo, TrMRL, \vbad) by up to $1.5\times$, achieving this with merely $10\%$ of their step budget. Crucially, as discussed in \Secref{background}, none of these PPO-based baselines can be trivially integrated with SAC, as they lack permutation invariance critical for off-policy methods \cite{rakelly2019efficient}. Furthermore, \nom outperforms the strong SAC-based baseline \pearl by up to $1.8\times$, highlighting the advantage of \nom's exact Bayesian task inference.

\subsection{MetaWorld Results}
\label{metares}
The bottom row of \Figref{allres} shows that \nom, when integrated with PPO, achieves the highest observed aggregate success rate among the evaluated methods on both ML10 and ML45. This superior aggregate result, however, does not imply uniform task-level dominance: on ML10, \maml attains a slightly better mean per-task rank over training than \nom ($2.25$ versus $2.50$). On ML45, \nom has the best mean per-task rank ($2.32$). Detailed per-task success rates and ranks are provided in \Apref{ml10pertask} and \Apref{ml45pertask}~\footnote{SDVT per-task results are excluded as they were not reported at $2e7/5e7$ steps in \cite{lee2023parameterizing}.}.

\subsection{Qualitative Analysis of Task Identifiability}
\label{qualitativeres}
We qualitatively examine the geometry of the learnt task representations on \texttt{HalfCheetahVel} using 2D Principal Component Analysis (PCA). As detailed in \Apref{taskid}, after $5e5$ training steps, \nom produces a visibly ordered representation of the continuous target velocities across five runs. In comparison, TrMRL shows less apparent separation despite a $10\times$ larger training budget. These visualisations qualitatively illustrate that the representation geometry can reflect task similarity, consistent with the exact identity in \Secref{plearning}.

\subsection{Ablation Studies}
\label{ablationstudy}
We have conducted ablation studies demonstrating (1) different choices of task representations mentioned in \Secref{plearning} (2) the effect of the regularisation mentioned in \Secref{mlearning1}, and (3) performing explicit Bayesian inference over model noises mitigates the predictive errors in both transition and reward models. Full experimental results and mathematical derivations are deferred to \Apref{beliefrep} and \Apref{ablation}.

%% file: sections/conclusion.tex
\section{Conclusion}
We introduced \nom, a novel deep BRL framework that enables fully tractable inference over task parameters and efficient learning of basis functions via exact, ELBO-free marginal likelihood optimisation. By performing Bayesian inference over the model noise, \nom reduces predictive errors. Furthermore, \nom naturally leads to permutation invariance, allowing seamless integration with both on- and off-policy RL algorithms. \nom also establishes an exact kernel identity that relates distances between its learned task representations to signed-measure kernel discrepancies over the task contexts, making the representation geometry analytically traceable. Empirically, \nom achieves the highest observed aggregate performance among the evaluated methods across the MuJoCo locomotion and MetaWorld ML10/45 benchmarks.

Several promising directions for future work arise naturally. Because \nom learns transition and reward models, it is well-suited for integration with model-based RL algorithms, which could further improve sample efficiency. Alternatively, within the model-free paradigm, it is interesting to explore how to incorporate uncertainties within task representations. Empirically, we have observed that directly feeding the covariances in the policy network will introduce high training instability.

%% file: sections/appendix.tex
\newpage
\section{Appendix}
\subsection{Table of Important Notations}
\begin{table}[h]
  \centering
  \label{tab:notation}
  \small
  \renewcommand{\arraystretch}{1.3}
  \begin{tabular}{@{}p{0.26\linewidth}p{0.69\linewidth}@{}}
    \toprule
    \textbf{Symbol} & \textbf{Meaning} \\
    \midrule
    \multicolumn{2}{@{}l}{\emph{MDPs and BAMDPs}}\\
    $\mathcal{M}=(S,A,R,T,\gamma)$            & MDP with state space $S$, action space $A$, reward $R$, transition $T$, discount $\gamma$. \\
    $S^{+}=S\times \mathcal{B}_{T}\times \mathcal{B}_{R}$         & BAMDP hyper-state space; $B_{T},B_{R}$ are the belief spaces of $\theta_{T},\theta_{R}$. \\
    $b_{t}=p(\theta_{T},\theta_{R})$          & Belief at time $t$ over task parameters. \\
    $\tau$                                    & Belief-update function, $b_{t+1}=\tau(b_{t},s_{t},a_{t},s_{t+1},r_{t+1})$. \\
    $\pi^{+}_{\psi}(a\mid s,b)$               & BAMDP (BRL) policy parameterised by $\psi$. \\
    $H^{+}$, $\mathcal{J}^{+}(\pi^{+})$                 & BAMDP horizon and expected return. \\
    \midrule
    \multicolumn{2}{@{}l}{\emph{Datasets and dimensions}}\\
    $c_{t}=(s_{t},a_{t},s_{t+1},r_{t+1})$     & Single sample at timestep $t$. \\
    $\mathcal{C}^{\mathcal{M}}=\{c_{t}\}_{t=1}^{N}$     & $N$ samples collected within MDP $\mathcal{M}$. \\
    $\mathcal{C}=\bigcup_{i=1}^{M} \mathcal{C}^{\mathcal{M}_{i}}$ & Joint dataset over a batch of $M$ MDPs. \\
    $\mathcal{C}^{\mathcal{M}_{i}}=(\mathbf{S}_{i},\mathbf{A}_{i},\mathbf{S}'_{i},\mathbf{r}_{i})$ & Samples in compact matrix forms. \\
    $D_{S},\,D_{A}$                           & State and action dimensions. \\
    $D_{T},\,D_{R}$                           & Transition and reward task dimensions. \\
    $D_{P}$                                   & Task-representation dimension. \\
    \midrule
    \multicolumn{2}{@{}l}{\emph{Task parameters}}\\
    $\theta_{T}=(\mut, \sigmat)$         & Transition task parameter: mean $\mut$, noise covariance $\sigmat$. \\
    $\theta_{R}=(\mur,\sigmar)$         & Reward task parameter: mean $\mur$, noise covariance $\sigmar$. \\
    \midrule
    \multicolumn{2}{@{}l}{\emph{Learnable basis functions}}\\
    $\phi_{T},\,\phi_{R}$                     & Neural networks acting as basis functions for transition and reward. \\
    $\mathbf{C}_{T}=\phi_{T}(S_{i},A_{i})$ & Transition features (design matrix). \\
    $\mathbf{C}_{R}=\phi_{R}(S_{i},A_{i},S'_{i})$ & Reward features (design matrix). \\
    $\zeta$                                   & Optional network parameterising the prior $p_{\zeta}(\theta_{T},\theta_{R})$. \\
    \midrule
    \multicolumn{2}{@{}l}{\emph{Distributions}}\\
    $\mathcal{MN}(\mathbf{W}\mid \mathbf{X},\mathbf{Y},\mathbf{Z})$               & Matrix-normal: mean $\mathbf{X}$, row covariance $\mathbf{Y}$, column covariance $\mathbf{Z}$. \\
    $\mathcal{W}(\mathbf{W}\mid \mathbf{X},\nu)$                & Wishart: scale matrix $\mathbf{X}$, degrees of freedom $\nu$. \\
    $\operatorname{Gam}(\cdot\mid a,b)$       & Gamma: shape $a$ and rate $b$. \\
    \midrule
    \multicolumn{2}{@{}l}{\emph{Prior/posterior parameters}}\\
    $\Mt$/$\Mt'$  & Prior/posterior mean of $T_{\mu}$. \\
    $\xit^{-1}$/${\xit'}^{-1}$ & Prior/posterior row covariance of $T_{\mu}$ ($\xit$ is the row precision). \\
    $\omegat$/$\omegat'$, $\nut$/$\nut'$ & Prior/posterior Wishart scale and dof on $T_{\sigma}^{-1}$. \\
    $(a_{T,d},b_{T,d})$ / $(a_{T,d}',b_{T,d}')$ & Prior/posterior Gamma shape and rate for transition coordinate $d$ under diagonal noise. \\
            & Parameters for $\theta_{R}$ follow analogously. \\
    \midrule
    \multicolumn{2}{@{}l}{\emph{Task representations and kernel correspondence}}\\
    {$\mathcal{N}_\epsilon$}                   & {$\epsilon$-normalisation $x/(\lVert x\rVert_2+\epsilon)$ (\Eqref{zerosafenorm}).} \\
    {$k_T,\,k_{TT},\,k_R$}                & {Kernels induced by the learned basis functions (\Apref{theory}).} \\
    {$\mathrm{sCMS}_{k}(\mathcal{C}^{\mathcal{M}_1},\mathcal{C}^{\mathcal{M}_2})$} & {Signed-measure Cosine Mean Similarity under kernel $k$ (\Apref{theory}).} \\
    {$\mathrm{sMMD}_{k}^2(\mathcal{C}^{\mathcal{M}_1},\mathcal{C}^{\mathcal{M}_2})$} & {Squared signed-measure Maximum Mean Discrepancy under kernel $k$.} \\
    \bottomrule
  \end{tabular}
\end{table}
\newpage

\subsection{\nom Algorithm}
\vspace{-1em}
\begin{algorithm}[h]
    \caption{\nom \label{glibrlalg}}
    {For $\mathcal{C}=(\mathbf{S},\mathbf{A},\mathbf{S}',\mathbf{r})$, define $\operatorname{Post}(\mathcal{C}) := p_{\phit}(\thetat\mid\phit(\mathbf{S},\mathbf{A}),\mathbf{S}')\,p_{\phir}(\thetar\mid\phir(\mathbf{S},\mathbf{A},\mathbf{S}'),\mathbf{r})$}\;
    Initialise: policy $\pi^+_{\psi}$, horizon $H$, $\phit, \phir$\;
    \While {Training}{
        Sample $K$ MDPs $\left\{\mathcal{M}_i\right\}_{i=1}^{K}$; \quad Initialise: Joint data $\mathcal{C} = \{\}$\;
        \tcp{collecting samples}
        \For {$i \in \{1, 2, \cdots, K\}$}{
            Initialise: samples in ${\mathcal{M}_i}$,  $\mathcal{C}^{\mathcal{M}_i} = \{\}$, state $s$\; 

            \For{$t < H$} {
                {$b \gets \operatorname{Post}(\mathcal{C}^{\mathcal{M}_i})$};\quad
                $a \sim \pi_{\psi}^{+} (a|s, b)$\;
                Execute $a$ from $s$ in $\mathcal{M}_i$ to get $s', r$\;
                $\mathcal{C}^{\mathcal{M}_i} \gets \mathcal{C}^{\mathcal{M}_i} \cup \{s, a, s', r\}$\;
                $s \gets s'$\;
            }
            $\mathcal{C} =\mathcal{C} \cup \mathcal{C}^{\mathcal{M}_i}$
        }
        \tcp{learning policy, transition and reward models}
        \While{Learning}{
            Sample $\mathcal{D} \subseteq \mathcal{C}$\;
            $\psi \gets \psi - \nabla_{\psi} \mathcal{L}_{\text{policy}}(\mathcal{D})$\;
            $\phit \gets \phit - \nabla_{\phit} \mathcal{L}_{\text{model}}(\mathcal{D})$\;
            $\phir \gets \phir - \nabla_{\phir} \mathcal{L}_{\text{model}}(\mathcal{D})$\;
        }
    }
\end{algorithm}

\subsection{GLiBRL-SAC}
\vspace{-1em}
\begin{algorithm}[h]
    \caption{GLiBRL with SAC \label{alg:glibrl_sac}}
    \While {Training}{
    \tcp{collecting samples following \Algref{glibrlalg}}
        \While{Learning}{
            Sample $\mathcal{D} \subseteq \mathcal{C}$ of \textbf{variable-length} $N \in [1, \mathrm{Horizon}]$ \;
            {$b \gets \operatorname{Post}(\mathcal{D})$}\;
            $\psi \gets \psi - \nabla_{\psi} (\mathcal{L}_{\text{actor}}(\mathcal{D}, b) + \mathcal{L}_{\text{critic}}(\mathcal{D}, b) + \mathcal{L}_{\alpha})$\;
            Polyak averaging for target Q-networks within $\psi$\; 
            \tcp{Model updates following \Algref{glibrlalg}}
        }
    }
\end{algorithm}

\subsection{GLiBRL-PPO}
\vspace{-1em}
\begin{algorithm}[h]
    \caption{GLiBRL with PPO \label{alg:glibrl_ppo}}
    \While {Training}{
        \tcp{collecting samples following \Algref{glibrlalg}}
        Compute advantages $\hat{A}$ and return targets using $\mathcal{C}$\;
        \While{Learning}{
            Sample $\mathcal{D} \subseteq \mathcal{C}$ of \textbf{fixed length} $N$\;
            {$\left\{b_i\right\}_{i=1}^N \gets \left\{\operatorname{Post}(\mathcal{D}[1:i])\right\}_{i=1}^N$}\;
            $\psi \gets \psi - \nabla_{\psi} \mathcal{L}_{\text{actor}}\left(\mathcal{D}, \hat{A}, \left\{b_i\right\}_{i=1}^N\right)$\;
            \tcp{Model updates following \Algref{glibrlalg}}
        }
    }
\end{algorithm}

\clearpage

\subsection{Evidence Lower Bound}
\label{elboproof}
The evidence lower bound in \Eqref{elbo} is derived as follows
\begin{align*}
    \begin{split}
        \log p_{\phit, \phir}(\mathcal{C}) &= \sum_{i=1}^M\log \iint p(\mathcal{C}^{\mathcal{M}_i}, \thetat, \thetar) \frac{q(\thetat, \thetar|\mathcal{C}^{\mathcal{M}_i})}{q(\thetat, \thetar|\mathcal{C}^{\mathcal{M}_i})} d\thetat d\thetar \\
        &\ge \sum_{i=1}^M \mathbb{E}_{q}\left[\log \frac{p(\thetat, \thetar)}{q(\thetat, \thetar|\mathcal{C}^{\mathcal{M}_i})} + \log p(\mathcal{C}^{\mathcal{M}_i}|\thetat, \thetar)  \right] \\
        &= \sum_{i=1}^M { \mathbb{E}_{q}\left[\log p(\mathcal{C}^{\mathcal{M}_i}|\thetat, \thetar) \right] - D_{KL}\left({q(\thetat, \thetar|\mathcal{C}^{\mathcal{M}_i})} \Vert p(\thetat, \thetar)\right) } 
    \end{split}
\end{align*}

\subsection{Normal-Wishart-Normal Conjugacy}
\vspace{1em}
\label{nwconjugacy}
Given \Eqref{tprior} 
\begin{equation*}
    p(\thetat)  = \mathcal{MN}(\mut|\Mt, \xit^{-1}, \sigmat) \cdot \mathcal{W}(\sigmat^{-1}|\omegat^{-1}, \nut)
\end{equation*}
and \Eqref{tllh}
\begin{align*}
    \recallLabel{tllh}
\end{align*}
We prove \Eqref{tpost} 
\begin{align*}
    \recallLabel{tpost}
\end{align*}
where the posterior parameters are defined in \Eqref{post}.
\vspace{1em}
\subsubsection*{Proof:}
The density of the prior distribution $p(\thetat)$ is
\begin{align}
    \begin{split}
        p(\thetat) &=\frac{|\xit|^{D_S/2}}{\sqrt{(2\pi)^{D_TD_S}}|\sigmat|^{D_T/2}} \cdot \exp{\left[ -\frac{1}{2}\Tr\left( \sigmat^{-1} (\mut - \Mt)^\text{T} \xit  (\mut - \Mt) \right) \right]}\\
        & \cdot 
        \sqrt{\frac{|\omegat|^{\nut}}{2^{\nut D_S}}} \frac{|\sigmat|^{(1 + D_S -\nut)/2}}{\Gamma_{D_S}(\nut / 2)} \cdot \exp{\left[ -\frac{1}{2} \Tr{\left( \omegat \sigmat^{-1} \right)}\right]} \label{a210}
    \end{split}\\
    \begin{split}
            &\propto|\sigmat|^{{(1+D_S-\nut - D_T)/2}}  \cdot \exp{\Bigl\{ -\frac{1}{2} \Tr\Bigl[ \sigmat^{-1} \Bigl[(\mut - \Mt)^\text{T} \xit  (\mut - \Mt)  +\omegat \Bigr]\Bigr ] \Bigr\}} \label{a211}
    \end{split}
\end{align}

where from \Eqref{a210} to \Eqref{a211} we treat multiplicative parameters irrelevant to $\thetat$ as constants. The joint density of $p_{\phit}(\thetat, \mathbf{S}_i' | \mathbf{S}_i, \mathbf{A}_i) = p(\thetat) \cdot p_{\phit}(\mathbf{S}_i'|\mathbf{S}_i, \mathbf{A}_i, \thetat)$ is hence

{
\begin{align}
    \begin{split}
           & \frac{|\xit|^{D_S/2}}{\sqrt{(2\pi)^{D_TD_S}}|\sigmat|^{D_T/2}} \cdot \exp{\left[ -\frac{1}{2}\Tr\left( \sigmat^{-1} (\mut - \Mt)^\text{T} \xit  (\mut - \Mt) \right) \right]}\\
            & \cdot 
            \sqrt{\frac{|\omegat|^{\nut}}{2^{\nut D_S}}} \frac{|\sigmat|^{(1 + D_S -\nut)/2}}{\Gamma_{D_S}(\nut / 2)} \cdot \exp{\left[ -\frac{1}{2} \Tr{\left( \omegat \sigmat^{-1} \right)}\right]} \\
            &  \cdot \frac{1}{\sqrt{(2\pi)^{ND_S}}|\sigmat|^{N/2}} \cdot \exp{\left[ -\frac{1}{2} \Tr\left( \sigmat^{-1} (\mathbf{S}_i' - \Ct\mut)^{\text{T}} (\mathbf{S}_i' - \Ct\mut)\right) \right]} \label{a215}
    \end{split} \\[10pt]
    \begin{split}
        &\propto |\sigmat|^{{(1 +D_S-\nut - N - D_T)/2}}  \cdot \exp{\left[ -\frac{1}{2}\Tr\left( \sigmat^{-1} (\mut - \Mt)^\text{T} \xit  (\mut - \Mt) \right) \right]} \\
        & \qquad \qquad \qquad \qquad \cdot\exp{\left[ -\frac{1}{2} \Tr\left( \sigmat^{-1} \left( (\mathbf{S}_i' - \Ct\mut)^{\text{T}} (\mathbf{S}_i' - \Ct\mut) + \omegat\right) \right)  \right]}
        \label{a216}
    \end{split}
\end{align}
}
Matching the second-order then the first-order term with respect to $\mut$, we can rewrite 
\begin{align}
    \begin{split}
        \text{\ref{a216}} = |\sigmat|^{{(1+D_S-\nut' - D_T)/2}}  \cdot \exp{\Bigl\{ -\frac{1}{2} \Tr\Bigl[ \sigmat^{-1} \Bigl[(\mut - \Mt')^\text{T} \xit'  (\mut - \Mt')  +\omegat' \Bigr]\Bigr ] \Bigr\}} \label{a217}
    \end{split}
\end{align}
We can find \Eqref{a211} and \Eqref{a217} match exactly, indicating the Normal-Wishart-Normal conjugacy. Note, we just use the posterior update of $p(\thetat)$ as an example. The exact same proof applies to the posterior update of $p(\thetar)$ as well. Such conjugacy allows exact posterior update and marginal likelihood, enabling efficient learning.  

\subsection{Marginal Log-likelihood of Normal-Wishart-Normal}
\label{llhproof}
We prove \Eqref{llhour} has the following closed form
\begin{align*}
    \recallLabel{llhour}
\end{align*}

\subsubsection*{Proof:} 

Consider
\begin{equation}
    \begin{split}
        p_{\phit} ( \mathbf{S}_i'|\mathbf{S}_i, \mathbf{A}_i) = \frac{p_{\phit}(\thetat,  \mathbf{S}_i'|\mathbf{S}_i, \mathbf{A}_i)}{p_{\phit}(\thetat|\mathbf{S}_i, \mathbf{A}_i, \mathbf{S}_i')}
    \end{split}
\end{equation}
From \Apref{nwconjugacy}, we know the numerator is

\begin{equation}
    \frac{|\xit|^{D_S/2}  |\omegat|^{\nut/2}} {2^{\nut D_S/2}\cdot(2\pi)^{D_S(D_T+N)/2}\cdot\Gamma_{D_S}(\nut/2)} \cdot \text{\Eqref{a217}}
\end{equation}

The denominator is
\begin{equation}
    \frac{|\xit'|^{D_S/2}  |\omegat'|^{\nut'/2}} {2^{\nut' D_S/2}\cdot(2\pi)^{D_SD_T/2}\cdot\Gamma_{D_S}(\nut'/2)} \cdot \text{\Eqref{a217}}
\end{equation}

Hence,
\begin{equation}
    \begin{split}
        p_{\phit} (\mathbf{S}_i'|\mathbf{S}_i, \mathbf{A}_i) = \frac{1}{(2\pi)^{D_SN/2}} \cdot \frac{|\xit|^{D_S/2}  |\frac{1}{2}\omegat|^{\nut/2} \cdot\Gamma_{D_S}(\nut'/2)}{|\xit'|^{D_S/2}  |\frac{1}{2}\omegat'|^{\nut'/2} \cdot\Gamma_{D_S}(\nut/2)} 
    \end{split}
\end{equation}

Similarly,
\begin{equation}
    \begin{split}
        p_{\phir} (\mathbf{r}_i | \mathbf{S}_i, \mathbf{A}_i, \mathbf{S}_i') = \frac{1}{(2\pi)^{N/2}} \cdot \frac{|\xir|^{1/2}  |\frac{1}{2}\omegar|^{\nur/2} \cdot\Gamma(\nur'/2)}{|\xir'|^{1/2}  |\frac{1}{2}\omegar'|^{\nur'/2} \cdot\Gamma(\nur/2)} 
    \end{split}
\end{equation}

Note that
\begin{equation}
    p_{\phit, \phir}(\mathcal{C}^{\mathcal{M}_i}) = p_{\phit} (\mathbf{S}_i'|\mathbf{S}_i, \mathbf{A}_i) p_{\phir} (\mathbf{r}_i | \mathbf{S}_i, \mathbf{A}_i, \mathbf{S}_i') p(\mathbf{S}_i, \mathbf{A}_i)
\end{equation}

By taking the logarithm, and note the independence of $p(\mathbf{S}_i, \mathbf{A}_i)$ with respect to $\phit, \phir$,
\begin{equation}
    \log p_{\phit, \phir}(\mathcal{C}^{\mathcal{M}_i}) = \log p_{\phit} (\mathbf{S}_i'|\mathbf{S}_i, \mathbf{A}_i) + \log p_{\phir} (\mathbf{r}_i | \mathbf{S}_i, \mathbf{A}_i, \mathbf{S}_i') + \text{const.}
\end{equation}

Sum the above equation on both sides with respect to $i$, then we have \Eqref{llhour}.
\newpage
\subsection{Diagonal Noise Inference with Matrix-Normal-Gamma}
\label{mngconjugacy}
The Matrix-Normal-Wishart formulation in \Secref{mlearning1} learns a full covariance. For simplicity, a diagonal formulation can be considered as follows 
\begin{align}
    p^{\mathrm{diag}}(\thetat)
    &=\mathcal{MN}(\mut\mid\Mt,\xit^{-1},\sigmat)
    \prod_{d=1}^{D_S}\operatorname{Gam}\left([\sigmat^{-1}]_{dd}\mid a_{T,d},b_{T,d}\right)\label{mngprior}\\
    p^{\mathrm{diag}}(\thetar)
    &=\mathcal{MN}(\mur\mid\Mr,\xir^{-1},\sigmar)
    \operatorname{Gam}\left(\sigmar^{-1}\mid a_R,b_R\right) \label{mngrewardprior}
\end{align}

where $\operatorname{Gam}(X\,\vert\,a, b)$ defines a Gamma distribution with random variable $X$, shape $a$ and rate $b$, and $\sigmat=\operatorname{diag}\left([\sigmat]_{11},\ldots,[\sigmat]_{D_S,D_S}\right)$ is the diagonal transition covariance. 

\subsubsection*{Closed-form posterior update and conjugacy}
The posterior distributions remain Matrix-Normal-Gamma:
\begin{align}
    p_{\phit}^{\mathrm{diag}}(\thetat\mid\mathbf{S}_i,\mathbf{A}_i,\mathbf{S}_i')
    &=\mathcal{MN}(\mut\mid\Mt',{\xit'}^{-1},\sigmat)
    \prod_{d=1}^{D_S}\operatorname{Gam}\left([\sigmat^{-1}]_{dd}\mid a_{T,d}',b_{T,d}'\right) \label{mngposterior}\\
    p_{\phir}^{\mathrm{diag}}(\thetar\mid\mathbf{S}_i,\mathbf{A}_i,\mathbf{S}_i',\mathbf{r}_i)
    &=\mathcal{MN}(\mur\mid\Mr',{\xir'}^{-1},\sigmar)
    \operatorname{Gam}\left(\sigmar^{-1}\mid a_R',b_R'\right) \label{mngrewardposterior}
\end{align}

For the transition model, the posterior parameters are
\begin{align}
    \Mt'&={\xit'}^{-1}\left(\Ct^{\text{T}}\mathbf{S}'+\xit\Mt\right)
    &\xit'&=\Ct^{\text{T}}\Ct+\xit \label{mngmeanupdate}\\
    a_{T,d}'&=a_{T,d}+\frac{N}{2}
    &b_{T,d}'&=b_{T,d}+\frac{1}{2}\left[
    {\mathbf{S}'}^{\text{T}}\mathbf{S}'
    +\Mt^{\text{T}}\xit\Mt
    -{\Mt'}^{\text{T}}\xit'\Mt'\right]_{dd} \label{mngnoiseupdate}
\end{align}

For the reward model, the posterior parameters are
\begin{align}
    \Mr'&={\xir'}^{-1}\left(\Cr^{\text{T}}\mathbf{r}+\xir\Mr\right)
    &\xir'&=\Cr^{\text{T}}\Cr+\xir \label{mngrewardmeanupdate}\\
    a_R'&=a_R+\frac{N}{2}
    &b_R'&=b_R+\frac{1}{2}\left(
    \mathbf{r}^{\text{T}}\mathbf{r}
    +\Mr^{\text{T}}\xir\Mr
    -{\Mr'}^{\text{T}}\xir'\Mr'\right) \label{mngrewardupdate}
\end{align}

The posterior means and row precisions are identical to those in \Eqref{post}. Only the full Wishart update is replaced by independent Gamma updates.

\subsubsection*{Online execution and time complexity}
The posterior update of Matrix-Normal-Gamma only differs with that of Matrix-Normal-Wishart in the noise update. Therefore, during online execution, the time complexity of the transition model update is $\mathcal{O}\left(D_T\cdot\max(D_T, D_S)\right)$. The quadratic Wishart update term now disappears. The reward model update remains $\mathcal{O}(D_R^2)$.

\subsubsection*{ELBO-free marginal log-likelihood}
\label{mngllhproof}
Follow the proof in \Apref{llhproof}, the marginal likelihood can be derived as
\begin{align}
    \log p_{\phit,\phir}^{\mathrm{diag}}(\mathcal{C})
    =-\sum_{i=1}^{M}\left[
    \frac{D_S}{2}\log|\xit'|
    +\sum_{d=1}^{D_S}a_{T,d}'\log b_{T,d}'
    +\frac{1}{2}\log|\xir'|
    +a_R'\log b_R'\right]+\mathrm{const.} \label{llhourdiag}
\end{align}

\subsubsection*{Task representation}
The posterior means $\Mt'$ and $\Mr'$ in \Eqref{mngmeanupdate} and \Eqref{mngrewardmeanupdate} are identical to those under the Wishart model. Consequently, the task representation $f_{T,R}$ in \Eqref{usedrep} is unchanged. The correspondence identities in \Eqref{finalrep} and \Eqref{finalrepunnorm}, as well as the alternative representations in \Apref{beliefrep}, therefore continue to hold without modification under diagonal noise.

\newpage
\subsection{Choices of Belief Representations}
\label{beliefrep}
\begin{figure}[h]
    \centering
    \includegraphics[width=0.9\linewidth]{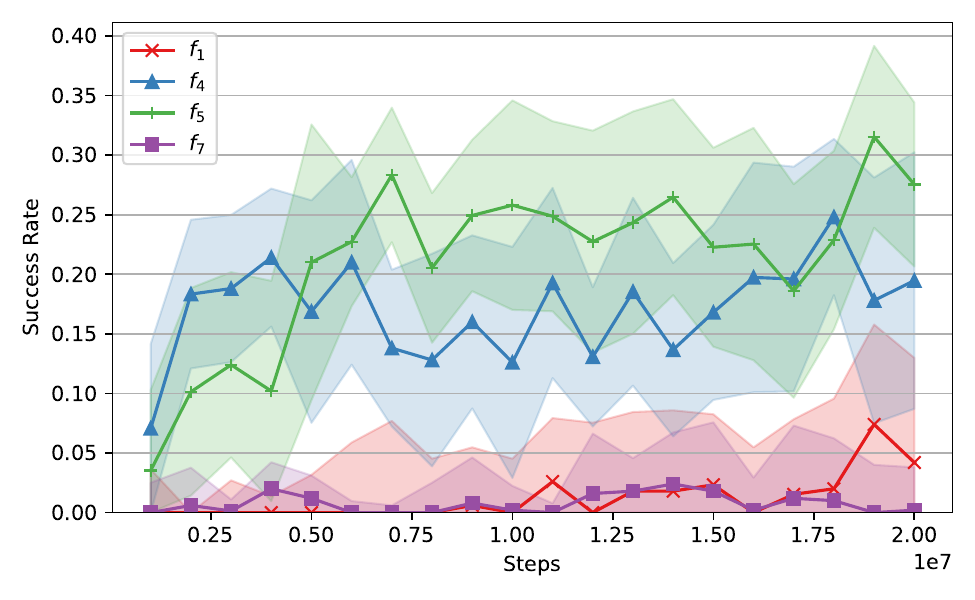}
    \caption{Success rate of different task representation functions in ML10. \label{repfunc}}
\end{figure}

We list several candidate $f$s:
{
\begin{align*}
    f_1\left(\Mt', \Mr'\right) &= \left[\mathrm{vec}\left(\Mt'\right)^{\text{T}} \;  {\Mr'}^{\text{T}}\right]^{\text{T}} \\ 
    f_2\left(\Mt', \Mr'\right) &= \left[\mathrm{tril}\left(\Mt'{\Mt'}^{\text{T}}\right)^{\text{T}} \;  {\Mr'}^{\text{T}}\right]^{\text{T}}  \\
    f_3\left(\Mt', \Mr'\right) &= \left[\mathrm{tril}\left({\Mt'}^{\text{T}}\Mt'\right)^{\text{T}} \;  {\Mr'}^{\text{T}}\right]^{\text{T}}  \\
    f_4\left(\Mt', \Mr'\right) &= \left[\mathcal{N}_\epsilon\!\left(\mathrm{vec}\left(\Mt'\right)\right)^{\text{T}} \;  \mathcal{N}_\epsilon\!\left(\Mr'\right)^{\text{T}}\right]^{\text{T}} \\
    f_5\left(\Mt', \Mr'\right) &= \left[\mathcal{N}_\epsilon\!\left(\mathrm{tril}\left({\Mt'}{\Mt'}^{\text{T}}\right)\right)^{\text{T}} \;  \mathcal{N}_\epsilon\!\left(\Mr'\right)^{\text{T}}\right]^{\text{T}} \\
    f_6\left(\Mt', \Mr'\right) &= \left[\mathcal{N}_\epsilon\!\left(\mathrm{tril}\left({\Mt'}^{\text{T}}{\Mt'}\right)\right)^{\text{T}} \;  \mathcal{N}_\epsilon\!\left(\Mr'\right)^{\text{T}}\right]^{\text{T}} \\
    f_7\left(\Mt', \Mr'\right) &= \left[g_T\left(\Mt'\right)^{\text{T}} \; g_R\left(\Mr'\right)^{\text{}}\right]^{\text{T}}
\end{align*}}

\noindent \noindent where $\mathrm{tril}(\cdot)$ retrieves the flattened lower triangle of matrices and $g_T(\cdot), g_R(\cdot)$ are neural networks. $f_1$ directly flattens and concatenates the task means, resulting in $D_T \times D_S + D_R$ dimensional representations. $f_2$ and $f_3$ are used for scenarios with large $D_S$ and $D_T$, giving $D_T(D_T+1)/2 + D_R$ and $D_S(D_S+1)/2 + D_R$ dimensional representations respectively. $f_4$, $f_5$, and $f_6$ are their $\epsilon$-normalised counterparts (\Eqref{zerosafenorm}). $f_7$ suggests learnable belief representations, trained by backpropagating the reinforcement-learning loss. While we focus on $f_5$, i.e. $f_{T, R}$ (\Eqref{usedrep}), we compare these design choices empirically and report the success rates of $f_1$, $f_4$, $f_5$, and $f_7$ on MetaWorld ML10 over 10 seeds in \Figref{repfunc}.

As shown in \Figref{repfunc}, $f_5$, i.e., $f_{T,R}$ (\Eqref{usedrep}), performs best overall. The low success rates of $f_1$ and $f_7$ show that neither raw posterior nor an unconstrained learned mapping suffice in this setting. Comparing $f_1$ and $f_4$ indicates that normalisation provides most of the gain. It substantially stabilises training and, consistently with \Eqref{finalrep}, bounds the squared distance between task representations in $[0,8)$. Comparing $f_4$ and $f_5$ suggests that the quadratic transition feature further improves success, partly by reducing the input dimension seen by the policy.

\newpage
\input{sections/theory_fixed}
\subsection{Implementation Details}
\label{impdetails}
We report the details of our implementations of \nom and other methods. The implementation of \nom is simple with only model networks $\phit, \phir$ and policy networks $\pi^+_{\psi}$. The model networks $\phit, \phir$ are Multi-Layer Perceptrons (MLPs) consisting of \textit{feature} and \textit{mixture} networks. Feature networks convert raw states and actions to features and are shared in $\phit$ and $\phir$. Mixture networks mix the state and action features, further improving the representativeness. The policy network is actor-only for PPO with linear baseline as suggested by \cite{mclean2025meta} in MetaWorld ML10/45, and actor-critic for SAC in MuJoCo tasks. We adopt the diagonal noise inference in ML10 and full noise inference in ML45. 

For reproducing all MuJoCo results, we follow the official implementations in TrMRL, \pearl and \vbad. All results are run with \textsc{gymnasium==1.2.3} and \textsc{mujoco==3.6.0}. For MetaWorld results, we rerun \maml and \rltwo using the more performant implementation provided by \cite{beck2023hypernetworks}, while re-implement and rerun \vbad as the official implementation does not contain MetaWorld environments. We use the tuned hyperparameters from official implementations in \maml, \rltwo and \vbad. For remaining baselines, we use the reported results from \cite{shala2025efficient}.

The table of all related hyperparameters of \nom is shown in \Apref{hyperparam}.

\newpage
\clearpage
\subsection{ML10 Success Rate Comparisons Per-task}
\label{ml10pertask}
\begin{figure}[!hp]
    \centering
    \begin{subfigure}{\textwidth}
        \centering
        \includegraphics[width=0.49\textwidth]{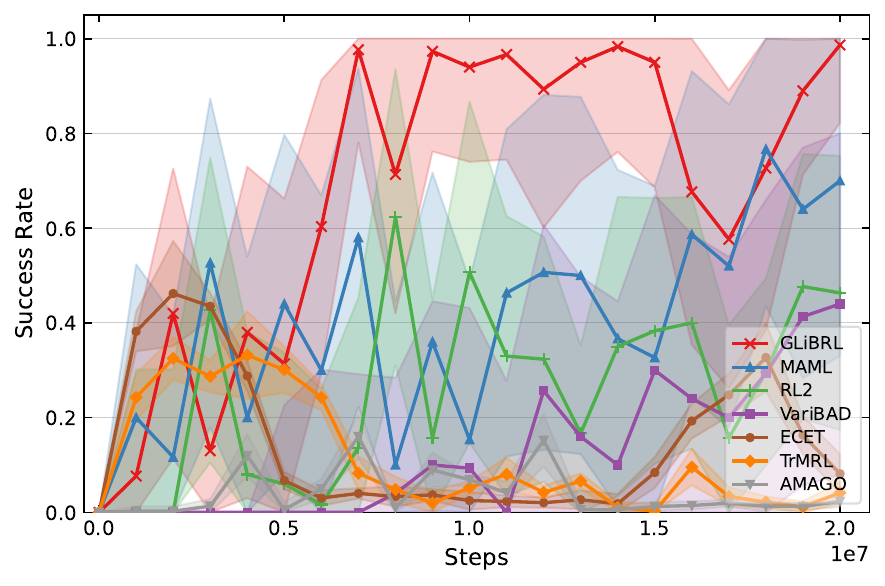}
        \hfill
        \includegraphics[width=0.49\textwidth]{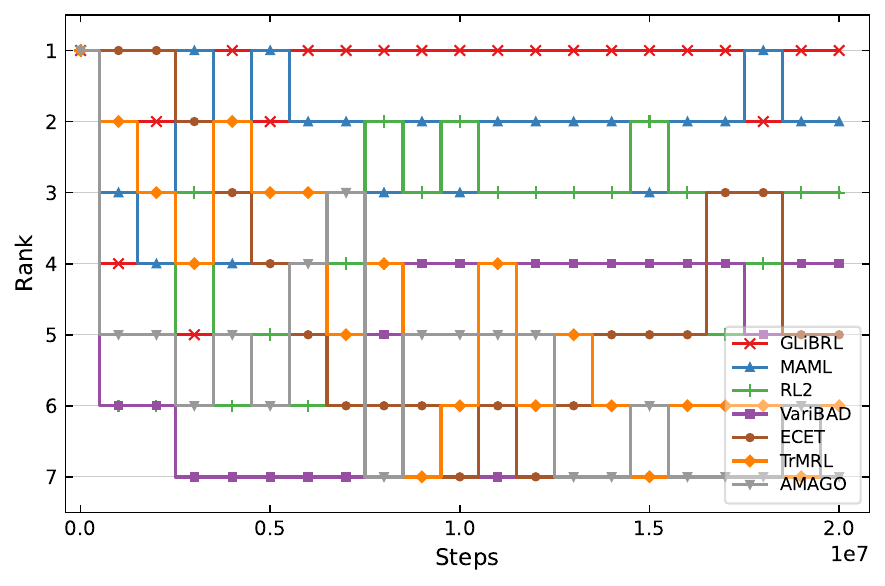}
        \caption{Door Close}
    \end{subfigure}
    \\
    \begin{subfigure}{\textwidth}
        \centering
        \includegraphics[width=0.49\textwidth]{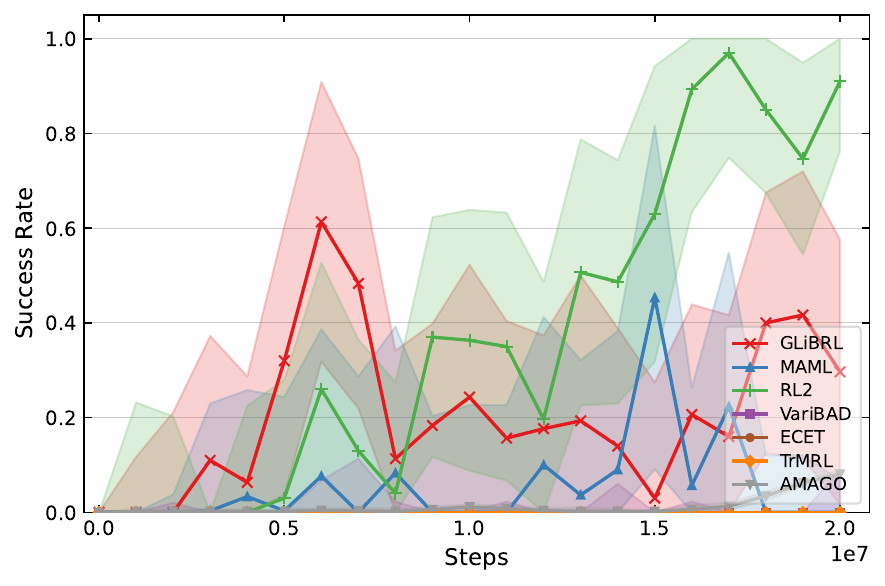}
        \hfill
        \includegraphics[width=0.49\textwidth]{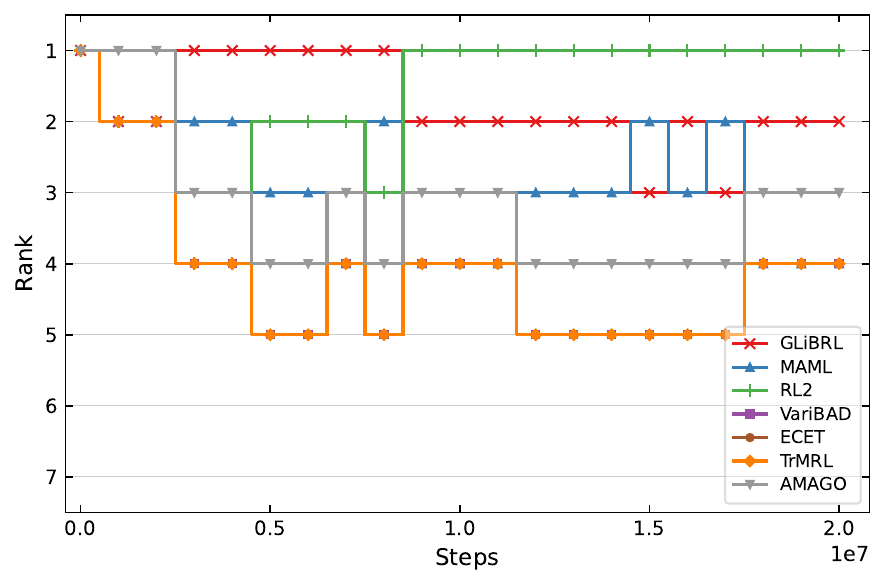}
        \caption{Drawer Open}
    \end{subfigure}
    \\
    \begin{subfigure}{\textwidth}
        \centering
        \includegraphics[width=0.49\textwidth]{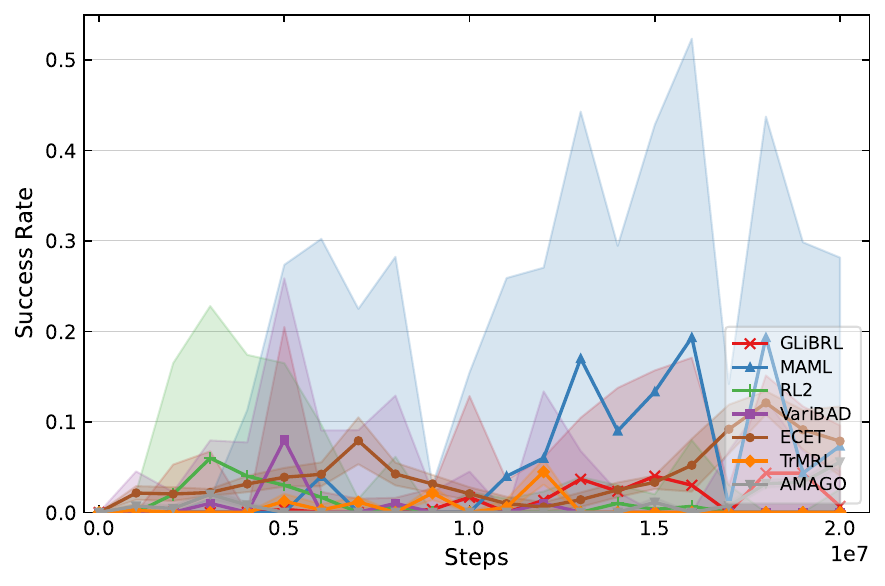}
        \hfill
        \includegraphics[width=0.49\textwidth]{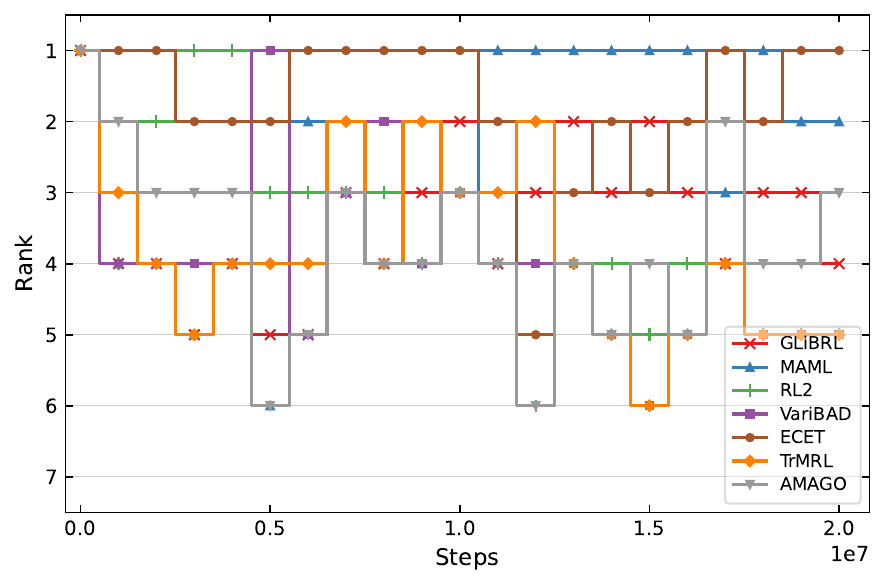}
        \caption{Lever Pull}
    \end{subfigure}
    \\
    \begin{subfigure}{\textwidth}
        \centering
        \includegraphics[width=0.49\textwidth]{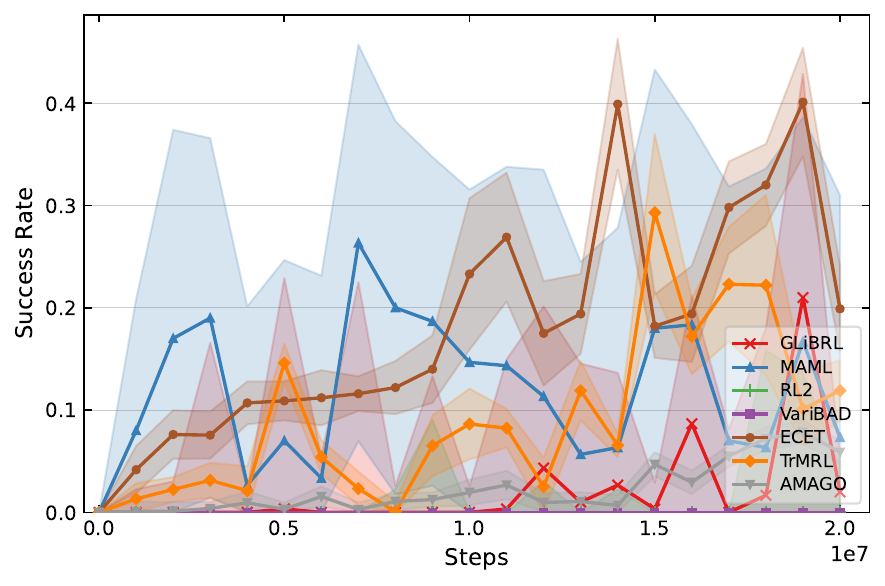}
        \hfill
        \includegraphics[width=0.49\textwidth]{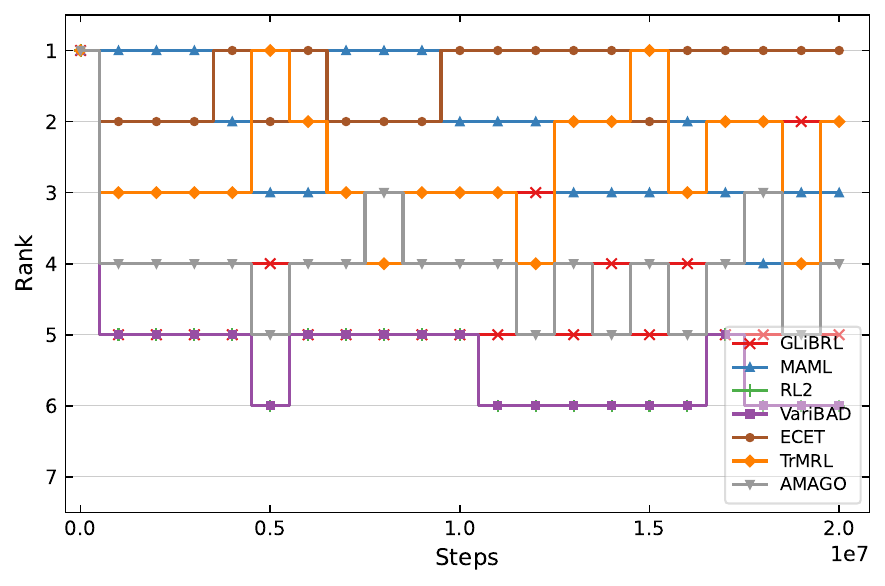}
        \caption{Sweep Into}
    \end{subfigure}
    \caption{Testing success rate (left) and success rate rank $(\downarrow)$ (right) for each scenario in ML10. }
    \label{fig:ml10pertask}
\end{figure}

\newpage

\begin{figure}[!tp]
    \ContinuedFloat
    \centering
    \begin{subfigure}{\textwidth}
        \centering
        \includegraphics[width=0.49\textwidth]{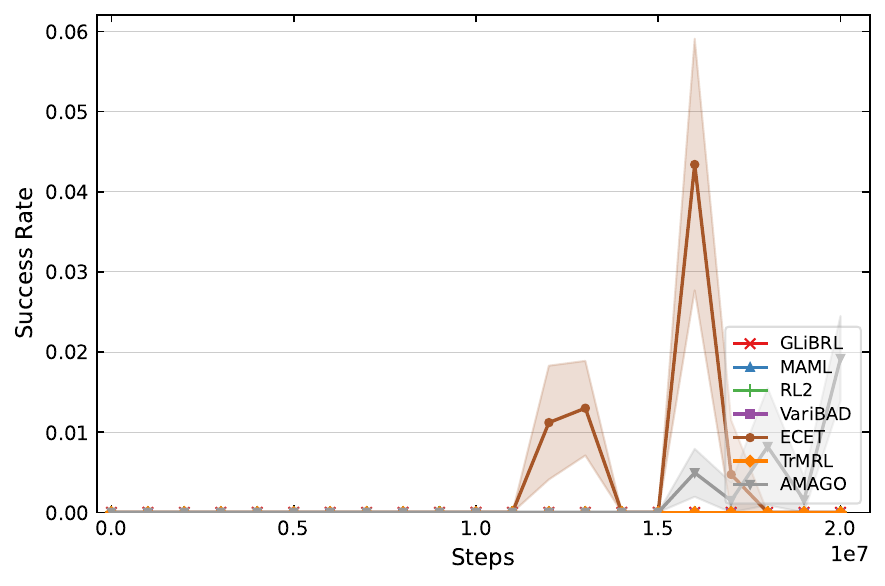}
        \hfill
        \includegraphics[width=0.49\textwidth]{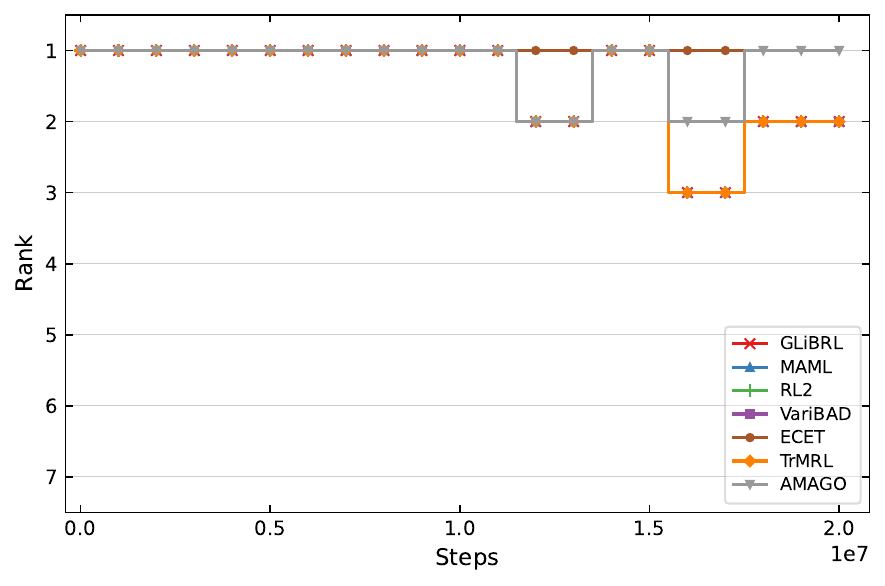}
        \caption{Shelf Place}
    \end{subfigure}
    \caption{Testing success rate (left) and success rate rank $(\downarrow)$ (right) for each scenario in ML10.}
\end{figure}

\vspace{3em}

\begin{table}[h]
  \centering
  \caption{Mean rank $(\downarrow)$ of each method over training in ML10.}
  \label{tab:ml10-pertask-rank}
  \setstretch{1.3}
  \begin{tabular}{lccccccc}
    \toprule
    Environment & GLiBRL & MAML & RL2 & VariBAD & ECET & TrMRL & AMAGO \\
    \midrule
    door-close & \textbf{1.48} & 2.19 & 3.62 & 5.00 & 4.38 & 4.71 & 5.52 \\
    drawer-open & 1.76 & 2.90 & \textbf{1.62} & 4.10 & 4.10 & 4.10 & 3.14 \\
    lever-pull & 3.38 & 2.57 & 3.52 & 3.90 & \textbf{1.71} & 3.81 & 3.71 \\
    shelf-place & 1.43 & 1.43 & 1.43 & 1.43 & \textbf{1.14} & 1.43 & 1.19 \\
    sweep-into & 4.43 & 2.14 & 5.29 & 5.29 & \textbf{1.38} & 2.57 & 4.00 \\
    \cmidrule(lr){1-8}
    \textit{Average} & 2.50 & \textbf{2.25} & 3.10 & 3.94 & 2.54 & 3.32 & 3.51 \\
    \bottomrule
  \end{tabular}
\end{table}

\newpage
\clearpage
\subsection{ML45 Success Rate Comparisons Per-task}
\label{ml45pertask}
\begin{figure}[!hp]
    \centering
    \begin{subfigure}{\textwidth}
        \centering
        \includegraphics[width=0.49\textwidth]{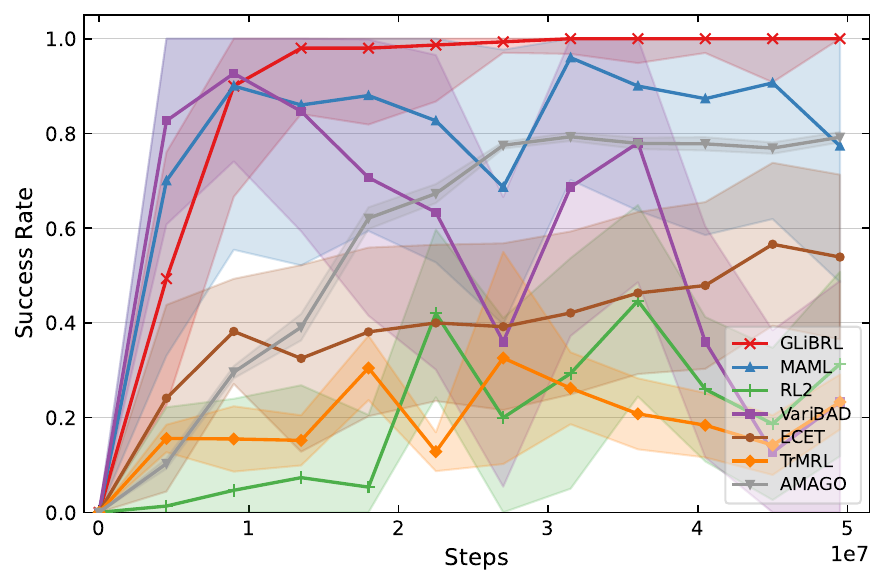}
        \hfill
        \includegraphics[width=0.49\textwidth]{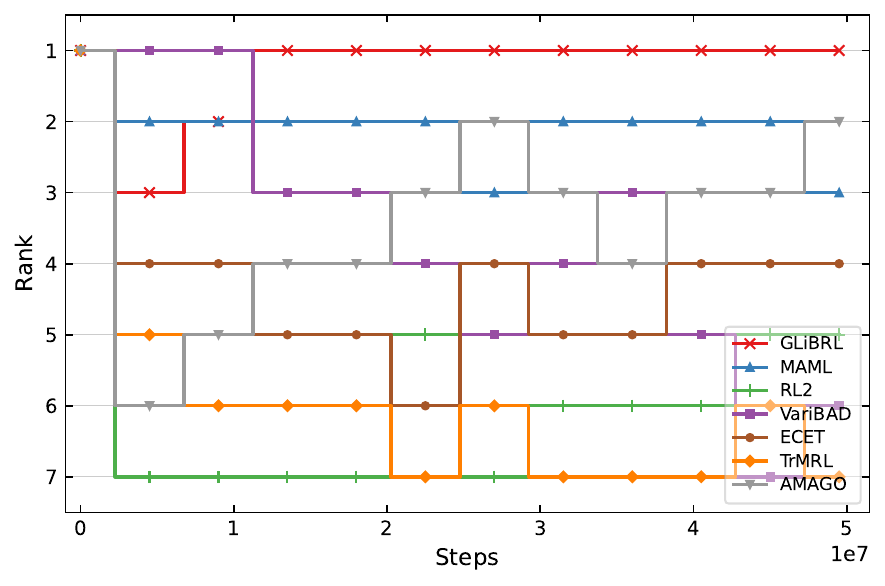}
        \caption{Door Lock}
    \end{subfigure}
    \\[1ex]
    \begin{subfigure}{\textwidth}
        \centering
        \includegraphics[width=0.49\textwidth]{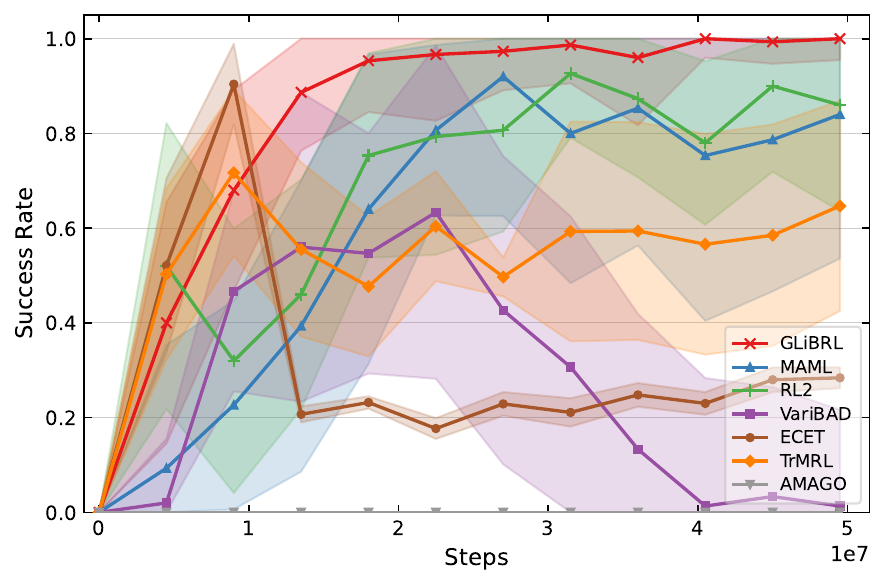}
        \hfill
        \includegraphics[width=0.49\textwidth]{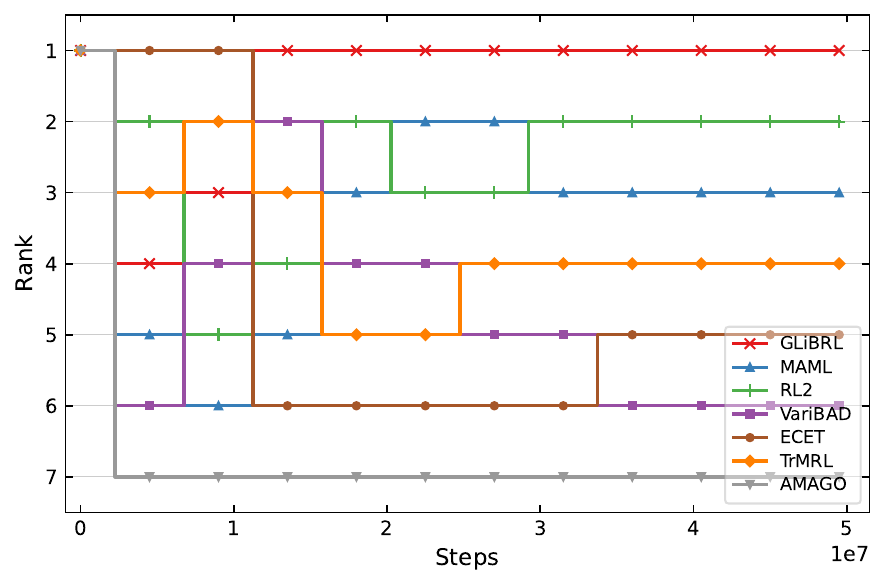}
        \caption{Door Unlock}
    \end{subfigure}
    \\[1ex]
    \begin{subfigure}{\textwidth}
        \centering
        \includegraphics[width=0.49\textwidth]{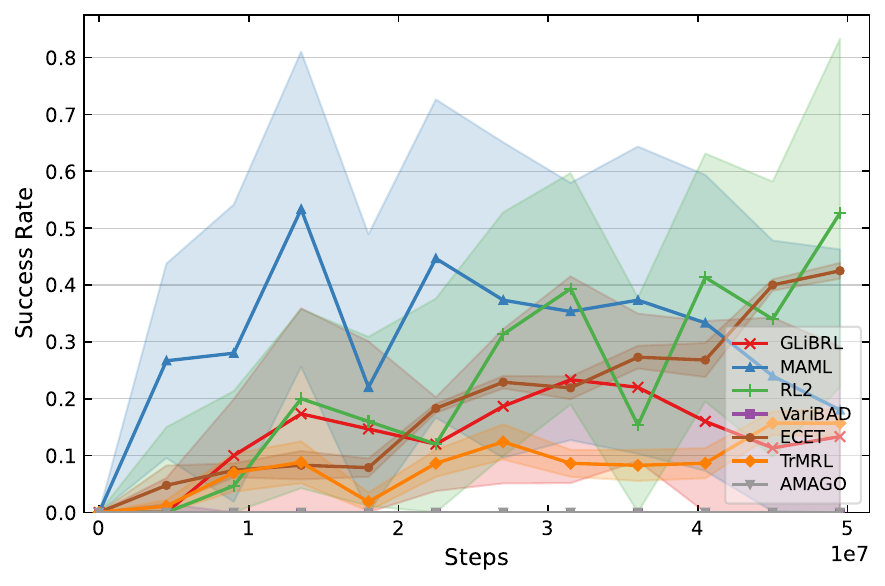}
        \hfill
        \includegraphics[width=0.49\textwidth]{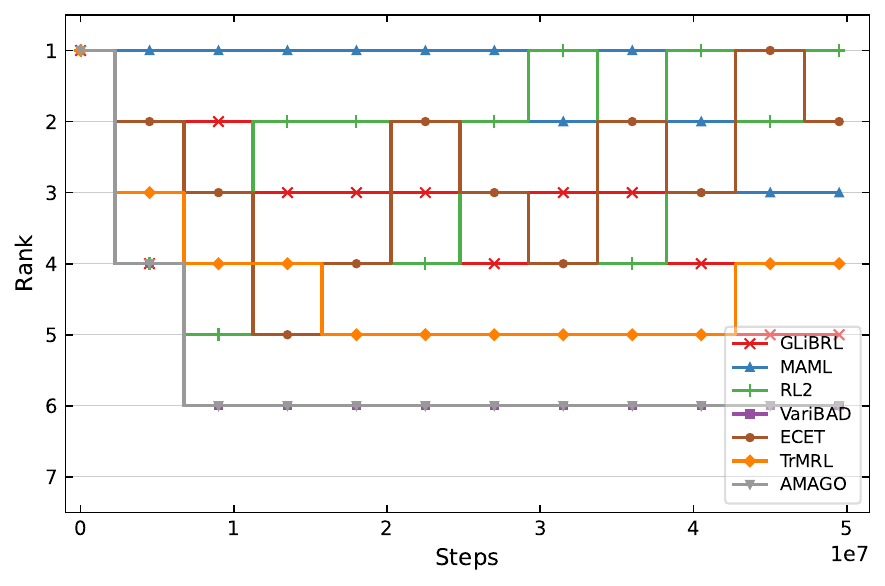}
        \caption{Hand Insert}
    \end{subfigure}
    \\
    \begin{subfigure}{\textwidth}
        \centering
        \includegraphics[width=0.49\textwidth]{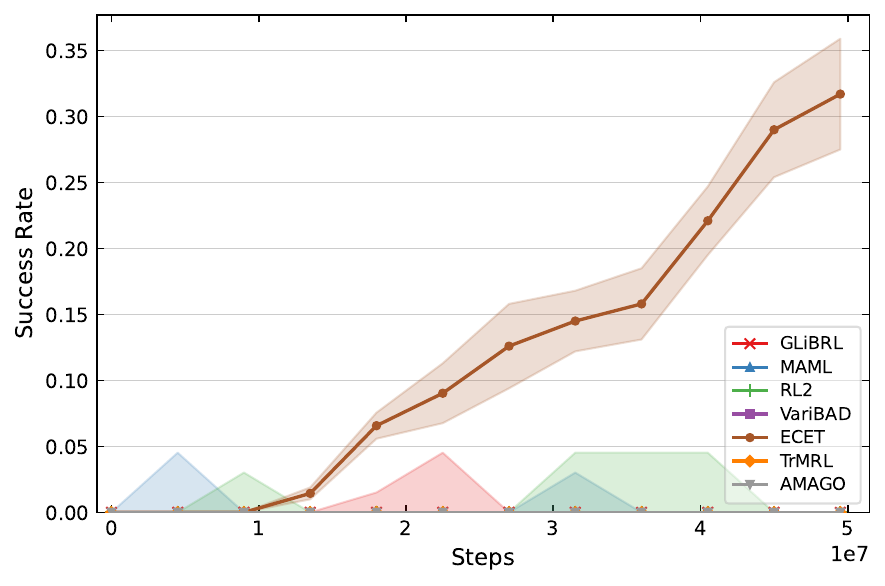}
        \hfill
        \includegraphics[width=0.49\textwidth]{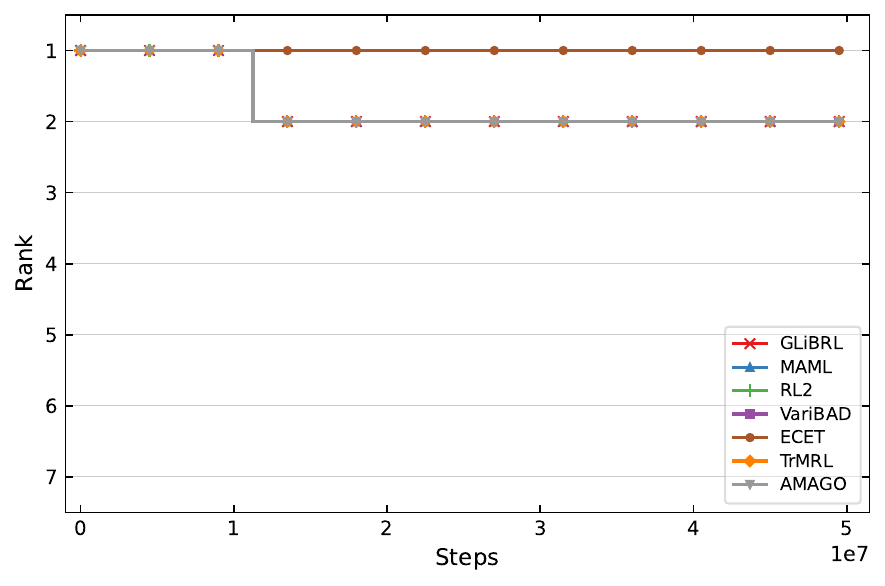}
        \caption{Bin Picking}
    \end{subfigure}
    \caption{Testing success rate (left) and success rate rank $(\downarrow)$ (right) for each scenario in ML45.}
    \label{fig:ml45pertask}
\end{figure}

\newpage

\begin{figure}[!tp]
    \ContinuedFloat
    \centering
    \begin{subfigure}{\textwidth}
        \centering
        \includegraphics[width=0.49\textwidth]{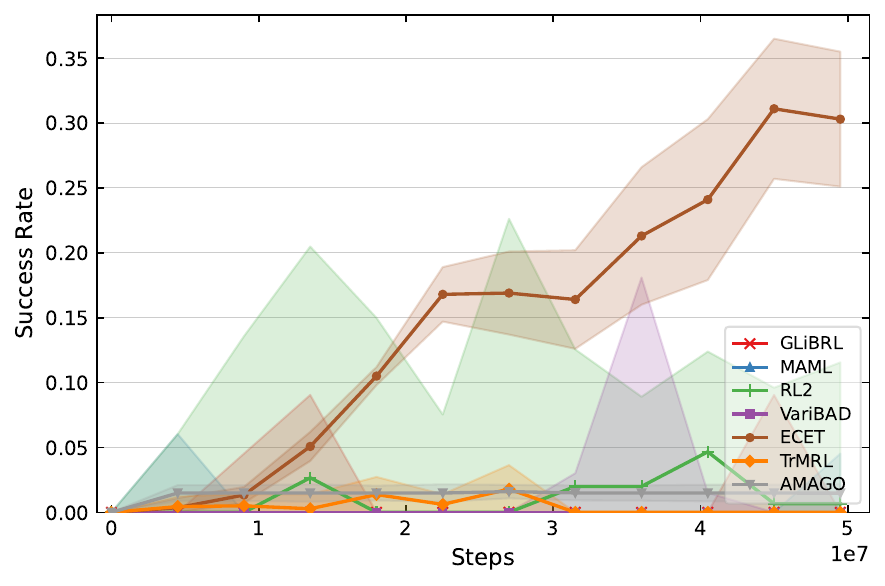}
        \hfill
        \includegraphics[width=0.49\textwidth]{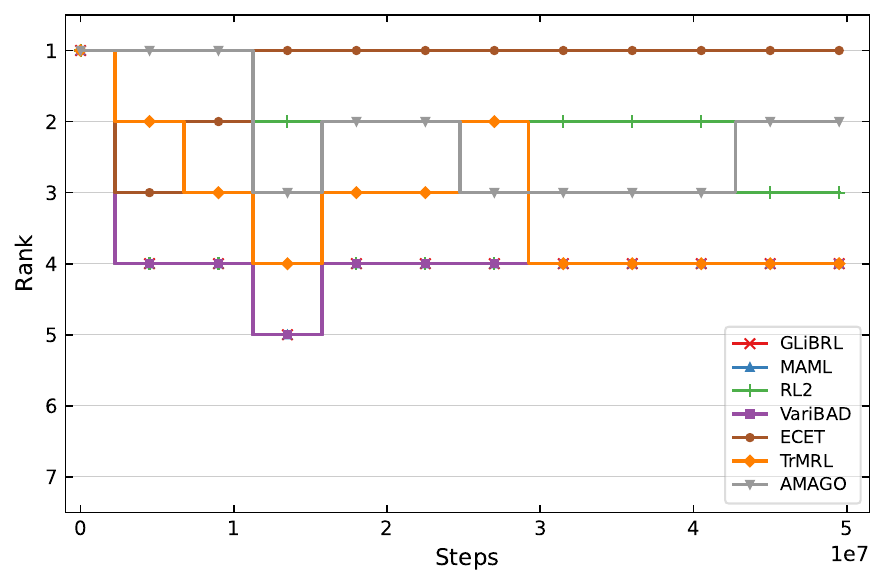}
        \caption{Box Close}
    \end{subfigure}
    \caption{Testing success rate (left) and success rate rank $(\downarrow)$ (right) for each scenario in ML45.}
\end{figure}

\vspace{3em}

\begin{table}[h]
  \centering
  \caption{Mean rank $(\downarrow)$ of each method over training in ML45.}
  \label{tab:ml45-pertask-rank}
  \setstretch{1.3}
  \begin{tabular}{lccccccc}
    \toprule
    Environment & GLiBRL & MAML & RL2 & VariBAD & ECET & TrMRL & AMAGO \\
    \midrule
    bin-picking & 1.75 & 1.75 & 1.75 & 1.75 & \textbf{1.00} & 1.75 & 1.75 \\
    box-close & 3.83 & 3.83 & 2.92 & 3.83 & \textbf{1.25} & 3.17 & 2.17 \\
    door-lock & \textbf{1.25} & 2.08 & 5.75 & 3.58 & 4.25 & 5.92 & 3.33 \\
    door-unlock & \textbf{1.42} & 3.25 & 2.50 & 4.58 & 4.42 & 3.58 & 6.50 \\
    hand-insert & 3.33 & \textbf{1.50} & 2.42 & 5.42 & 2.67 & 4.17 & 5.42 \\
    \cmidrule(lr){1-8}
    \textit{Average} & \textbf{2.32} & 2.48 & 3.07 & 3.83 & 2.72 & 3.72 & 3.83 \\
    \bottomrule
  \end{tabular}
\end{table}

\newpage
\newpage

\subsection{Qualitative Results on Task Identifiability}
\label{taskid}
\begin{figure}[h]
    \centering
    \setlength{\tabcolsep}{2pt}
    \begin{tabular}{ m{0.07\linewidth} m{0.21\linewidth} m{0.21\linewidth} | m{0.21\linewidth} m{0.21\linewidth} }
        \hline \\
        & \multicolumn{2}{c|}{\textbf{\nom}} & \multicolumn{2}{c}{\textbf{TrMRL}} \\
        & \centering $\text{step} = 0$ & \centering $\text{step} = 5e5$ & \centering $\text{step} = 0$ & \centering\arraybackslash $\text{step} = 5e6$ \\[2mm]
        
        \textbf{Run 1} & 
        \includegraphics[width=\linewidth]{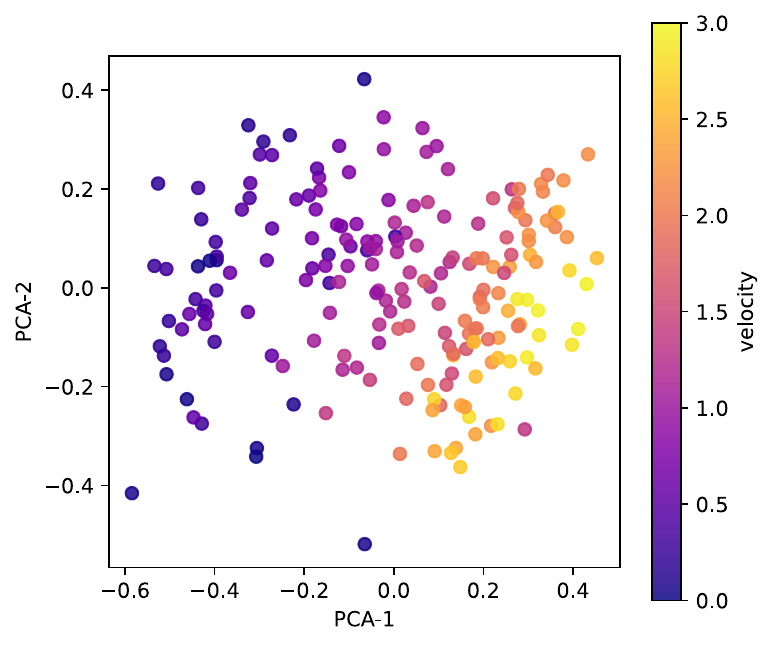} & 
        \includegraphics[width=\linewidth]{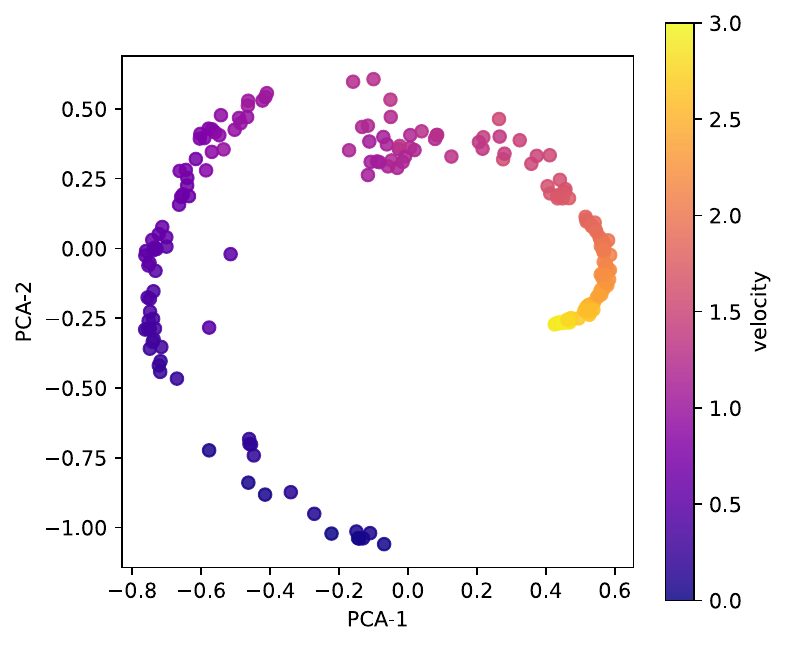} & 
        \includegraphics[width=\linewidth]{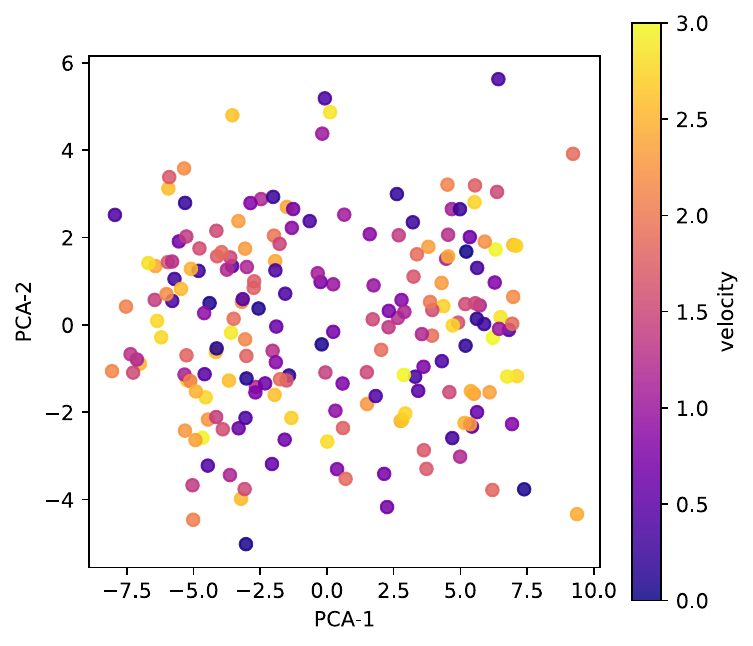} & 
        \includegraphics[width=\linewidth]{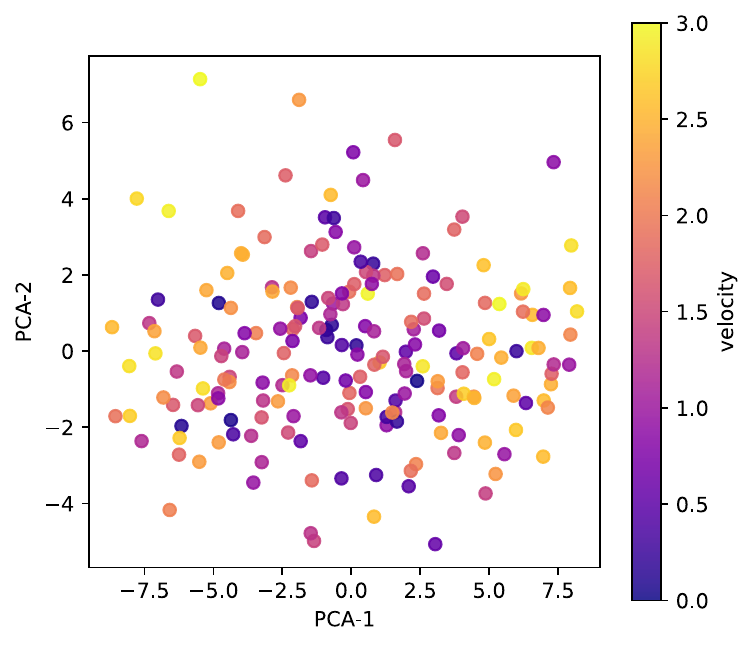} \\[1mm]
        
        \textbf{Run 2} & 
        \includegraphics[width=\linewidth]{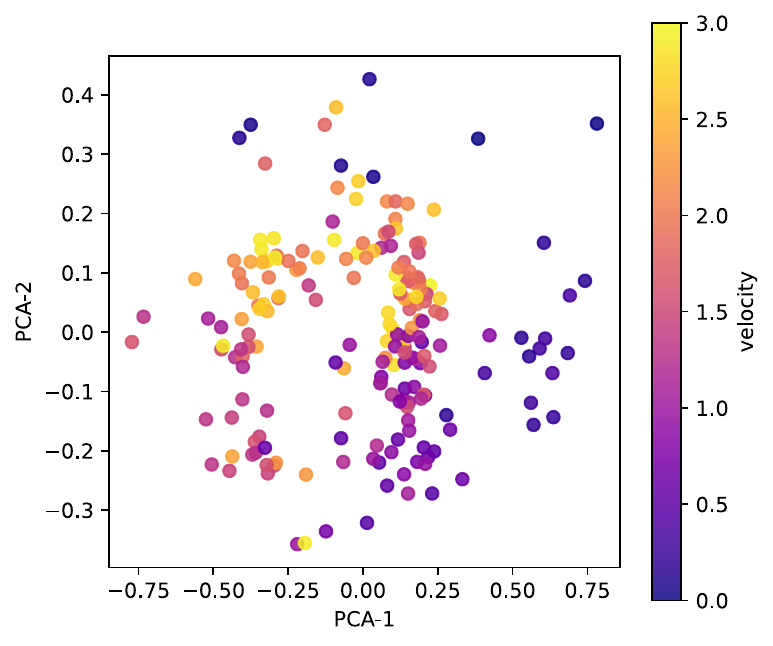} & 
        \includegraphics[width=\linewidth]{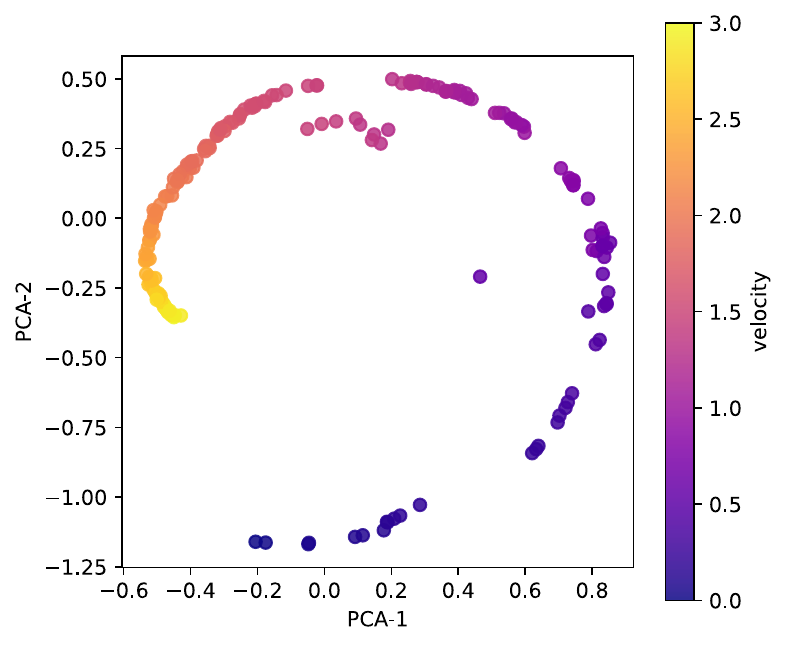} & 
        \includegraphics[width=\linewidth]{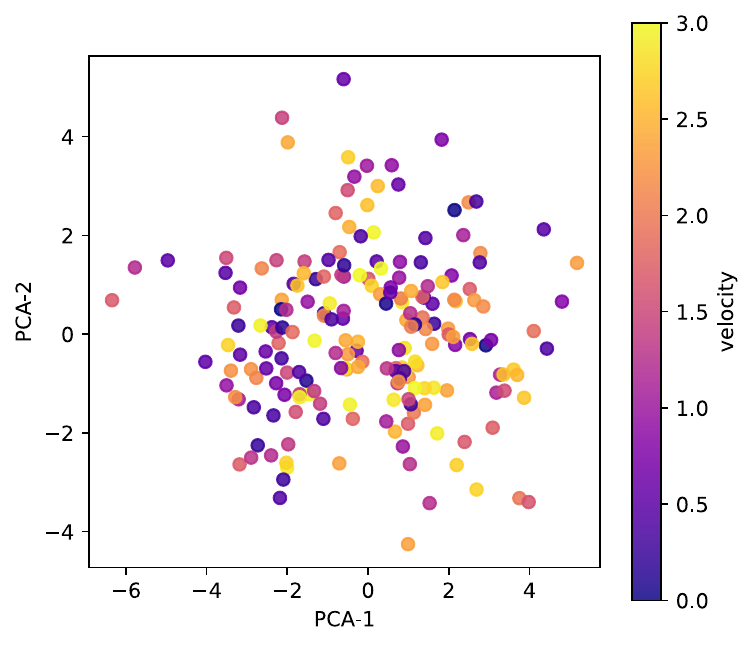} & 
        \includegraphics[width=\linewidth]{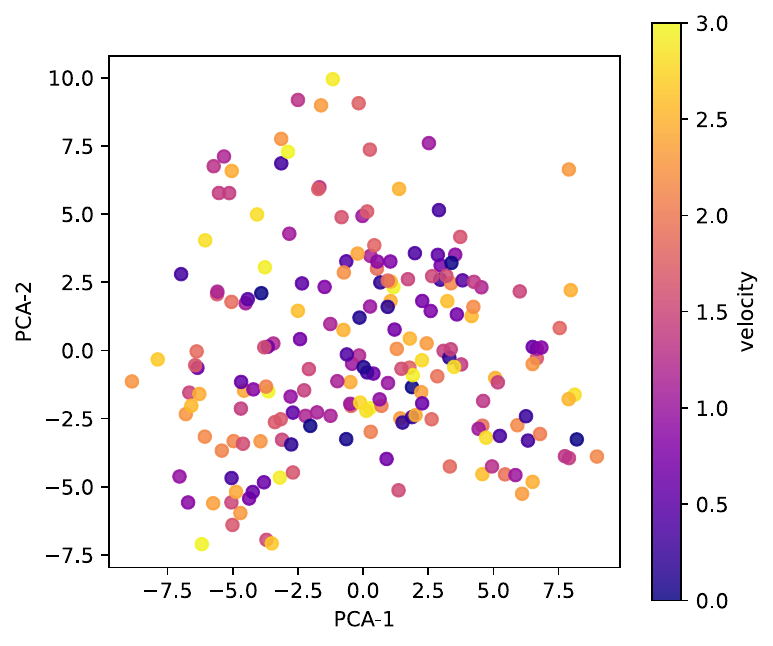} \\[1mm]

        \textbf{Run 3} & 
        \includegraphics[width=\linewidth]{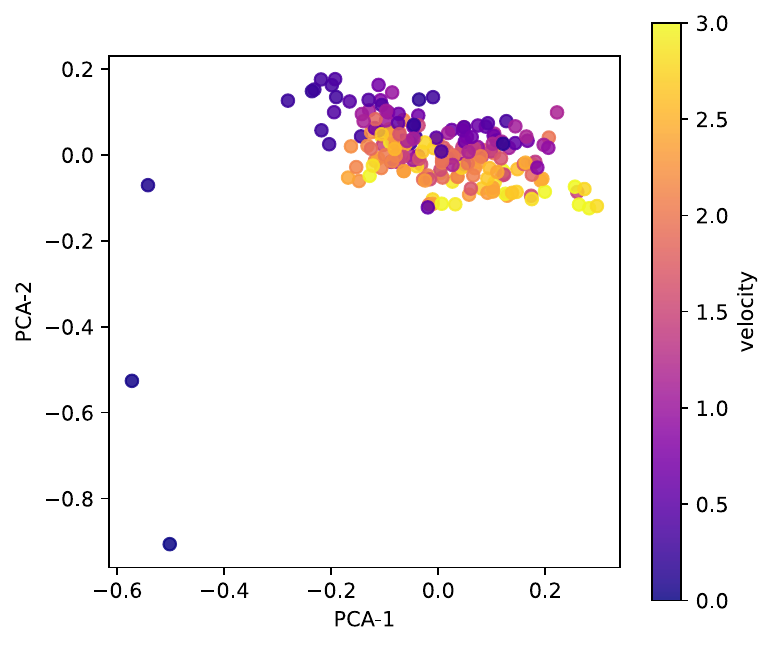} & 
        \includegraphics[width=\linewidth]{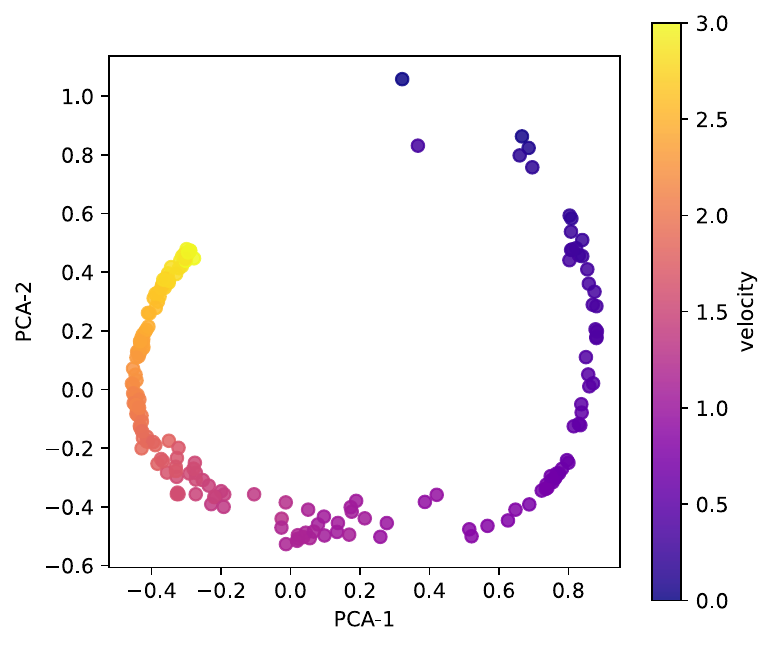} & 
        \includegraphics[width=\linewidth]{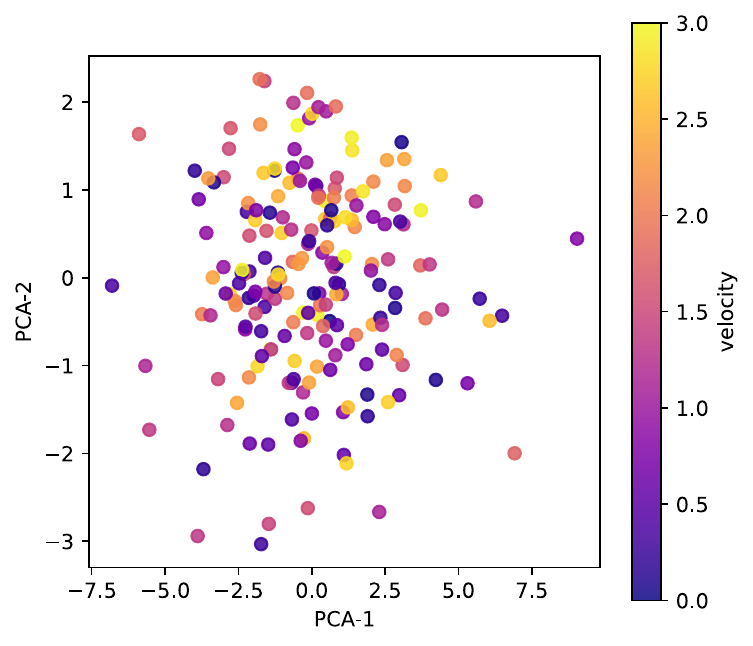} & 
        \includegraphics[width=\linewidth]{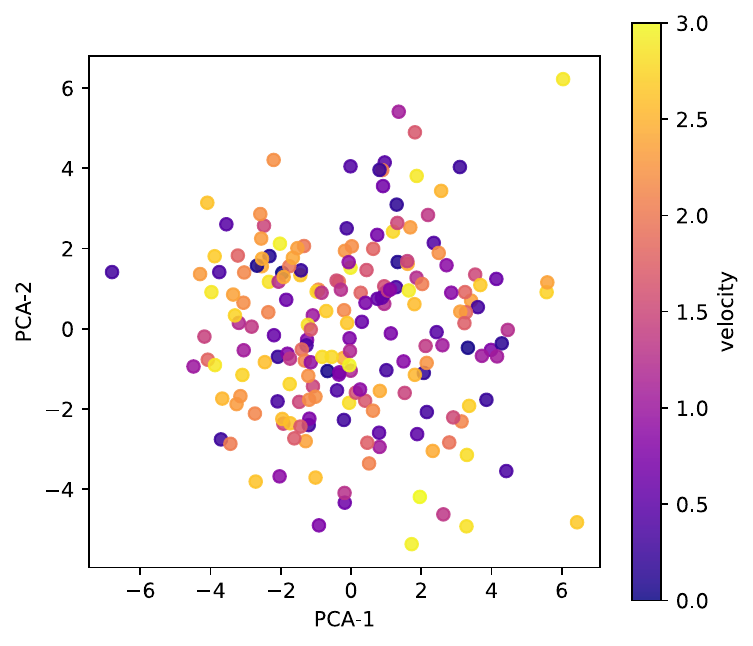} \\[1mm]

        \textbf{Run 4} & 
        \includegraphics[width=\linewidth]{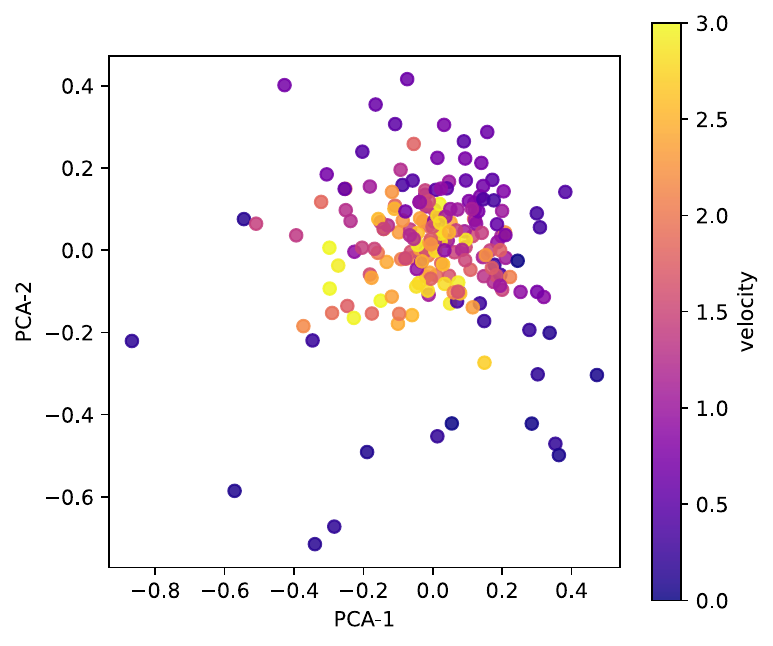} & 
        \includegraphics[width=\linewidth]{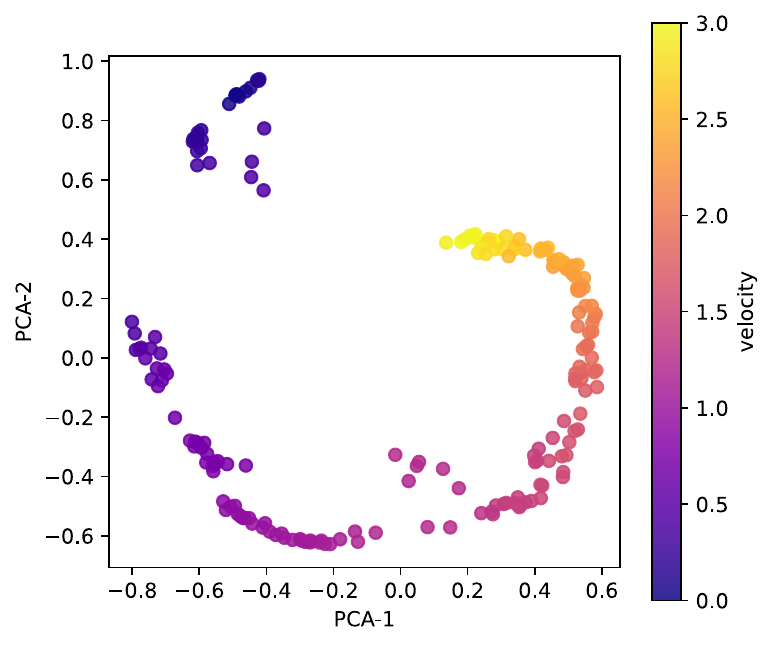} & 
        \includegraphics[width=\linewidth]{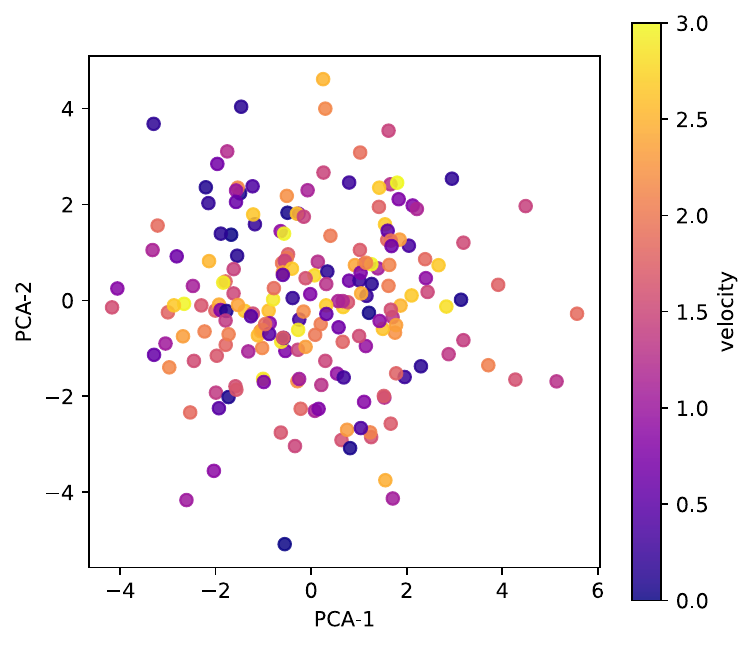} & 
        \includegraphics[width=\linewidth]{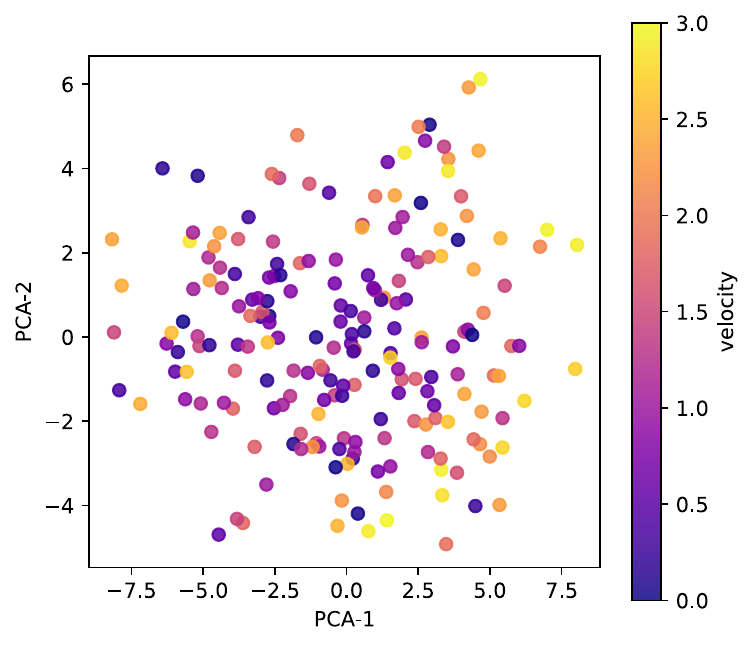} \\[1mm]

        \textbf{Run 5} & 
        \includegraphics[width=\linewidth]{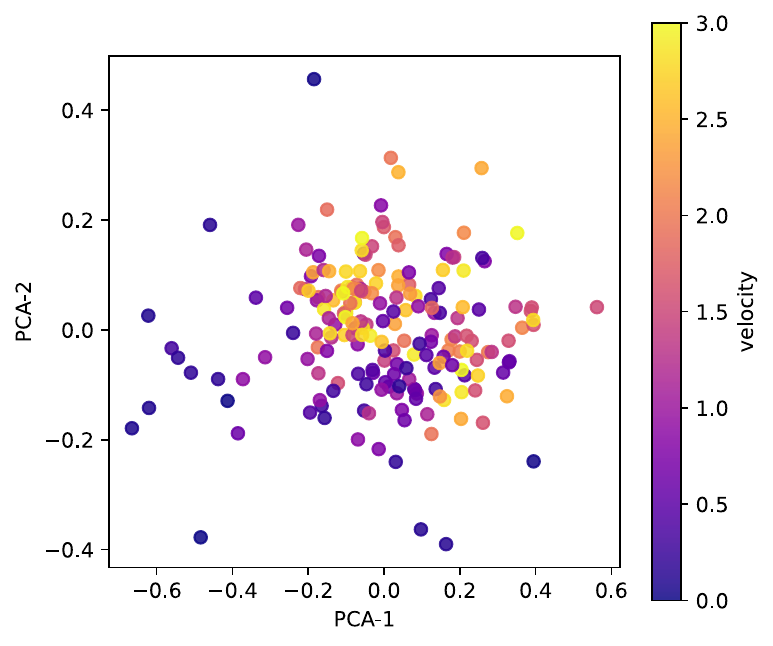} & 
        \includegraphics[width=\linewidth]{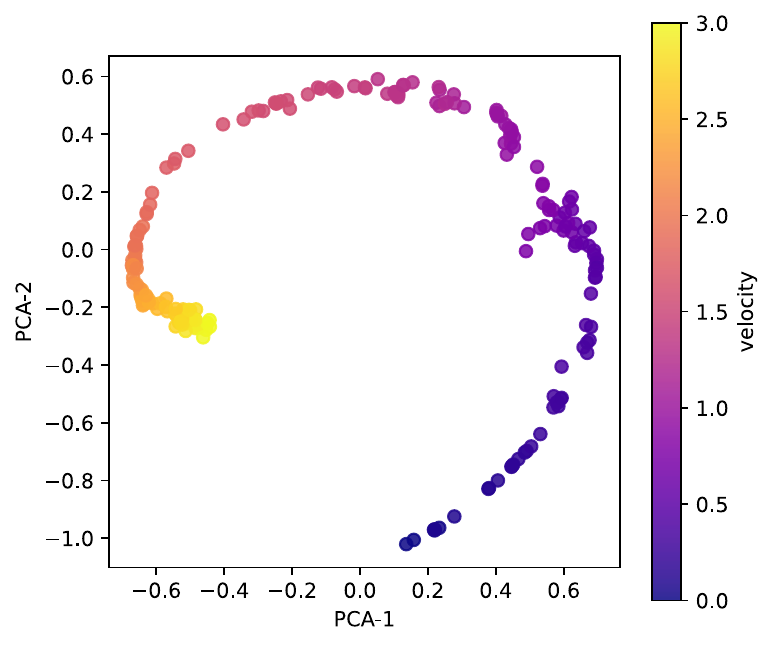} & 
        \includegraphics[width=\linewidth]{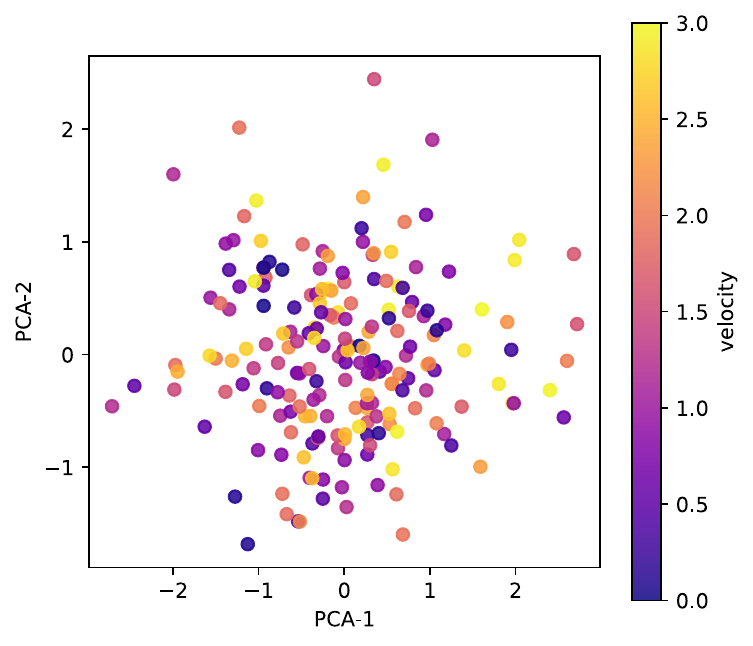} & 
        \includegraphics[width=\linewidth]{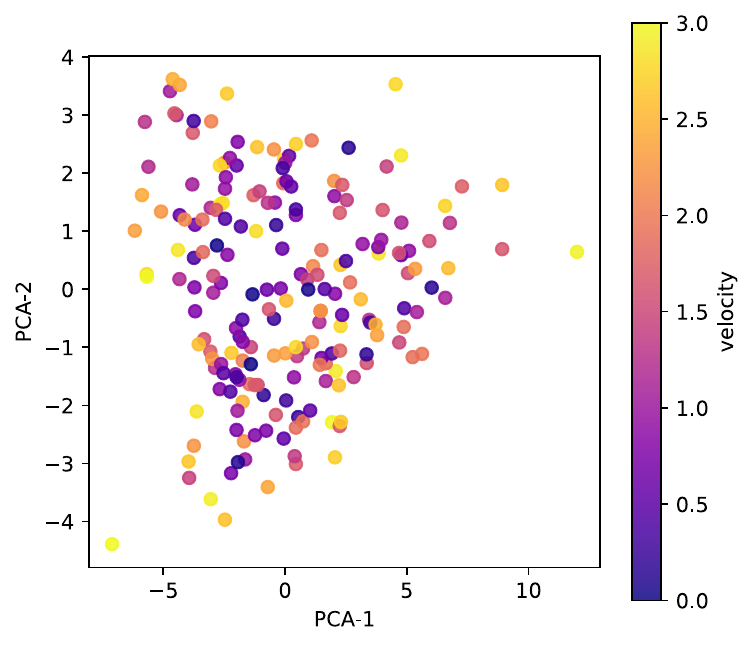} \\[1mm]
        \\ \hline 
    \end{tabular}

    \caption{PCA projections of task representations from \nom and TrMRL across five independent runs. These projections provide a qualitative view of task separation.}
    \label{fig:method_step_comparison_5rows_vline}
\end{figure}
We visualise the task representations of \nom and TrMRL on the \texttt{HalfCheetahVel} benchmark using 2D PCA across five independent runs; TrMRL is allocated a $10\times$ larger training-step budget for fairness. Even before training ($\text{step}=0$), the \nom projections show more apparent task separation than those of TrMRL at $5e6$ steps. After $5e5$ training steps, the \nom projections display a clear ordering of the continuous target velocities and the spherical pattern induced by the normalised representation function $f_{T,R}$. Across these qualitative examples, nearby target velocities generally remain close in the projected space, whereas more distinct velocities tend to separate. 

\newpage
\subsection{Sensitivity of Latent Task Dimensions}
\label{sensitivity}
{\color{black}
To demonstrate the sensitivity of \nom with respect to latent task dimensions $D_T$ and $D_R$, we perform hyperparameter search in ML10 on (1) $D_T = \left\{ 4, 8, 16, 32 \right\}$ while fixing $D_R = 256$ and (2) $D_R = \left\{32, 64, 128, 256, 512\right\}$ while fixing $D_T = 16$. We compare on (1) the success rate, (2) the predictive error in transitions, and (3) the predictive error in rewards.

\begin{figure}[h]
    \centering
    \begin{subfigure}{0.49\textwidth}
        \includegraphics[width=\textwidth]{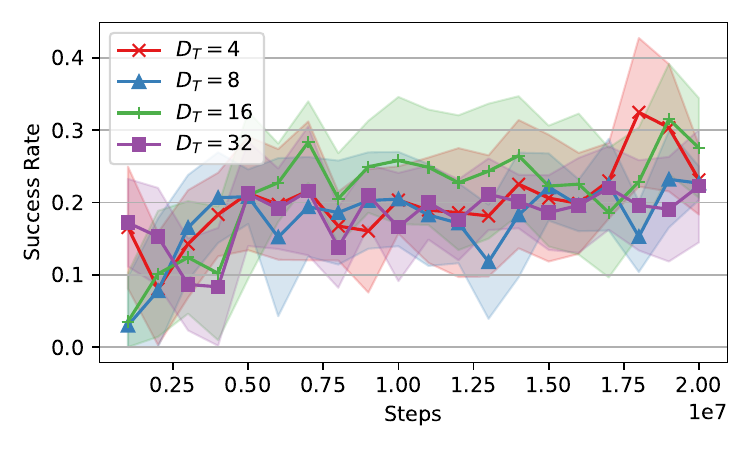}
        \caption{Success Rate (Fixed $D_R$)}
    \end{subfigure}
    ~
    \begin{subfigure}{0.49\textwidth}
        \includegraphics[width=\textwidth]{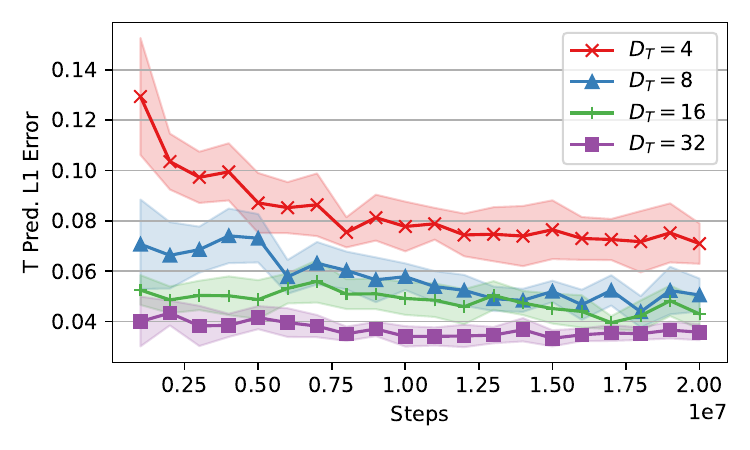}
        \caption{Pred. Error in Transitions (Fixed $D_R$)}
    \end{subfigure}
    \\
    \begin{subfigure}{0.49\textwidth}
        \includegraphics[width=\textwidth]{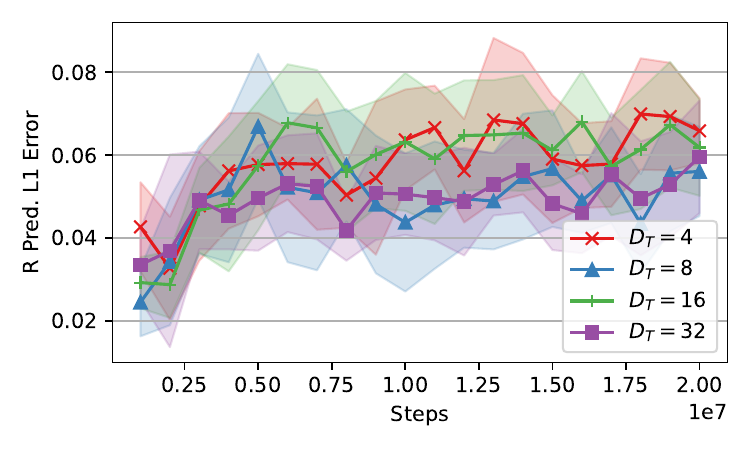}
        \caption{Pred. Error in Rewards (Fixed $D_R$)}
    \end{subfigure}
    ~
    \begin{subfigure}{0.49\textwidth}
        \includegraphics[width=\textwidth]{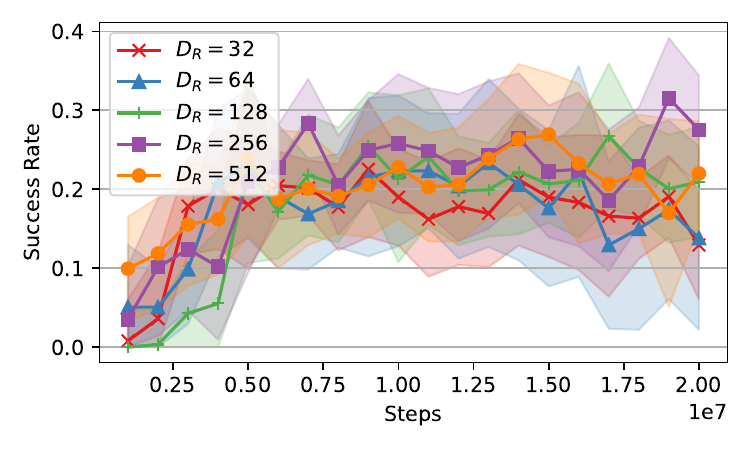}
        \caption{Success Rate (Fixed $D_T$)}
    \end{subfigure}
    \\
    \begin{subfigure}{0.49\textwidth}
        \includegraphics[width=\textwidth]{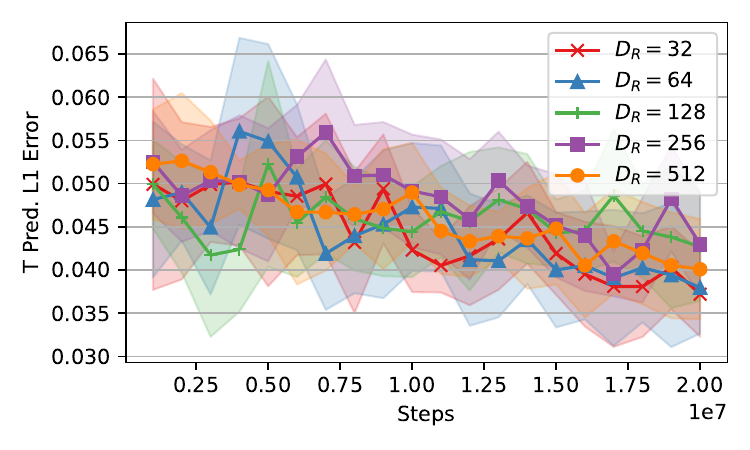}
        \caption{Pred. Error in Transitions (Fixed $D_T$)}
    \end{subfigure}
    ~
    \begin{subfigure}{0.49\textwidth}
        \includegraphics[width=\textwidth]{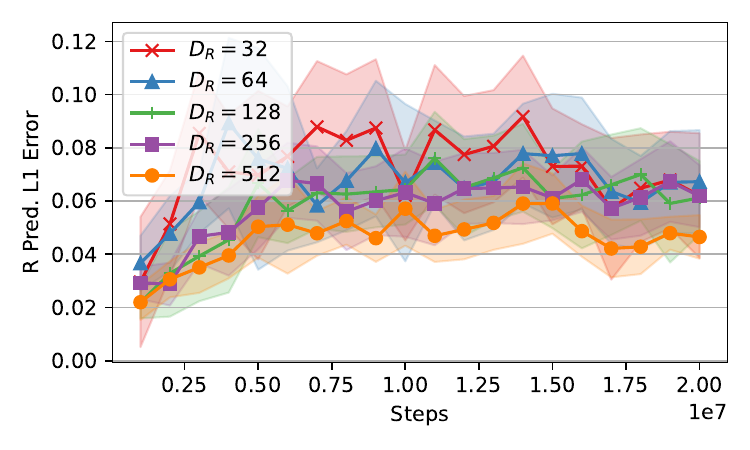}
        \caption{Pred. Error in Rewards (Fixed $D_T$)}
    \end{subfigure}
    \caption{Sensitivity analysis of latent task dimensions in \nom. }
\end{figure}
In general, \nom shows robust success rates with respect to latent task dimensions $D_T, D_R$, which can be inferred from Figure (a) and (d). Figure (b) and (f) indicate that as $D_T, D_R$ grows, the corresponding error reduces, offering an efficiency-accuracy trade-off. Figure (c) and (e) are sanity checks that have confirmed transitions do not affect rewards, and vice versa. Overall, $D_R$ are set to be much larger than $D_T$, as reward functions are generally much harder to learn, compared to transition function. 
}

\newpage
\subsection{Ablation Studies}
\label{ablation}

\label{cnorm}
\begin{figure}[h]
    \centering
    \begin{subfigure}{0.49\textwidth}
        \includegraphics[width=\textwidth]{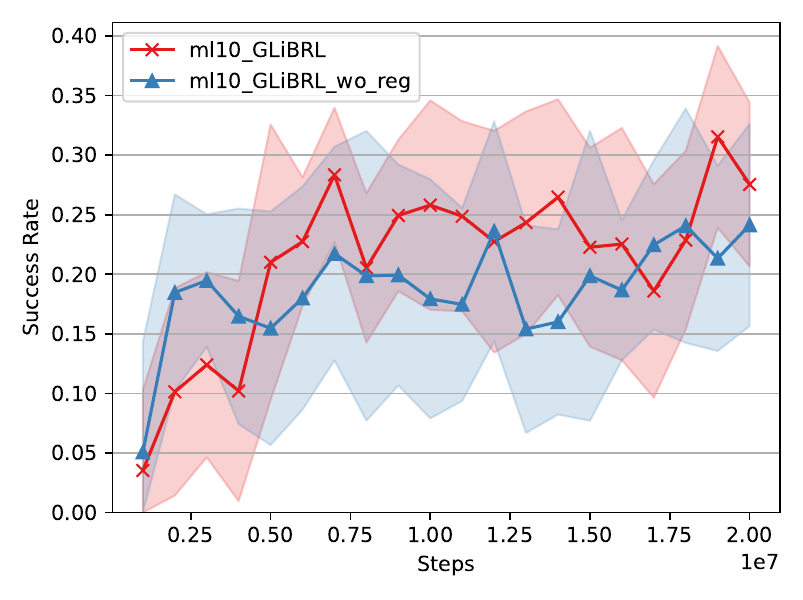}
        \caption{Success Rate}
    \end{subfigure}
    ~
    \begin{subfigure}{0.49\textwidth}
        \includegraphics[width=\textwidth]{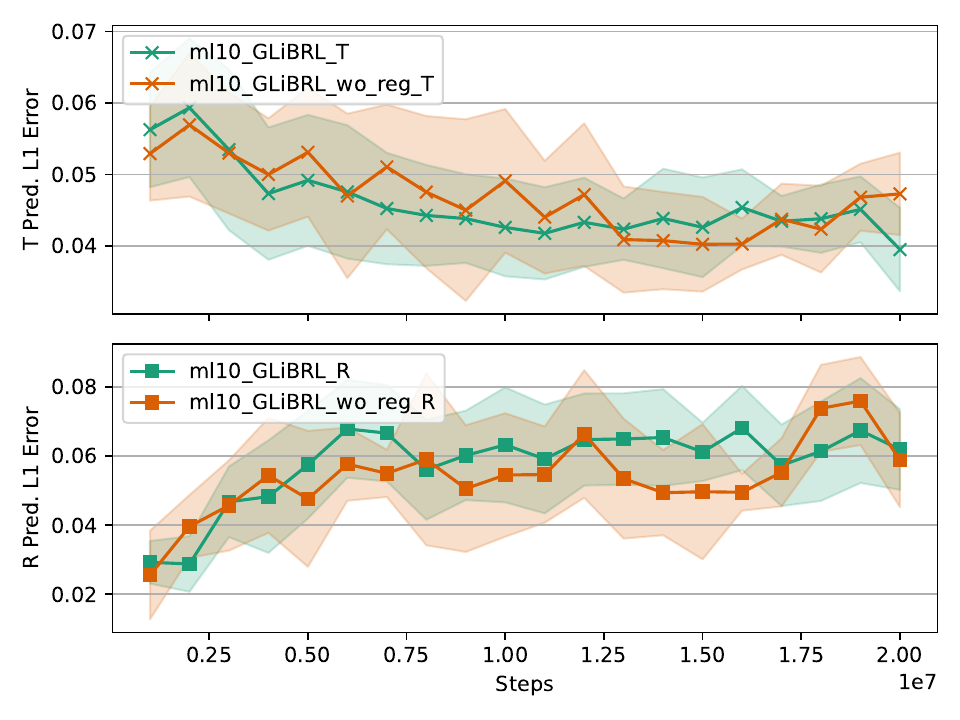}
        \caption{Predictive Error}
    \end{subfigure}
    \caption{Success rates and predictive errors of \nom with and without regularisation in ML10. }
\end{figure}
Running \nom and its variant without regularisation in the ML10 benchmark, we can tell the regularisations do not change the overall trend in both success rate and prediction error. However, as the number of training step increases, \nom begins to show in (a): higher success rates; in (b): lower predictive error in transitions. The higher predictive error in reward is due to the higher magnitude of reward from higher success rates in \nom.

\begin{figure}[h]
    \centering
    \begin{subfigure}{0.49\textwidth}
        \centering
        \includegraphics[width=\textwidth]{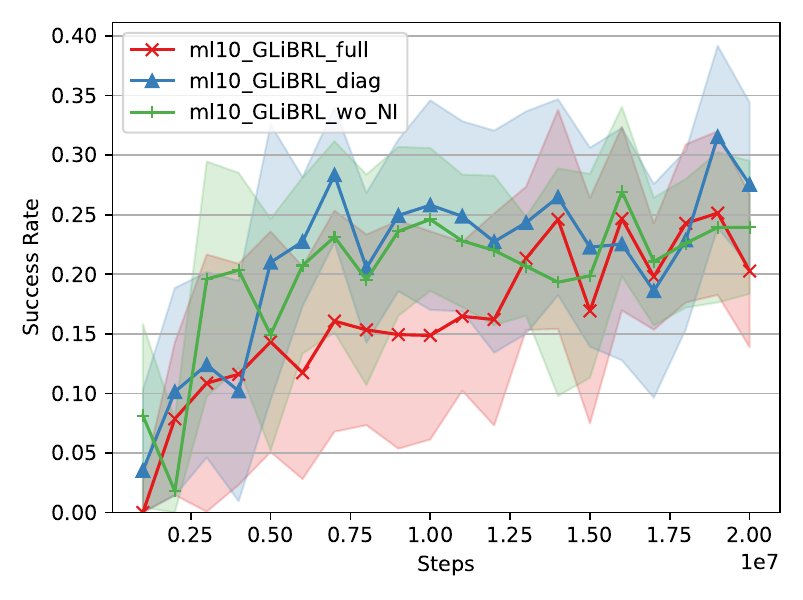}
    \end{subfigure}%
    ~
    \begin{subfigure}{0.49\textwidth}
        \centering
        \includegraphics[width=\textwidth]{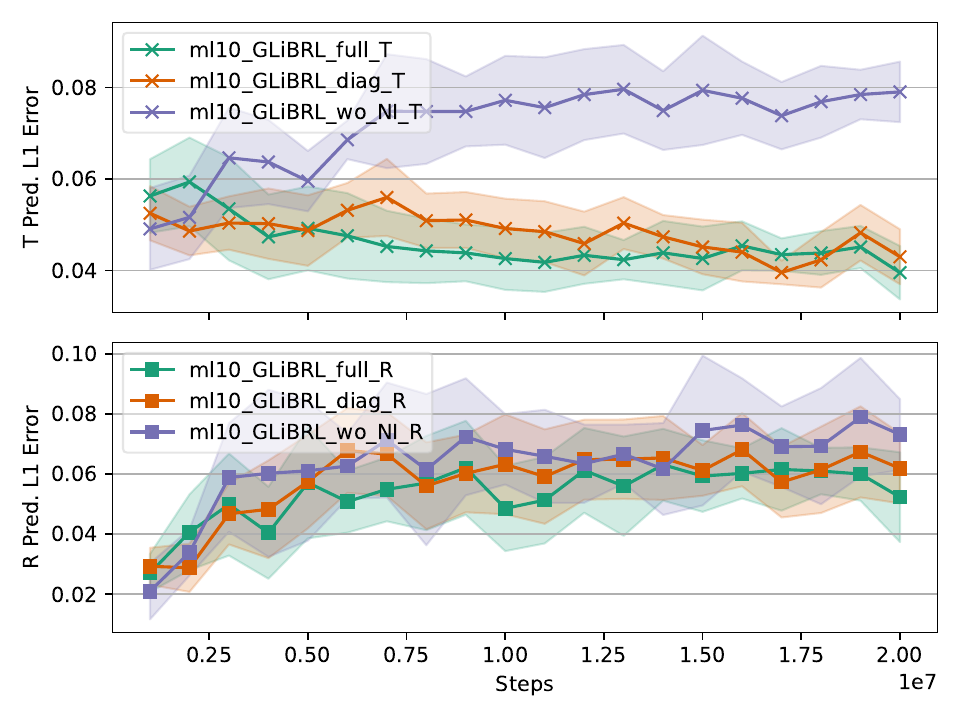}
    \end{subfigure} \\
    \begin{subfigure}{0.49\textwidth}
        \centering
        \includegraphics[width=\textwidth]{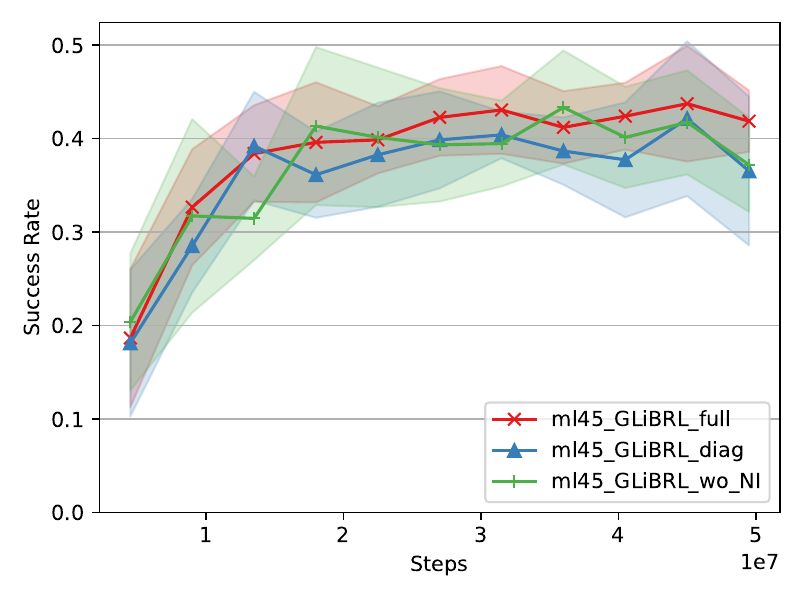}
    \end{subfigure}%
    ~
    \begin{subfigure}{0.49\textwidth}
        \centering
        \includegraphics[width=\textwidth]{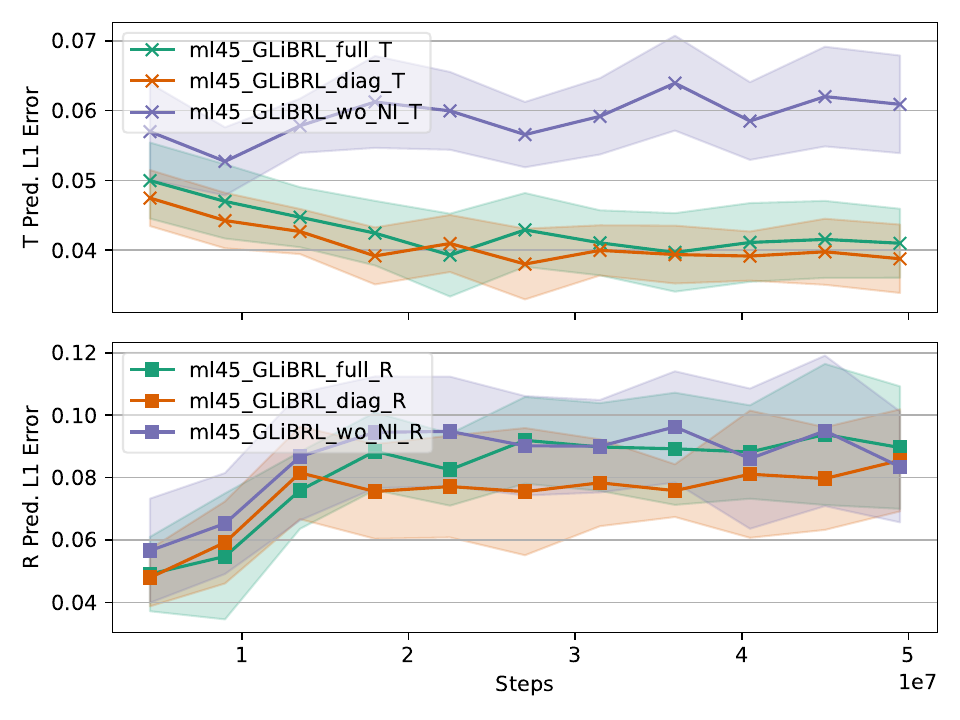}
    \end{subfigure}
    \caption{Success rates (Left) and predictive errors in transition and reward (Right) in ML10 and ML45, comparing \text{GLiBRL\_full}, \text{GLiBRL\_diag} and \text{GLiBRL\_wo\_NI}. \label{lossfig}}
\end{figure}

\nom can be viewed as a generalised, deep BRL version of \alpaca; whereas \alpaca  assumes the model noises $\sigmat = \SigmaT, \sigmar = \SigmaR$ are fully known a priori, \nom performs full Bayesian inference over them. Under the assumption of \alpaca, \Eqref{llhour} reduces to (see \Apref{alpaca})
\begin{align} 
    \labelAndRemember{tllh_simple}{
    \begin{split}
        \log_{\phit, \phir}(\mathcal{C}) = -\frac{1}{2}\sum_{i=1}^M \biggl( D_S\log |\xit'| &- \Tr \left(\SigmaT^{-1} \Mt'^{\text{T}} \xit' \Mt'\right)  \\
         &+ \log |\xir'| - \Tr \left(\SigmaR^{-1} \Mr'^{\text{T}} \xir' \Mr'\right)\biggr) +  \text{const.}
    \end{split}
    }
\end{align}
We studied whether performing Bayesian inference on model noises is necessary for learning accurate transition and reward models. We test \nom with full noise inference (\text{GLiBRL\_full}),  \nom with diagonal noise inference (\text{GLiBRL\_diag}) and \nom without noise inference (\text{GLiBRL\_wo\_NI}) with identical hyperparameters on ML10 and ML45 for 10 seeds. We have set $\Sigma_T=0.025$ and $\Sigma_R = 0.5$ to match the initial covariance of \text{GLiBRL\_full} and \text{GLiBRL\_diag}. The metrics being evaluated are success rates and $\mathcal{L}_1$ norms of prediction errors in both transitions (defined as $|\mathbf{S}' - \Ct\Mt'|_1$) and rewards (defined as $|\mathbf{r} - \Cr\Mr'|_1$).

As demonstrated in \Figref{lossfig}, both noise-inference variants substantially reduce transition prediction error relative to \text{GLiBRL\_wo\_NI}; the differences in reward error are smaller and partly reflect the increasing reward magnitude as policies become more successful. The increasing transition error of \text{GLiBRL\_wo\_NI} is undesirable because state magnitudes remain comparatively bounded. More accurate predictive models also make \text{GLiBRL\_diag} and \text{GLiBRL\_full}) better suited to model-based extensions that rely on imagined samples.

Interestingly, the relative behaviour of \text{GLiBRL\_diag} and \text{GLiBRL\_full} reverses between ML10 and ML45. On ML10, \text{GLiBRL\_diag} achieves a higher success rate despite having a higher predictive error, whereas on ML45, the same pattern is observed for \text{GLiBRL\_full}. A plausible explanation is a bias--variance trade-off in noise modelling. The greater task diversity of ML45 makes correlations between state dimensions more informative, allowing the full-covariance marginal likelihood to learn more useful task representations. On the less diverse ML10 benchmark, modelling these correlations may instead increase estimation and optimisation variance or fit spurious correlations, while the diagonal formulation acts as an effective regulariser. The predictive-error ordering is likely a consequence of the resulting policies: the more successful variant reaches a broader range of contact-rich states and receives larger rewards, making its on-policy samples harder to predict. 

\subsection{Marginal Log-likelihood of Normal-Normal}
\label{alpaca}
We prove \Eqref{tllh_simple} has the following form
\begin{align*}
    \recallLabel{tllh_simple}
\end{align*}
\subsection*{Proof:}
The distributions without inferring on the noise are listed as follows:

\textbf{Likelihood:}
\begin{align}
    p_{\phit}(\mathbf{S}_i'|\mathbf{S}_i, \mathbf{A}_i, \thetat) &= \mathcal{MN}\left(\mathbf{S}_i'|\Ct \mut, \mathbf{I}_N,  \SigmaT \in \mathbb{R}^{D_S \times D_S}\right) \\
    p_{\phir}(\mathbf{r}_i|\mathbf{S}_i, \mathbf{A}_i, \mathbf{S}_i', \thetar) &= \mathcal{MN}\left(\mathbf{r}_i|\Cr \mur, \mathbf{I}_N, \SigmaR \in \mathbb{R}^{1 \times 1} \right)
\end{align}

\textbf{Prior:}
\begin{align}
    \begin{split}
         p(\thetat) = p(\mut) =& \mathcal{MN}(\mut|\Mt \in \mathbb{R}^{D_T \times D_S}, \xit^{-1} \in \mathbb{R}^{D_T \times D_T}, \SigmaT )
    \end{split}\\
   \begin{split}
        p(\thetar) = p(\mur) =& \mathcal{MN}(\mur|\Mr\in \mathbb{R}^{D_R \times 1}, \xir^{-1}\in \mathbb{R}^{D_R \times D_R}, \SigmaR) 
   \end{split}
\end{align}

\textbf{Posterior:}
\begin{align}
    p_{\phit}(\thetat|\mathbf{S}_i, \mathbf{A}_i, \mathbf{S}_i') &= \mathcal{MN}(\mut|\Mt', \xit'^{-1}, \SigmaT)  \\
    p_{\phir}(\thetar|\mathbf{S}_i, \mathbf{A}_i, \mathbf{S}_i', \mathbf{r}) &= \mathcal{MN}(\mur|\Mr', \xir'^{-1}, \SigmaR) 
\end{align}
where
\begin{align}
    \begin{split}
         \Mt' &= {\xit'}^{-1} \left[\Ct^\text{T}\mathbf{S}' + \xit \Mt \right]\\
        {\xit'} &=\Ct^\text{T}\Ct + \xit \\
   \end{split}
    \begin{split}
        \Mr' &= {\xir'}^{-1} \left[\Cr^\text{T}\mathbf{r} +  \xir \Mr \right]\\
        {\xir'} &= \Cr^\text{T}\Cr + \xir\\
    \end{split}
\end{align}

Similar to \Apref{llhproof},
\begin{equation}
    \begin{split}
        p_{\phit} ( \mathbf{S}_i'|\mathbf{S}_i, \mathbf{A}_i) = \frac{p_{\phit}(\thetat,  \mathbf{S}_i'|\mathbf{S}_i, \mathbf{A}_i)}{p_{\phit}(\thetat|\mathbf{S}_i, \mathbf{A}_i, \mathbf{S}_i')}
    \end{split}
\end{equation}
As
\begin{align}
    \begin{split}
        &p_{\phit}(\thetat,  \mathbf{S}_i'|\mathbf{S}_i, \mathbf{A}_i) = \frac{|\xit|^{D_S/2}|\SigmaT|^{-D_T/2}}{\sqrt{(2\pi)^{D_TD_S}}} \cdot \exp{\left[ -\frac{1}{2}\Tr\left( \SigmaT^{-1} (\mut - \Mt)^\text{T} \xit  (\mut - \Mt) \right) \right]}\\
        &\qquad\qquad\qquad\qquad\qquad \cdot \frac{|\SigmaT|^{-N/2}}{\sqrt{(2\pi)^{ND_S}}} \cdot \exp{\left[ -\frac{1}{2} \Tr\left( \SigmaT^{-1} (\mathbf{S}_i' - \Ct\mut)^{\text{T}} (\mathbf{S}_i' - \Ct\mut)\right) \right]}
    \end{split}
\end{align}
And
\begin{align}
   p_{\phit}(\thetat|\mathbf{S}_i, \mathbf{A}_i, \mathbf{S}_i') = \frac{|\xit'|^{D_S/2}|\SigmaT|^{-D_T/2}}{\sqrt{(2\pi)^{D_TD_S}}} \cdot \exp{\left[ -\frac{1}{2}\Tr\left( \SigmaT^{-1} (\mut - \Mt')^\text{T} \xit'  (\mut - \Mt') \right) \right]}
\end{align}
Hence
\begin{align}
\begin{split}
        p_{\phit} ( \mathbf{S}_i'|\mathbf{S}_i, \mathbf{A}_i) &\propto |\xit'|^{-D_S/2} \cdot \exp\left[ -\frac{1}{2} \Tr\left( -\SigmaT^{-1}\Mt'^{\text{T}} \xit' \Mt' \right) \right]
\end{split}\\
\begin{split}
    \log p_{\phit} ( \mathbf{S}_i'|\mathbf{S}_i, \mathbf{A}_i) &= -\frac{D_S}{2}\log |\xit'| + \frac{1}{2}  \Tr\left(\SigmaT^{-1}\Mt'^{\text{T}} \xit' \Mt' \right) + \text{const.}  
\end{split}
\end{align}
Similarly,
\begin{align}
    \begin{split}
    \log p_{\phir} ( \mathbf{r}_i|\mathbf{S}_i, \mathbf{A}_i, \mathbf{S}'_i) &= -\frac{1}{2} \log |\xir'| + \frac{1}{2}  \Tr\left(\SigmaR^{-1}\Mr'^{\text{T}} \xir' \Mr' \right)  + \text{const.}
\end{split}
\end{align}
Following \Apref{llhproof},
\begin{align*}
    \recallLabel{tllh_simple}
\end{align*}

\newpage
\subsection{Hyperparameters and Runtime}
\label{hyperparam}
We list all hyperparameters of \nom in the following tables. 

In \texttt{HalfCheetahDir}:
\begin{table*}[h]
    \centering
    \setstretch{1.4}
    \begin{tabular}{l|l|l|l}
        \toprule
         \multicolumn{1}{l}{\textbf{Name}} &  \multicolumn{1}{l}{\textbf{Value}} &  \multicolumn{1}{l}{\textbf{Name}} &  \multicolumn{1}{l}{\textbf{Value}} \\
         \cmidrule{1-4}
         \textbf{policy\_learner} & SAC & \textbf{feat\_out\_activation} & True\\
         \textbf{critic\_layer} & [256, 256] & \textbf{t\_mix\_layers} & [64, 32] \\ \textbf{actor\_layer} & [256, 256, 256] & \textbf{t\_mix\_layernorm} & True\\
         \textbf{policy\_activation} & Tanh & \textbf{t\_mix\_out\_activation} & False \\
         \textbf{policy\_optimiser} & Adam & \textbf{r\_mix\_layers} & [128, 64]\\
         \textbf{actor\_lr/critic\_lr} & 3e-4/4e-4 & \textbf{r\_mix\_layernorm} & True \\
         \textbf{policy\_max\_norm} & 10 & \textbf{r\_mix\_out\_activation} & False \\
         \textbf{policy\_weight\_init} & Xavier & \textbf{t\_reg\_coef} & 5e-3 \\
         \textbf{policy\_bias\_init} & 0 & \textbf{r\_reg\_coef} & 1e-3\\
         \textbf{actor\_logstd\_range} & $[\ln 10^{-6}, \ln 2]$ & \textbf{model\_activation} & ReLU\\
         \textbf{tau} & 5e-3 & \textbf{model\_optimiser} & Adam \\
         \textbf{actor\_grad\_steps} & 500 & \textbf{model\_lr} & 2e-4\\
         \textbf{critic\_grad\_steps} & 500 & \textbf{model\_opt\_max\_norm} & None   \\
         \textbf{policy\_update\_freq} & 200 &  \textbf{model\_grad\_epochs} & 1 \\
         \textbf{init\_alpha} & 1 &\textbf{model\_grad\_steps} & 1 \\
         \textbf{alpha\_lr} & 3e-4 & \textbf{init\_mt} & zeros \\
         \textbf{policy\_bs\_range}& [1, 200] &  \textbf{init\_mr} & zeros\\
         \textbf{s\_feat\_layers} & [64, 32] &  \textbf{init\_xit} & ones\\
         \textbf{s\_feat\_outdim} & 32 & \textbf{init\_xir} & ones \\
         \textbf{s\_feat\_layernorm} & False & \textbf{init\_omegat} & ones  \\
         \textbf{a\_feat\_layers} & [32, 16] & \textbf{init\_omegar} & ones \\
         \textbf{a\_feat\_outdim} & 16 & \textbf{init\_nut} & 22 \\
         \textbf{a\_feat\_layernorm} & False & \textbf{init\_nur} & 2 \\
         \bottomrule
    \end{tabular}
\end{table*}
\newpage
In \texttt{HalfCheetahVel}:
\begin{table*}[h]
    \centering
    \setstretch{1.4}
    \begin{tabular}{l|l|l|l}
        \toprule
         \multicolumn{1}{l}{\textbf{Name}} &  \multicolumn{1}{l}{\textbf{Value}} &  \multicolumn{1}{l}{\textbf{Name}} &  \multicolumn{1}{l}{\textbf{Value}} \\
         \cmidrule{1-4}
         \textbf{policy\_learner} & SAC & \textbf{feat\_out\_activation} & True\\
         \textbf{policy\_layers} & [256, 256] & \textbf{t\_mix\_layers} & [64, 32] \\ \textbf{a\_feat\_out\_activation} & True & \textbf{t\_mix\_layernorm} & True\\
         \textbf{policy\_activation} & Tanh & \textbf{t\_mix\_out\_activation} & False \\
         \textbf{policy\_optimiser} & Adam & \textbf{r\_mix\_layers} & [128, 64]\\
         \textbf{actor\_lr/critic\_lr} & 3e-4/4e-4 & \textbf{r\_mix\_layernorm} & True \\
         \textbf{policy\_max\_norm} & 10 & \textbf{r\_mix\_out\_activation} & False \\
         \textbf{policy\_weight\_init} & Xavier & \textbf{t\_reg\_coef} & 5e-3 \\
         \textbf{policy\_bias\_init} & 0 & \textbf{r\_reg\_coef} & 1e-3\\
         \textbf{actor\_logstd\_range} & $[\ln 10^{-6}, \ln 2]$ & \textbf{model\_activation} & ReLU\\
         \textbf{tau} & 1e-2 & \textbf{model\_optimiser} & Adam \\
         \textbf{actor\_grad\_steps} & 2000 & \textbf{model\_lr} & 2e-4\\
         \textbf{critic\_grad\_steps} & 2000 & \textbf{model\_opt\_max\_norm} & None   \\
         \textbf{policy\_update\_freq} & 200 &  \textbf{model\_grad\_epochs} & 1 \\
         \textbf{init\_alpha} & 0.2 &\textbf{model\_grad\_steps} & 50 \\
         \textbf{alpha\_lr} & 3e-4 & \textbf{init\_mt} & zeros \\
         \textbf{policy\_bs\_range}& [1, 200] &  \textbf{init\_mr} & zeros\\
         \textbf{s\_feat\_layers} & [64, 32] &  \textbf{init\_xit} & $\mathbf{I}$ \\
         \textbf{s\_feat\_outdim} & 32 & \textbf{init\_xir} & $\mathbf{I}$  \\
         \textbf{s\_feat\_layernorm} & False & \textbf{init\_omegat} & $\mathbf{I}$  \\
         \textbf{a\_feat\_layers} & [32, 16] & \textbf{init\_omegar} & $\mathbf{I}$  \\
         \textbf{a\_feat\_outdim} & 16 & \textbf{init\_nut} & 22 \\
         \textbf{a\_feat\_layernorm} & False & \textbf{init\_nur} & 2 \\
         \bottomrule
    \end{tabular}
\end{table*}
\newpage
In \texttt{AntDir}:
\begin{table*}[h]
    \centering
    \setstretch{1.4}
    \begin{tabular}{l|l|l|l}
        \toprule
         \multicolumn{1}{l}{\textbf{Name}} &  \multicolumn{1}{l}{\textbf{Value}} &  \multicolumn{1}{l}{\textbf{Name}} &  \multicolumn{1}{l}{\textbf{Value}} \\
         \cmidrule{1-4}
         \textbf{policy\_learner} & SAC & \textbf{feat\_out\_activation} & True\\
         \textbf{policy\_layers} & [256, 256] & \textbf{t\_mix\_layers} & [64, 32] \\ \textbf{a\_feat\_out\_activation} & True & \textbf{t\_mix\_layernorm} & True\\
         \textbf{policy\_activation} & Tanh & \textbf{t\_mix\_out\_activation} & False \\
         \textbf{policy\_optimiser} & Adam & \textbf{r\_mix\_layers} & [128, 64]\\
         \textbf{actor\_lr/critic\_lr} & 3e-4/4e-4 & \textbf{r\_mix\_layernorm} & True \\
         \textbf{policy\_max\_norm} & 1 & \textbf{r\_mix\_out\_activation} & False \\
         \textbf{policy\_weight\_init} & Xavier & \textbf{t\_reg\_coef} & 5e-3 \\
         \textbf{policy\_bias\_init} & 0 & \textbf{r\_reg\_coef} & 1e-3\\
         \textbf{actor\_logstd\_range} & $[\ln 10^{-6}, \ln 2]$ & \textbf{model\_activation} & ReLU\\
         \textbf{tau} & 1e-2 & \textbf{model\_optimiser} & Adam \\
         \textbf{actor\_grad\_steps} & 1000 & \textbf{model\_lr} & 2e-4\\
         \textbf{critic\_grad\_steps} & 2000 & \textbf{model\_opt\_max\_norm} & None   \\
         \textbf{policy\_update\_freq} & 400 &  \textbf{model\_grad\_epochs} & 1 \\
         \textbf{init\_alpha} & 0.1 &\textbf{model\_grad\_steps} & 5 \\
         \textbf{alpha\_lr} & 3e-4 & \textbf{init\_mt} & zeros \\
         \textbf{policy\_bs\_range}& [1, 200] &  \textbf{init\_mr} & zeros\\
         \textbf{s\_feat\_layers} & [64, 32] &  \textbf{init\_xit} & $\mathbf{I}$ \\
         \textbf{s\_feat\_outdim} & 32 & \textbf{init\_xir} & $\mathbf{I}$  \\
         \textbf{s\_feat\_layernorm} & False & \textbf{init\_omegat} & $\mathbf{I}$   \\
         \textbf{a\_feat\_layers} & [32, 16] & \textbf{init\_omegar} & $\mathbf{I}$  \\
         \textbf{a\_feat\_outdim} & 16 & \textbf{init\_nut} & 28 \\
         \textbf{a\_feat\_layernorm} & False & \textbf{init\_nur} & 2 \\
         \bottomrule
    \end{tabular}
\end{table*}

\newpage

We use the same hyperparameters for both ML10 and ML45.

\begin{table*}[h]
    \centering
    \setstretch{1.4}
    \begin{tabular}{l|l|l|l}
        \toprule
         \multicolumn{1}{l}{\textbf{Name}} &  \multicolumn{1}{l}{\textbf{Value}} &  \multicolumn{1}{l}{\textbf{Name}} &  \multicolumn{1}{l}{\textbf{Value}} \\
         \cmidrule{1-4}
         \textbf{policy\_learner} & PPO & \textbf{feat\_out\_activation} & True\\
         \textbf{policy\_layers} & [256, 256] & \textbf{t\_mix\_layers} & [64, 32] \\ \textbf{a\_feat\_out\_activation} & True & \textbf{t\_mix\_layernorm} & True\\
         \textbf{policy\_activation} & Tanh & \textbf{t\_mix\_out\_activation} & False \\
         \textbf{policy\_optimiser} & Adam & \textbf{r\_mix\_layers} & [128, 64]\\
         \textbf{policy\_lr} & 5e-4 & \textbf{r\_mix\_layernorm} & True \\
         \textbf{policy\_opt\_max\_norm} & N/A & \textbf{r\_mix\_out\_activation} & False \\
         \textbf{policy\_weight\_init} & Xavier & \textbf{t\_reg\_coef} & 5e-3 \\
         \textbf{policy\_bias\_init} & 0 & \textbf{r\_reg\_coef} & 1e-3\\
         \textbf{policy\_log\_std\_min} & $\ln 10^{-6}$ & \textbf{model\_activation} & ReLU\\
         \textbf{policy\_log\_std\_max} & $\ln 2$ & \textbf{model\_optimiser} & Adam \\
         \textbf{policy\_grad\_epochs} & 10 & \textbf{model\_lr} & 2e-4\\
         \textbf{policy\_grad\_steps} & 20 & \textbf{model\_opt\_max\_norm} & None   \\
         \textbf{ppo\_clip\_eps} & 0.5 &  \textbf{model\_grad\_epochs} & 1 \\
         \textbf{ppo\_gamma} & 0.99 &\textbf{model\_grad\_steps} & 20 \\
         \textbf{ppo\_gae\_lambda} & 0.95 & \textbf{init\_mt} & zeros \\
         \textbf{ppo\_entropy\_coef }& 5e-3 &  \textbf{init\_mr} & zeros\\
         \textbf{s\_feat\_layers} & [64, 32] &  \textbf{init\_xit} & $\mathbf{I}$ \\
         \textbf{s\_feat\_outdim} & 32 & \textbf{init\_xir} & $\mathbf{I}$  \\
         \textbf{s\_feat\_layernorm} & False & \textbf{init\_omegat} & $\mathbf{I}$   \\
         \textbf{a\_feat\_layers} & [32, 16] & \textbf{init\_omegar} & $\mathbf{I}$  \\
         \textbf{a\_feat\_outdim} & 16 & \textbf{init\_nut} & 40 \\
         \textbf{a\_feat\_layernorm} & False & \textbf{init\_nur} & 2 \\
         \bottomrule
    \end{tabular}
\end{table*}

\vspace{2em}

The runtime of \nom does not vary much with changes in $D_T$ and $D_R$. In the following table, we show the average runtime in minutes for each single run with different $D_T$ and $D_R$.

\begin{table*}[h]
    \centering
    \setstretch{1.4}
    \begin{tabular}{c|c|c}
        \toprule
        $D_T$ & $D_R$ & Avg. Runtime (min) \\
        \cmidrule{1-3}
         16 & 32 & 72 \\
         16 & 512 & 80 \\
         4 & 256 & 75 \\
         32 & 256 & 77 \\
         \bottomrule
    \end{tabular}
\end{table*}

\newpage

%% file: sections/theory_fixed.tex
\subsection{Kernel Correspondence of Task Representations}
\label{theory}
Let $\mathcal{C}^{\mathcal{M}_i}=(\mathbf{S}_i,\mathbf{A}_i,\mathbf{S}'_i,\mathbf{r}_i)$ contain $N_i$ samples $\mathbf{c}_n^i=(s_n^i,a_n^i,{s_n^i}',r_n^i)$. We take the shared prior means to be zero. The shared row precisions can be absorbed by the reparameterisations $\widetilde{\Ct}=\Ct\xit^{-1/2}$, $\widetilde{\Mt}=\xit^{1/2}\Mt$ and $\widetilde{\Cr}=\Cr\xir^{-1/2}$, $\widetilde{\Mr}=\xir^{1/2}\Mr$, which preserve $\Ct\Mt$ and $\Cr\Mr$ and give identity row precisions. We may therefore set $\xit=\xir=\mathbf{I}$ without loss of generality and drop the tildes. We also suppress the task index until comparing tasks. The identities for $f_1,f_2,f_4$, and $f_5$ in \Apref{beliefrep} use three blocks: $\Mt'$, $\Mr'$, and $\mathrm{tril}(\Mt'{\Mt'}^{\mathrm{T}})$. We first express these blocks as kernel embeddings and discuss $f_3$ and $f_6$ separately.

\paragraph{Linear transition mean.} From \Eqref{post} and the push-through identity,
\begin{align}
    \Mt'&=\Ct^{\mathrm{T}}\mathbf{U}_T
    =\sum_{n=1}^{N}\boldsymbol{\psi}_T(\mathbf{c}_n)\mathbf{u}_{T,n}^{\mathrm{T}},
    &\mathbf{U}_T&=(\mathbf{I}+\Ct\Ct^{\mathrm{T}})^{-1}\mathbf{S}',
\end{align}
where $\mathbf{u}_{T,n}^{\mathrm{T}}$ is the $n$-th row of $\mathbf{U}_T$ and $\boldsymbol{\psi}_T(\mathbf{c}_n)=\phit(s_n,a_n)^{\mathrm{T}}$. Define $k_T(\mathbf{c},\mathbf{d})=\boldsymbol{\psi}_T(\mathbf{c})^{\mathrm{T}}\boldsymbol{\psi}_T(\mathbf{d})$ and, for each output dimension $\ell$,
\begin{align}
    \eta_{T,\ell}=\sum_{n=1}^{N}u_{T,n,\ell}\,\delta_{\mathbf{c}_n}.
\end{align}
Identifying the RKHS $\mathcal{H}_{k_T}$ with its coefficient space gives $\mu_{k_T}(\eta_{T,\ell})=\sum_n u_{T,n,\ell}\boldsymbol{\psi}_T(\mathbf{c}_n)$. Thus, in the direct sum $\mathcal{H}_T=\bigoplus_{\ell=1}^{D_S}\mathcal{H}_{k_T}$,
\begin{align}
    \mu_T:=\bigoplus_{\ell=1}^{D_S}\mu_{k_T}(\eta_{T,\ell})\equiv\Mt',
\end{align}
where the direct-sum norm is the Frobenius norm of $\Mt'$.

\paragraph{Reward mean.} The scalar-output analogue uses $\mathbf{U}_R=(\mathbf{I}+\Cr\Cr^{\mathrm{T}})^{-1}\mathbf{r}$, its $n$-th entry $u_{R,n}$, and $\boldsymbol{\psi}_R(\mathbf{c}_n)=\phir(s_n,a_n,s'_n)^{\mathrm{T}}$. With
\begin{align}
    k_R(\mathbf{c},\mathbf{d})&=\boldsymbol{\psi}_R(\mathbf{c})^{\mathrm{T}}\boldsymbol{\psi}_R(\mathbf{d}),
    &\eta_R&=\sum_{n=1}^{N}u_{R,n}\delta_{\mathbf{c}_n},
\end{align}
the same argument gives $\mu_{k_R}(\eta_R)=\sum_n u_{R,n}\boldsymbol{\psi}_R(\mathbf{c}_n)=\Mr'$.

\paragraph{Quadratic transition block.} Set $\beta_{mn}=\mathbf{u}_{T,m}^{\mathrm{T}}\mathbf{u}_{T,n}$ and $\mathbf{h}_T(\mathbf{c}_m,\mathbf{c}_n)=\mathrm{tril}(\boldsymbol{\psi}_T(\mathbf{c}_m)\boldsymbol{\psi}_T(\mathbf{c}_n)^{\mathrm{T}})$. The pair kernel and signed measure
\begin{align}
    k_{TT}((\mathbf{c},\mathbf{d}),(\mathbf{c}',\mathbf{d}'))
    &=\mathbf{h}_T(\mathbf{c},\mathbf{d})^{\mathrm{T}}\mathbf{h}_T(\mathbf{c}',\mathbf{d}'),
    &\eta_{TT}&=\sum_{m,n=1}^{N}\beta_{mn}\delta_{(\mathbf{c}_m,\mathbf{c}_n)}
\end{align}
therefore satisfy
\begin{align}
    \mu_{k_{TT}}(\eta_{TT})
    =\sum_{m,n=1}^{N}\beta_{mn}\mathbf{h}_T(\mathbf{c}_m,\mathbf{c}_n)
    =\mathrm{tril}(\Mt'{\Mt'}^{\mathrm{T}}).
\end{align}
Thus $k_T$ and $k_R$ act on individual samples, whereas $k_{TT}$ acts on ordered sample pairs; all three are positive semidefinite by construction.

For task $i$, write $\eta_{T,\ell}^i$, $\eta_R^i$, and $\eta_{TT}^i$ for these scalar measures and define
\begin{align}
    \mu_{k_T}^i&:=\bigoplus_{\ell=1}^{D_S}\mu_{k_T}(\eta_{T,\ell}^i)\equiv{{\Mt^i}'},\\
    \mu_{k_R}^i&:=\mu_{k_R}(\eta_R^i)={{\Mr^i}'},\qquad
    \mu_{k_{TT}}^i:=\mu_{k_{TT}}(\eta_{TT}^i)=\mathrm{tril}({\Mt^i}'{{\Mt^i}'}^{\mathrm{T}}).
\end{align}
For any scalar signed measures, the standard bilinear identity
\begin{align}
    \left\langle\mu_k(\eta^1),\mu_k(\eta^2)\right\rangle
    =\sum_{a,b}w_a^1w_b^2\,k(\mathbf{z}_a^1,\mathbf{z}_b^2),
    \quad \eta^i=\sum_a w_a^i\delta_{\mathbf{z}_a^i},
\end{align}
applies directly to $\eta_R$ and $\eta_{TT}$ and is summed over $\ell$ for the transition direct sum. Hence all inner products and norms below can be evaluated from the corresponding kernel.

\paragraph{sMMD and sCMS.} All measures above are scalar finite signed measures, whose RKHS embeddings are standard \citep{sriperumbudur2011universality}. For $k\in\{k_T,k_R,k_{TT}\}$, define
\begin{align}
    \mathrm{sMMD}_k^2(\mathcal{C}^{\mathcal{M}_1},\mathcal{C}^{\mathcal{M}_2})
    &=\left\Vert\mu_k^1-\mu_k^2\right\Vert^2, \label{smmddef}\\
    \mathrm{sCMS}_k(\mathcal{C}^{\mathcal{M}_1},\mathcal{C}^{\mathcal{M}_2})
    &=\left\langle\overline{\mu}_k^1,\overline{\mu}_k^2\right\rangle
    =\frac{\left\langle\mu_k^1,\mu_k^2\right\rangle}{(\lVert\mu_k^1\rVert+\epsilon)(\lVert\mu_k^2\rVert+\epsilon)}, \label{scmsdef}
\end{align}
where $\overline{\mu}_k^i=\mu_k^i/(\lVert\mu_k^i\rVert+\epsilon)$ and $\rho_k(\mathcal{C}^{\mathcal{M}_i})=\lVert\mu_k^i\rVert/(\lVert\mu_k^i\rVert+\epsilon)$. These extend MMD \citep{gretton2012kernel} and Cosine Mean Similarity \citep{gruber2024disentangling} to finitely supported signed measures; for $k_T$, the norm and inner product are those of the direct sum. 

\paragraph{Representation identities.} Write $f_j(\mathcal{C}^{\mathcal{M}_i})$ for $f_j({\Mt^i}',{\Mr^i}')$. Because vectorisation is an isometry and concatenated blocks occupy orthogonal coordinates,
\begin{align}
    \left\Vert f_1(\mathcal{C}^{\mathcal{M}_1})-f_1(\mathcal{C}^{\mathcal{M}_2})\right\Vert_2^2
    &=\mathrm{sMMD}_{k_T}^2(\mathcal{C}^{\mathcal{M}_1},\mathcal{C}^{\mathcal{M}_2})+\mathrm{sMMD}_{k_R}^2(\mathcal{C}^{\mathcal{M}_1},\mathcal{C}^{\mathcal{M}_2}), \label{idf1}\\
    \left\Vert f_2(\mathcal{C}^{\mathcal{M}_1})-f_2(\mathcal{C}^{\mathcal{M}_2})\right\Vert_2^2
    &=\mathrm{sMMD}_{k_{TT}}^2(\mathcal{C}^{\mathcal{M}_1},\mathcal{C}^{\mathcal{M}_2})+\mathrm{sMMD}_{k_R}^2(\mathcal{C}^{\mathcal{M}_1},\mathcal{C}^{\mathcal{M}_2}), \label{idf2}
\end{align}
where \Eqref{idf2} is \Eqref{finalrepunnorm}. For each normalised block,
\begin{align}
    \left\Vert\overline{\mu}_k^1-\overline{\mu}_k^2\right\Vert^2
    =\rho_k(\mathcal{C}^{\mathcal{M}_1})^2+\rho_k(\mathcal{C}^{\mathcal{M}_2})^2-2\,\mathrm{sCMS}_k(\mathcal{C}^{\mathcal{M}_1},\mathcal{C}^{\mathcal{M}_2}). \label{polar}
\end{align}
Consequently,
\begin{align}
    \left\Vert f_4(\mathcal{C}^{\mathcal{M}_1})-f_4(\mathcal{C}^{\mathcal{M}_2})\right\Vert_2^2
    &=\sum_{k\in\{k_T,k_R\}}\left[\rho_k(\mathcal{C}^{\mathcal{M}_1})^2+\rho_k(\mathcal{C}^{\mathcal{M}_2})^2-2\,\mathrm{sCMS}_k(\mathcal{C}^{\mathcal{M}_1},\mathcal{C}^{\mathcal{M}_2})\right] \label{idf4}\\
    \left\Vert f_5(\mathcal{C}^{\mathcal{M}_1})-f_5(\mathcal{C}^{\mathcal{M}_2})\right\Vert_2^2
    &=\sum_{k\in\{k_{TT},k_R\}}\left[\rho_k(\mathcal{C}^{\mathcal{M}_1})^2+\rho_k(\mathcal{C}^{\mathcal{M}_2})^2\right.\left.-2\,\mathrm{sCMS}_k(\mathcal{C}^{\mathcal{M}_1},\mathcal{C}^{\mathcal{M}_2})\right] \label{idf5}
\end{align}
where \Eqref{idf5} is \Eqref{finalrep}.

Analogous kernel identities also hold for \(f_3\) and \(f_6\), using a different pair-kernel construction. The kernel identities derived above do not directly extend to the nonlinear neural representation \(f_7\).

%% file: neurips_2026.bbl
\begin{thebibliography}{10}

\bibitem{agarwal2021deep}
Rishabh Agarwal, Max Schwarzer, Pablo~Samuel Castro, Aaron~C Courville, and Marc Bellemare.
\newblock Deep reinforcement learning at the edge of the statistical precipice.
\newblock {\em Advances in neural information processing systems}, 34:29304--29320, 2021.

\bibitem{beck2023hypernetworks}
Jacob Beck, Matthew~Thomas Jackson, Risto Vuorio, and Shimon Whiteson.
\newblock Hypernetworks in meta-reinforcement learning.
\newblock In {\em Conference on Robot Learning}, pages 1478--1487. PMLR, 2023.

\bibitem{beck2023survey}
Jacob Beck, Risto Vuorio, Evan~Zheran Liu, Zheng Xiong, Luisa Zintgraf, Chelsea Finn, and Shimon Whiteson.
\newblock A survey of meta-reinforcement learning.
\newblock {\em arXiv preprint arXiv:2301.08028}, 2023.

\bibitem{bowman2016generating}
Samuel Bowman, Luke Vilnis, Oriol Vinyals, Andrew Dai, Rafal Jozefowicz, and Samy Bengio.
\newblock Generating sentences from a continuous space.
\newblock In {\em Proceedings of the 20th SIGNLL conference on computational natural language learning}, pages 10--21, 2016.

\bibitem{cremer2018inference}
Chris Cremer, Xuechen Li, and David Duvenaud.
\newblock Inference suboptimality in variational autoencoders.
\newblock In {\em International conference on machine learning}, pages 1078--1086. PMLR, 2018.

\bibitem{dai2020usual}
Bin Dai, Ziyu Wang, and David Wipf.
\newblock The usual suspects? reassessing blame for vae posterior collapse.
\newblock In {\em International conference on machine learning}, pages 2313--2322. PMLR, 2020.

\bibitem{doshi2016hidden}
Finale Doshi-Velez and George Konidaris.
\newblock Hidden parameter markov decision processes: A semiparametric regression approach for discovering latent task parametrizations.
\newblock In {\em IJCAI: proceedings of the conference}, volume 2016, page 1432, 2016.

\bibitem{duan2016rl}
Yan Duan, John Schulman, Xi~Chen, Peter~L Bartlett, Ilya Sutskever, and Pieter Abbeel.
\newblock Rl$^2$: Fast reinforcement learning via slow reinforcement learning.
\newblock {\em arXiv preprint arXiv:1611.02779}, 2016.

\bibitem{duff2002optimal}
Michael~O'Gordon Duff.
\newblock {\em Optimal Learning: Computational procedures for Bayes-adaptive Markov decision processes}.
\newblock University of Massachusetts Amherst, 2002.

\bibitem{finn2017model}
Chelsea Finn, Pieter Abbeel, and Sergey Levine.
\newblock Model-agnostic meta-learning for fast adaptation of deep networks.
\newblock In {\em International conference on machine learning}, pages 1126--1135. PMLR, 2017.

\bibitem{finn2018probabilistic}
Chelsea Finn, Kelvin Xu, and Sergey Levine.
\newblock Probabilistic model-agnostic meta-learning.
\newblock {\em Advances in neural information processing systems}, 31, 2018.

\bibitem{ghavamzadeh2015bayesian}
Mohammad Ghavamzadeh, Shie Mannor, Joelle Pineau, Aviv Tamar, et~al.
\newblock Bayesian reinforcement learning: A survey.
\newblock {\em Foundations and Trends{\textregistered} in Machine Learning}, 8(5-6):359--483, 2015.

\bibitem{gretton2012kernel}
Arthur Gretton, Karsten~M Borgwardt, Malte~J Rasch, Bernhard Sch{\"o}lkopf, and Alexander Smola.
\newblock A kernel two-sample test.
\newblock {\em The journal of machine learning research}, 13(1):723--773, 2012.

\bibitem{grigsby2024amago}
Jake Grigsby, Justin Sasek, Samyak Parajuli, Daniel Adebi, Amy Zhang, and Yuke Zhu.
\newblock Amago-2: Breaking the multi-task barrier in meta-reinforcement learning with transformers.
\newblock {\em Advances in Neural Information Processing Systems}, 37:87473--87508, 2024.

\bibitem{gruber2024disentangling}
Sebastian~G Gruber, Pascal~Tobias Ziegler, and Florian Buettner.
\newblock Disentangling mean embeddings for better diagnostics of image generators.
\newblock {\em arXiv preprint arXiv:2409.01314}, 2024.

\bibitem{guez2013scalable}
Arthur Guez, David Silver, and Peter Dayan.
\newblock Scalable and efficient bayes-adaptive reinforcement learning based on monte-carlo tree search.
\newblock {\em Journal of Artificial Intelligence Research}, 48:841--883, 2013.

\bibitem{haarnoja2018soft}
Tuomas Haarnoja, Aurick Zhou, Pieter Abbeel, and Sergey Levine.
\newblock Soft actor-critic: Off-policy maximum entropy deep reinforcement learning with a stochastic actor.
\newblock In {\em International conference on machine learning}, pages 1861--1870. Pmlr, 2018.

\bibitem{harrison2018control}
James Harrison, Apoorva Sharma, Roberto Calandra, and Marco Pavone.
\newblock Control adaptation via meta-learning dynamics.
\newblock In {\em Workshop on Meta-Learning at NeurIPS}, volume 2018, 2018.

\bibitem{harrison2018meta}
James Harrison, Apoorva Sharma, and Marco Pavone.
\newblock Meta-learning priors for efficient online bayesian regression.
\newblock In {\em International Workshop on the Algorithmic Foundations of Robotics}, pages 318--337. Springer, 2018.

\bibitem{harrison2024variational}
James Harrison, John Willes, and Jasper Snoek.
\newblock Variational bayesian last layers.
\newblock {\em arXiv preprint arXiv:2404.11599}, 2024.

\bibitem{kaelbling1998planning}
Leslie~Pack Kaelbling, Michael~L Littman, and Anthony~R Cassandra.
\newblock Planning and acting in partially observable stochastic domains.
\newblock {\em Artificial intelligence}, 101(1-2):99--134, 1998.

\bibitem{killian2017robust}
Taylor~W Killian, Samuel Daulton, George Konidaris, and Finale Doshi-Velez.
\newblock Robust and efficient transfer learning with hidden parameter markov decision processes.
\newblock {\em Advances in neural information processing systems}, 30, 2017.

\bibitem{lee2023parameterizing}
Suyoung Lee, Myungsik Cho, and Youngchul Sung.
\newblock Parameterizing non-parametric meta-reinforcement learning tasks via subtask decomposition.
\newblock {\em Advances in Neural Information Processing Systems}, 36:43356--43383, 2023.

\bibitem{mclean2025meta}
Reginald McLean, Evangelos Chatzaroulas, Luc McCutcheon, Frank R{\"o}der, Tianhe Yu, Zhanpeng He, KR~Zentner, Ryan Julian, JK~Terry, Isaac Woungang, et~al.
\newblock Meta-world+: An improved, standardized, rl benchmark.
\newblock {\em arXiv preprint arXiv:2505.11289}, 2025.

\bibitem{melo2022transformers}
Luckeciano~C Melo.
\newblock Transformers are meta-reinforcement learners.
\newblock In {\em international conference on machine learning}, pages 15340--15359. PMLR, 2022.

\bibitem{osband2013more}
Ian Osband, Daniel Russo, and Benjamin Van~Roy.
\newblock (more) efficient reinforcement learning via posterior sampling.
\newblock {\em Advances in Neural Information Processing Systems}, 26, 2013.

\bibitem{perez2020generalized}
Christian Perez, Felipe~Petroski Such, and Theofanis Karaletsos.
\newblock Generalized hidden parameter mdps: Transferable model-based rl in a handful of trials.
\newblock In {\em Proceedings of the AAAI Conference on Artificial Intelligence}, volume~34, pages 5403--5411, 2020.

\bibitem{poupart2006analytic}
Pascal Poupart, Nikos Vlassis, Jesse Hoey, and Kevin Regan.
\newblock An analytic solution to discrete bayesian reinforcement learning.
\newblock In {\em Proceedings of the 23rd international conference on Machine learning}, pages 697--704, 2006.

\bibitem{rakelly2019efficient}
Kate Rakelly, Aurick Zhou, Chelsea Finn, Sergey Levine, and Deirdre Quillen.
\newblock Efficient off-policy meta-reinforcement learning via probabilistic context variables.
\newblock In {\em International conference on machine learning}, pages 5331--5340. PMLR, 2019.

\bibitem{schulman2017proximal}
John Schulman, Filip Wolski, Prafulla Dhariwal, Alec Radford, and Oleg Klimov.
\newblock Proximal policy optimization algorithms.
\newblock {\em arXiv preprint arXiv:1707.06347}, 2017.

\bibitem{shala2025efficient}
Gresa Shala, Andr{\'e} Biedenkapp, Pierre Krack, Florian Walter, and Josif Grabocka.
\newblock Efficient cross-episode meta-rl.
\newblock In {\em The Thirteenth International Conference on Learning Representations}, 2025.

\bibitem{sriperumbudur2011universality}
Bharath~K Sriperumbudur, Kenji Fukumizu, and Gert~RG Lanckriet.
\newblock Universality, characteristic kernels and rkhs embedding of measures.
\newblock {\em Journal of Machine Learning Research}, 12(7), 2011.

\bibitem{strens2000bayesian}
Malcolm Strens.
\newblock A bayesian framework for reinforcement learning.
\newblock In {\em ICML}, volume 2000, pages 943--950, 2000.

\bibitem{todorov2012mujoco}
Emanuel Todorov, Tom Erez, and Yuval Tassa.
\newblock Mujoco: A physics engine for model-based control.
\newblock In {\em 2012 IEEE/RSJ international conference on intelligent robots and systems}, pages 5026--5033. IEEE, 2012.

\bibitem{tziortziotis2013linear}
Nikolaos Tziortziotis, Christos Dimitrakakis, and Konstantinos Blekas.
\newblock Linear bayesian reinforcement learning.
\newblock In {\em IJCAI}, pages 1721--1728, 2013.

\bibitem{vaswani2017attention}
Ashish Vaswani, Noam Shazeer, Niki Parmar, Jakob Uszkoreit, Llion Jones, Aidan~N Gomez, {\L}ukasz Kaiser, and Illia Polosukhin.
\newblock Attention is all you need.
\newblock {\em Advances in neural information processing systems}, 30, 2017.

\bibitem{wang2016learning}
Jane~X Wang, Zeb Kurth-Nelson, Dhruva Tirumala, Hubert Soyer, Joel~Z Leibo, Remi Munos, Charles Blundell, Dharshan Kumaran, and Matt Botvinick.
\newblock Learning to reinforcement learn.
\newblock {\em arXiv preprint arXiv:1611.05763}, 2016.

\bibitem{yang2019single}
Jiachen Yang, Brenden Petersen, Hongyuan Zha, and Daniel Faissol.
\newblock Single episode policy transfer in reinforcement learning.
\newblock {\em arXiv preprint arXiv:1910.07719}, 2019.

\bibitem{yao2018direct}
Jiayu Yao, Taylor Killian, George Konidaris, and Finale Doshi-Velez.
\newblock Direct policy transfer via hidden parameter markov decision processes.
\newblock In {\em LLARLA Workshop, FAIM}, volume 2018, 2018.

\bibitem{yoon2018bayesian}
Jaesik Yoon, Taesup Kim, Ousmane Dia, Sungwoong Kim, Yoshua Bengio, and Sungjin Ahn.
\newblock Bayesian model-agnostic meta-learning.
\newblock {\em Advances in neural information processing systems}, 31, 2018.

\bibitem{yu2020meta}
Tianhe Yu, Deirdre Quillen, Zhanpeng He, Ryan Julian, Avnish Narayan, Hayden Shively, Adithya Bellathur, Karol Hausman, Chelsea Finn, and Sergey Levine.
\newblock Meta-world: A benchmark and evaluation for multi-task and meta reinforcement learning, 2021.

\bibitem{zintgraf2021varibad}
Luisa Zintgraf, Sebastian Schulze, Cong Lu, Leo Feng, Maximilian Igl, Kyriacos Shiarlis, Yarin Gal, Katja Hofmann, and Shimon Whiteson.
\newblock Varibad: Variational bayes-adaptive deep rl via meta-learning.
\newblock {\em Journal of Machine Learning Research}, 22(289):1--39, 2021.

\end{thebibliography}
